\documentclass[dvipsnames]{article}

\usepackage[final]{delta_tuning}

\usepackage[utf8]{inputenc} 
\usepackage[T1]{fontenc}    
\usepackage{url}            
\usepackage{booktabs}       
\usepackage{amsfonts}       
\usepackage{nicefrac}       
\usepackage{booktabs} 
\usepackage{multirow}
\usepackage{wrapfig}
\usepackage{amssymb}
\usepackage{epigraph}
\usepackage{makecell}
\usepackage{listings}

\usepackage[labelfont=bf]{caption} 

\usepackage[ruled,vlined]{algorithm2e}
\usepackage{color, soul}
\usepackage{microtype}      
\usepackage{xcolor}         
\usepackage{geometry}
\usepackage{times}
\usepackage{ragged2e}
\usepackage{amsmath}
\usepackage{bm}
\usepackage{latexsym}
\usepackage{subfigure}
\usepackage{graphicx}
\usepackage{float}
\usepackage{longtable}
\usepackage{booktabs} 
\usepackage{multirow}
\usepackage{amssymb}
\usepackage{longtable}
\usepackage{tabu}
\usepackage{bbding}
\usepackage{awesomebox}
\usepackage{bbding}
\usepackage[most,breakable]{tcolorbox}
\usepackage{etoolbox}
\usepackage{fancyhdr}
\usepackage{lipsum}
\usepackage{CJKutf8}
\usepackage{pdfpages}
\usepackage{multicol}
\usepackage{pgffor}

\usepackage{caption} 
\usepackage{chngcntr}

\usepackage[colorlinks, linkcolor=RoyalBlue, anchorcolor=BrickRed, citecolor=RoyalBlue, urlcolor=RoyalBlue]{hyperref}

\usepackage{cleveref}
\crefname{section}{§}{§§}
\Crefname{section}{§}{§§}

\newcommand\refsec[1]{Section~\hyperref[sec:#1]{\ref{sec:#1}}}
\newcommand\refsecs[2]{\hyperref[sec:#1]{§\ref{sec:#1}:~\textsc{#1}}, \hyperref[sec:#2]{§\ref{sec:#2}:~\textsc{#2}}}

\usepackage[tikz]{bclogo}
\definecolor{msftBlue}{RGB}{0,164,239}
\definecolor{msftGreen}{RGB}{127,186,0}
\definecolor{msftYello}{RGB}{255,185,0}
\definecolor{msftBlack}{RGB}{0,0,0}

\newtcolorbox{myboxnote}[1][]{
  breakable,
  title=#1,
  colback=cyan!0,
  colbacktitle=cyan!0,
  coltitle=black,
  fonttitle=\bfseries,
  bottomrule=0pt,
  toprule=0pt,
  leftrule=1.5pt,
  rightrule=1.5pt,
  titlerule=0pt,
  arc=0pt,
  outer arc=0pt,
  colframe=lightgray,
}
\usepackage{mdframed}

\definecolor{academicblue}{RGB}{54, 95, 145}

\newtcbox{\smybox}[1][red]{on line,
arc=1pt,colback=#1!10!white,colframe=#1!100!black,
before upper={\rule[-3pt]{0pt}{10pt}},
boxsep=0pt,left=6pt,right=6pt,top=2pt,bottom=0pt,boxrule=0pt,leftrule=1pt,rightrule=1pt}

\newenvironment{itemize*}%
 {\leftmargini=20pt\begin{itemize}%
  \setlength{\itemsep}{3pt}%
  \setlength{\parskip}{0pt}%
  }%
 {\end{itemize}}
\newenvironment{enumerate*}%
 {\begin{enumerate}%
  \setlength{\itemsep}{0pt}%
  \setlength{\parskip}{0pt}}%
 {\end{enumerate}}

\usepackage{fancyhdr} 
\usepackage{blindtext} 
\usepackage{ulem}

\DeclareMathOperator*{\argmax}{arg\,max}

\DeclareMathOperator*{\E}{\mathbb{E}}

\usepackage{xparse}
\NewDocumentCommand{\heng}
{ mO{} }{\textcolor{red}{\textsuperscript{\textit{Heng}}\textsf{\textbf{\small[#1]}}}}

\usepackage{colortbl}

\title{
Tool Learning with Foundation Models
}

\author{Yujia Qin$^1$, Shengding Hu$^1$, Yankai Lin$^{2}$\thanks{ Corresponding authors.}\hspace{0.5em}, Weize Chen$^1$, Ning Ding$^1$, Ganqu Cui$^1$, \vspace{0.03in}\\ \textbf{Zheni Zeng$^1$, Xuanhe Zhou$^1$, Yufei Huang$^1$, Chaojun Xiao$^1$, Chi Han$^3$, Yi Ren Fung$^3$,} \vspace{0.03in}\\ \textbf{Yusheng Su$^1$, Huadong Wang$^1$, Cheng Qian$^1$, Runchu Tian$^1$, Kunlun Zhu$^8$, Shihao Liang$^8$, } \vspace{0.03in}\\ \textbf{Xingyu Shen$^1$, Bokai Xu$^1$, Zhen Zhang$^1$, Yining Ye$^1$, Bowen Li$^1$, Ziwei Tang$^5$, Jing Yi$^1$, } \vspace{0.03in}\\ \textbf{Yuzhang Zhu$^1$, Zhenning Dai$^1$, Lan Yan$^1$, Xin Cong$^1$, Yaxi Lu$^1$, Weilin Zhao$^1$, } \vspace{0.03in}\\ 
\textbf{Yuxiang Huang$^1$, Junxi Yan$^1$, Xu Han$^1$, Xian Sun$^7$, Dahai Li$^7$, Jason Phang$^4$, } \vspace{0.03in}\\ 
\textbf{Cheng Yang$^5$, Tongshuang Wu$^6$, Heng Ji$^3$, Guoliang Li$^{1}$, Zhiyuan Liu$^{1*}$, Maosong Sun$^{1*}$} \vspace{0.1in}\\ 
$^1$Tsinghua University, $^2$Renmin University of China, $^3$University of Illinois Urbana-Champaign,\vspace{0.03in}\\ 
$^4$New York University, $^5$Beijing University of Posts and Telecommunications, \vspace{0.03in}\\
$^6$Carnegie Mellon University, $^7$Zhihu Inc., $^8$ModelBest Inc.\vspace{0.03in}\\
\texttt{qyj20@mails.tsinghua.edu.cn}\\
}

\begin{document}

\maketitle

\vspace{-0.5cm}
\begin{abstract}
Humans possess an extraordinary ability to create and utilize tools, allowing them to overcome physical limitations and explore new frontiers. With the advent of recent powerful foundation models, artificial intelligence systems have the potential to be equally adept in tool use as humans. This paradigm, which is dubbed as \textit{tool learning with foundation models}, combines the strengths of specialized tools and foundation models to achieve enhanced accuracy, efficiency, and automation in problem-solving. Despite its immense potential, there is still a lack of a comprehensive understanding of key challenges, opportunities, and future endeavors in this field. To this end, we present a systematic investigation and comprehensive review of tool learning in this paper. We first introduce the background of tool learning, including its cognitive origins, the paradigm shift of foundation models, and the complementary roles of tools and models. We recapitulate existing tool learning research and formulate a general tool learning framework: starting from understanding the user instruction, models should learn to decompose a complex task into several subtasks, dynamically adjust their plan through reasoning, and effectively conquer each sub-task by selecting appropriate tools. We also discuss how to train models for improved tool-use capabilities and facilitate the generalization in tool learning. 
Considering the lack of a systematic tool learning evaluation in prior works, we experiment with $18$ representative tools and show the potential of current foundation models in skillfully utilizing tools.
Finally, we discuss several open problems that require further investigation for tool learning, such as ensuring safe and trustworthy tool use, enabling tool creation with foundation models, and addressing personalization challenges.
Overall, we hope this paper could inspire future research in integrating tools with foundation models. Relevant codes and datasets are publicly available for further research exploration\footnote[1]{\url{https://github.com/OpenBMB/BMTools} \\ \quad Author contributions are listed in \cref{sec:contributions}.}.

\end{abstract}

\vspace{-0.3cm}
\hspace{180pt}\parbox[b]{0.45\textwidth}
{
\epigraph{\textit{``It is not only the violin that shapes the violinist, we are all shaped by the tools we train ourselves to use.''}}{--- Edsger W. Dijkstra} 
}

\newpage
{
  \hypersetup{linkcolor=RoyalBlue, linktoc=page}
  \tableofcontents
}

\newpage

\section{Introduction}
\label{sec:introduction}

Tools are extensions of human capabilities designed to enhance productivity, efficiency, and problem-solving in human activities. Since the dawn of civilization, tools have been integral to the very essence of our existence~\citep{washburn1960tools}. Tool creation and utilization are motivated by a deep-rooted desire to overcome our physical limitations and discover new territories. 
More specifically, with advancements in tools, we can accomplish increasingly complex tasks with ease and efficiency, liberating time and resources to pursue more ambitious ventures. 
As such, tools have served as the crucial foundation upon which our cultural and social practices are built, transforming our modes of learning, communication, working, and entertainment, infusing these domains with new dimensions of accessibility and interactivity~\citep{gibson1993tools}. 
Throughout history, it is undeniable that human beings have played a pivotal role in the invention and manipulation of tools, which is a striking manifestation of intelligence~\citep{shumaker2011animal}.
Given the rise of Artificial Intelligence (AI), one natural question is, does AI possess the potential to be equally adept and capable as its creators?

The prerequisite of the manipulation of tools is a thorough comprehension of the tools' functionalities, as well as the ability to understand user intents and perform planning and reasoning for tool use. 
Before the advent of powerful foundation models~\citep{bommasani2021opportunities}, conducting tool-oriented AI research was exceedingly challenging. While certain basic tools could be fitted using shallow statistical models or deep neural models~\citep{pomerleau1988alvinn,mnih2013playing,akkaya2019solving}, their performance and stability remained inadequate to meet the demands of practical applications, let alone generalizing across various tools. This is due to the limitations of traditional supervised learning in capturing the complex operations essential for tool utilization and the insufficiency of trial-and-error approaches like reinforcement learning in mastering the extensive decision space associated with tool use.
In a nutshell, the fundamental limitations in tool use by earlier AI lie in the insufficient capabilities of the models.
Recently, the emergence of more capable foundation models, characterized by significantly improved capabilities, has rendered tool learning practicable.
They have shown enormous semantic understanding capacity in diverse tasks, spanning the fields of natural language processing (NLP)~\citep{brown2020language}, computer vision (CV)~\citep{ramesh2022hierarchical}, biology~\citep{jumper2021highly}, etc.
Additionally, they have demonstrated superior reasoning and decision-making abilities in complex interactive environments~\citep{nakano2021webgpt}. 
By harnessing the extensive world knowledge garnered during pre-training, they can perform grounded actions and interact with the real world. 
Notably, the emergence of ChatGPT~\citep{openaichatgptblog} highlights the potential of foundation models to understand human intentions, automate intricate processes, and generate natural responses; the advent of GPT-4~\citep{openai2023gpt4} offers immense potential for multi-modal perception, which is essential to the real-world grounding ability.

\begin{figure}[!b]
    \centering
    \includegraphics[width=\textwidth]{figures/intro.pdf}
    \caption{Tool learning paradigm aims to combine the strengths of specialized tools and foundation models.}
    \label{fig:introduction}
\end{figure}

Therefore, foundation models enable AI to harness tools, which can lead to more potent and streamlined solutions for real-world tasks.
Foundation models are able to decipher complex data, simulate human-like planning capabilities, and generate a broad spectrum of outputs.
Concurrently, specialized tools can be employed to refine and target specific goals. The amalgamation of tools and models unveils vast potential where sophisticated procedures can be automated with limited human involvement.
This paradigm, dubbed as \textbf{tool learning with foundation models} in this paper (Figure~\ref{fig:introduction}), aims to combine the strengths of specialized tools and foundation models, thereby culminating in greater accuracy, efficiency, and autonomy in problem-solving. 
Recent research has shed light on foundation models' potential to exhibit a level of dexterity and finesse in tool use~\citep{lazaridou2022internet,nakano2021webgpt,cobbe2021training,thoppilan2022lamda,huang2022inner,ahn2022can,yao2022webshop,yao2022react,schick2023toolformer,wu2023visual,bubeck2023sparks}. 
Despite these breakthroughs, the efforts mainly focus on applying foundation models to specific tasks and domains with delicate algorithm designs.
The current understanding of tool learning is still not comprehensive enough to estimate its characteristics and future developments. 
We believe that it is crucial to examine and summarize the current progress of tool learning with foundation models to explore their potential and challenges and to better pave the way for future technological advancements.

In this paper, we conduct a systematic investigation and comprehensive review of tool learning, attempting to build a full grasp of the key challenges, opportunities, and directions in this field.
Before delving into the tool learning framework, we introduce essential \textbf{backgrounds} (\cref{sec:backround}), covering both tools and foundation models and their interaction. Specifically, we first recapitulate the cognitive origins of tool use in human history and its potential implications for tool use in AI systems (\cref{sec:cognitive_origin}), followed by a categorization of tools from the perspective of the user interface (\cref{sec:tool_categorization}). Then we review the AI paradigm shift brought by foundation models and highlight the emergence and significance of tool learning (\cref{sec:pradigm_shift}). After that, we discuss the complementary roles of tools and foundation models, and argue that integrating both can bring various advantages, such as improving interpretability, enhancing robustness, and delivering better user experiences (\cref{sec:complementary_role}). 

We present a comprehensive literature review for existing exploration in tool learning. While previous works often focus on specific aspects in isolation, we strive to formulate a general tool learning \textbf{framework} (\cref{sec:overview}), which comprises the controller (typically modeled using a foundation model), tool set, environment, perceiver, and human. Based on the unified framework, we review existing works of tool learning, highlight core research problems, and introduce their existing solutions as well as future explorations. The whole procedure (\cref{sec:tool_learning_procedure}) of tool learning starts with a user instruction, and models are required to make an executable plan for tool execution. To bridge user instructions with appropriate tools, models should first learn to understand the user intents underlying the instruction (i.e., \textit{intent understanding}) and understand the functionalities and usage of tools (i.e., \textit{tool understanding}). Models should also learn to decompose a complex task into several subtasks, dynamically adjust their plan through reasoning, and effectively conquer each sub-task with the appropriate tools. Regarding the training strategy (\cref{sec:training_strategy}) to facilitate models for improved tool utilization, we conclude with two mainstream methods: learning from demonstrations and learning from feedback. We discuss how to construct effective training supervision under different settings. To facilitate transferring the learned tool-use skills to new tools and situations, i.e., generalizable tool learning, it is important to design a unified interface that enables the model to interact with different tools in a standardized manner.

Considering the lack of a systematic tool learning evaluation in prior works, we conduct \textbf{experiments}~(\cref{sec:applications}) on $18$ representative tools based on our framework to investigate the efficacy and limitations of foundation models in tool manipulation. We demonstrate that state-of-the-art foundation models (e.g., ChatGPT) can effectively use tools to solve tasks with simple prompting. These results highlight the potential of using the foundation model as a general agent for tool learning.

Finally, we discuss the remaining important \textbf{research topics} (\cref{sec:discussion}) for applying our general framework to real-world scenarios, including (1) \textbf{safety and trustworthiness}, where we emphasize the potential risks from adversaries, governance, and trustworthiness. We contend that careful considerations are necessary before deploying tool learning models in high-stakes scenarios (\cref{sec:safe_tool_learning}); (2) \textbf{tool learning for large complex systems}, where we showcase the unique characters of large complex systems and discuss the challenges in applying tool learning to these systems, such as complicated knowledge and function learning, representative data sampling with privacy concerns, and the strict requirements of efficient tool learning (\cref{sec:large-system}); (3) \textbf{tool creation}, where we discuss the possibility that AI can also create new tools, challenging the long-held beliefs about what makes humans unique (\cref{sec:tool_creation}); 
(4) \textbf{personalized tool learning}, where models provide tailored assistance to users in tool use. We highlight the challenges of aligning user preference with tool manipulation and introduce the shift from reactive to proactive systems, and the privacy-preserving concerns (\cref{sec:personalization}); 
(5) \textbf{embodied learning}, where the intersection of tool learning and embodied agent enables digital embodiment and manipulation of embodied tools (\cref{sec:embodied_learning}); (6) \textbf{knowledge conflicts in tool augmentation}, where we review how tools can be leveraged to enhance models' generation and the practical problems of knowledge conflicts, which can lead to inaccurate and unreliable model predictions (\cref{sec:knowledge_conflicts}); (7) other \textbf{open problems}, such as viewing tool use capability as a measure for machine intelligence and tool learning for scientific discovery (\cref{sec:open_problems}). Overall, we hope this paper could inspire further research in integrating tools with foundation models and developing more intelligent and capable AI systems.
\section{Background}
\label{sec:backround}

In this section, we first discuss the cognitive origins of human tool use (\cref{sec:cognitive_origin}), followed by a tool categorization through the lens of the user interface (\cref{sec:tool_categorization}). Then we review the recent AI paradigm shift brought by foundation models (\cref{sec:pradigm_shift}) and its significance in tool learning. After that, we examine the respective roles of specialized tools and foundation models in problem-solving, and discuss the benefits and challenges of their integration (\cref{sec:complementary_role}).

\subsection{Cognitive Origins of Tool Use}
\label{sec:cognitive_origin}
Tools have played a critical role in the long history of thousands of years of human evolution. Tools are commonly viewed as extensions of human beings, just as ancient fighting equipment and agricultural machinery were. Tool use is defined as a unique characteristic of human beings that is distinguished from other species~\citep{von1995cognitive}. Throughout evolution, the ability to use tools has been essential for animals, particularly those with advanced intellectual development~\citep{shumaker2011animal}. 
For example, chimpanzees have been observed using stones or other materials to crack nuts~\citep{boesch2019learning}, while New Caledonian crows can craft and utilize two distinct types of hook tools to aid in capturing prey~\citep{hunt1996manufacture}. However, human tool behavior diverges from these observations in several ways. Humans can create much more complicated tools than other animals, such as converting our actions into fundamentally different mechanical actions in tools like hammers and pencils~\citep{frey2007puts}. Additionally, we can harness natural forces such as wind turbines to create tools~\citep{shumaker2011animal}. This ability may be attributed to our deep comprehension of cause-and-effect relations, which allows us to engage in technical reasoning~\citep{osiurak2020elephant}.

\textbf{Neural Basis of Tool Use.}
To better understand human tool use behaviors, researchers analyze the neural basis of tool observation and execution. It is verified that humans have parietal systems involved in grasping objects and using tools, and the anterior supramarginal gyrus activation of observing tool use is typical of human subjects, of which macaques do not exhibit~\citep{orban2014neural}. This neurocognitive bases of tool observation may be related to the origins of cumulative technological evolution (e.g., the improvement in the efficiency and complexity of human tools and techniques over generations~\citep{osiurak2020elephant}) and some other human phenomena~\citep{reynaud2019watch}. While for the tool execution, researchers hold different views on manipulation-based versus reasoning-based approaches~\citep{osiurak2016tool}. The former claims that tool use has to be supported by the simulation of sensorimotor experiences, and the latter demonstrates the importance of reasoning based on mechanical knowledge in tool use. Nevertheless, the overall trend in cognitive science is understanding cognition as an enactive process that emphasizes interaction with the external world~\citep{engel2013s}, and the feedback from observation, communication, and hands-on practice is important for mastering tool use. 

\textbf{Three Intelligence Levels of Tool Use.}
Besides, there are specific frameworks designed to discuss the level of intelligence represented by human tool use. For instance, ``intoolligence''~\citep{osiurak2018looking} divides the tool use behavior into three modes: \textit{assistive tool use} is usually passive and unaware (e.g., walking in the rain shelter corridor); \textit{arbitrary tool use} requires active interaction (e.g., driving, using smart phones); \textit{free tool use} further needs to comprehend and choose appropriate tools for the scenarios (e.g., cooking new dishes). In this framework, the three modes of tool use present a progressive relationship, and the authors assume that the key cognitive process for achieving free tool use is technical reasoning, which allows someone to learn new actions by observing others using, selecting, or making a tool instead of numerous practices. 

\textbf{Transition from Physical Tools to Conceptual Tools.} 
Apart from tools in the physical world, we can also turn to more abstract tools. Take cognitive tools~\citep{heyes2018cognitive} as an example: it refers to an auxiliary aid that facilitates higher-order thinking (e.g., multi-step critical analysis, the generation of creative problem-solving solutions). Cognitive tools can be classified based on the functionalities they provide~\citep{lajoie2013computer}. These include (1) supporting cognitive processes (e.g., documenting intermediate reasoning outcomes), (2) alleviating lower-level cognitive load to free up resources for advanced-level thinking, (3) enabling learners to engage in activities out of their reach and (4) allowing learners to generate and test hypotheses (e.g., simulated diagnoses for medical students).

\textbf{Bridging the Gap between Human and Machine Tool Use.}
First, the abilities to manipulate tools are deeply rooted in our cognitive and perceptual systems and have evolved over millions of years. In contrast, foundation models rely primarily on statistical patterns of pre-training data, and significant gaps still exist between the tool-use capabilities of foundation models and their human counterparts. Humans can perceive the properties of tools, understand their functionalities, and identify the appropriate tools for each task. Gaining insights from this, we investigate and discuss how foundation models can learn such a process in \cref{sec:aligning_user_tools}. 
Second, humans excel at breaking down complex tasks into smaller, manageable subtasks and deftly manipulating tools to accomplish each sub-task. However, foundation models lack the physical embodiment and sensory experience necessary to fully understand and utilize tools. As a result, these models often struggle with tasks that require higher-order reasoning and adaptivity, and they cannot trustfully integrate multiple sources of knowledge and tools effectively. We will discuss how to better make executable plans leveraging models' reasoning abilities in \cref{sec:framework_reasoning}. Furthermore, current algorithms for adapting foundation models to learn specific tools generally require huge amounts of supervised data~\citep{nakano2021webgpt,reed2022generalist}, hindering their generalization and transferability to broader types of tools or novel situations. Hence we first summarize the training strategies for tool learning (\cref{sec:learning_from_demonstrations} and \cref{sec:learning_from_feedback}) and discuss how to facilitate the generalization and transferability of tool learning (\cref{sec:generalizable_tool_learning}).

\subsection{Tool Categorization: A User-Interface Perspective}
\label{sec:tool_categorization}

\begin{figure}[!t]
    \centering
    \includegraphics[width=\linewidth]{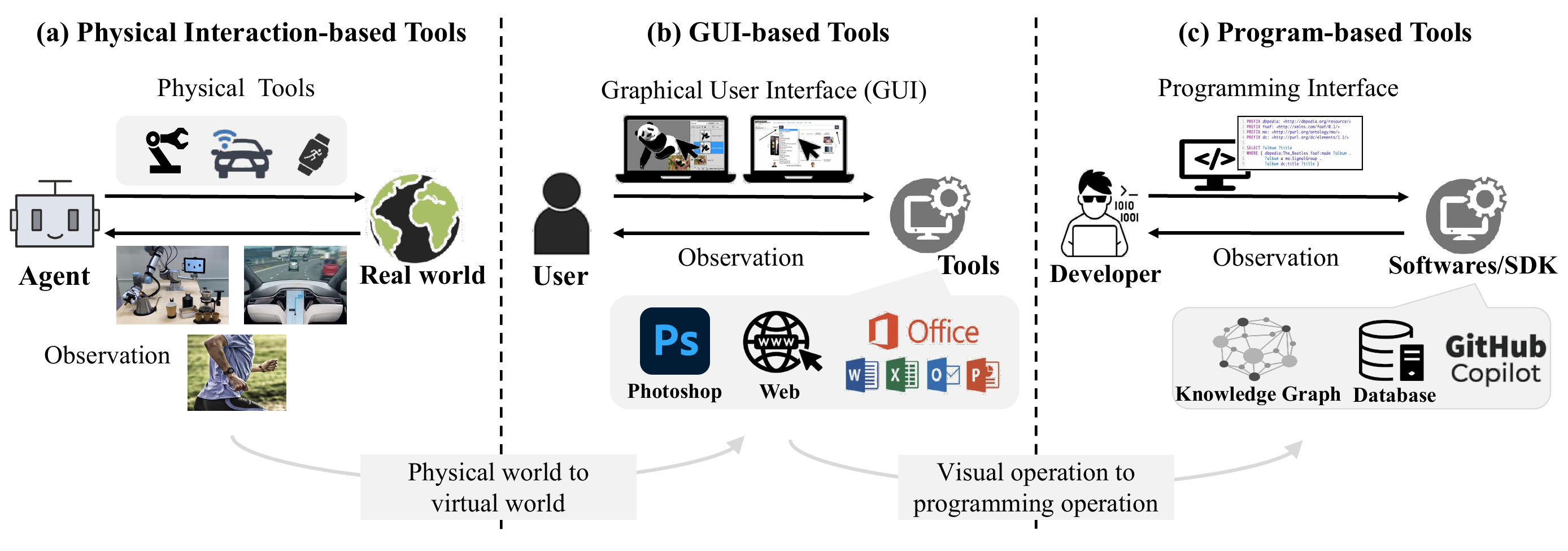}
    \caption{Tool categorization from the perspective of the user interface: (1) physical interaction-based tools, (b) GUI-based tools, and (c) program-based tools.}
    \label{fig:tool_categorization}
\end{figure}

The growing number and complexity of tools in our world make it increasingly important to understand and group them in a meaningful way. A system for classifying these tools helps us better grasp their uses, benefits, and potential for growth. In this paper, our focus is particularly on those tools that can be manipulated using instructions in conjunction with foundation models. We introduce a taxonomy that sorts these tools based on their modes of expression and interaction. As depicted in Figure~\ref{fig:tool_categorization}, this taxonomy incorporates three levels of interaction, arranged from the most tangible to the least. {\it The physical level} involves direct physical interactions with tools. {\it The graphical user interface} (GUI) level facilitates user interaction with visual representations of tools. {\it The program level} involves users engaging directly with the underlying source code of tools.

\textbf{Physical Interaction-based Tools.} We start with the most tangible genre of tools, physical interaction-based tools.
As the name suggests, this class of tools involves direct interactions with the physical world, including devices like robots, sensors, and wearables that could physically impact the environment. 
Physical interaction tools have the capability to sense and respond to the physical environment of users, making them useful in a wide range of applications, from manufacturing to healthcare and education.
Physical interaction tools are close to the real world, and they have the potential to substantially improve efficiency and productivity.
For example, robots can perform from simple to intricate, even adventurous tasks to reduce human errors and labor costs. Sensors can collect valuable data, such as temperature and pressure, allowing for real-time monitoring and optimization of industrial processes. 
Wearables, on the other hand, provide users with a personalized experience by tracking physiological or environmental parameters, promoting health and safety.
It is worth noting that although the output of such tools interacts with the real world at the physical level, users may also create the input of the tools at the GUI or source code level.

\textbf{GUI-based Tools.} 
Some tools allow users to manipulate them through an interactive interface, i.e., visual representations of tools, with pre-defined operations. 
These tools, defined as GUI-based tools, do not have a direct impact on the physical world. 
The GUI interface typically includes buttons, menus, text boxes, and other graphical elements that allow users to interact with the underlying system. 
These tools are extensively employed in various industries and applications such as software development, data analysis, and design.
Particularly, GUI-based tools can improve productivity by streamlining workflows and automating repetitive tasks.
GUI-based tools could considerably simplify complex tasks and reduce the learning curve for non-technical users. 
From this viewpoint, tool learning with foundation models share the same primary goal, which simplifies intricate tasks to a natural language format.
Representative GUI-based tools are usually well-developed software such as browsers, Microsoft Office, Adobe PhotoShop, etc.
These applications showcase the versatility that graphical interfaces can provide and enable users to access and manipulate complex features within the software.
On the other hand, the main limitation of GUI-based tools is that they may not provide the flexibility and customizability of command-line interfaces or APIs. 
In specific scenarios that require rapid and mass responses, as well as greater and flexible control, the visual interface may not always be the most effective way to interact with a system.

\textbf{Program-based Tools.} 
The innermost layer of tools that users can access is the source code, offering a high degree of flexibility for the input and output of these program-based tools. 
Program-based tools are software tools primarily designed for use through programming interfaces rather than visual interfaces.  
They can take various forms, including declarative languages, programming libraries, software development kits (SDKs), and even neural network-based tools.
These tools are typically used by developers or technical users who possess a deeper understanding of the underlying data, system or technology, with which the users could complete complex software applications.
The main advantage of program-based tools is that they provide greater flexibility and customizability than GUI-based tools, and users can build more sophisticated solutions for current problems. As a result, such tools also have a steeper learning curve than GUI-based tools, they require a greater degree of technical expertise and programming knowledge, which may not be accessible to non-technical users. 
For example, program-based tools can be more time-consuming to set up and configure and may require more maintenance and support in the learning process.
It is noteworthy that, although these tools pose difficulties for human beings in terms of the learning curve, they may not have the same level of challenges for foundation models.

It can be seen that the above three interaction modes have varying levels of connectivity with the tool kernel. They are not strictly mutually exclusive but indicate a tendency to intermingle with each other. Human beings have the ability to deal with complex tasks by flexibly executing tools of different types. In this paper, we contend that regardless of the tool type, it is fundamentally possible to leverage foundation models to execute them by setting up intermediary interfaces. We will introduce ways to unify the interface of different tools for foundation models in \cref{sec:generalizable_tool_learning}.

\subsection{Paradigm Shift of Foundation Models}
\label{sec:pradigm_shift}
In recent years, the field of natural language processing (NLP) has undergone a paradigm shift, marked by the advent of pre-trained language models (PLMs)~\citep{devlin2018bert,bommasani2021opportunities,HAN2021}. 
Prior to this breakthrough, NLP was a challenging field that necessitated designing separate learning objectives for distinct research domains, such as dependency parsing~\citep{kubler2009dependency}, named entity recognition~\citep{nadeau2007survey}, and summarization~\citep{nenkova2012survey}.
Despite the successful design of effective models and methods for these specific tasks, the separated nature of this paradigm impeded progress toward a holistic comprehension of language, thereby limiting its potential.

The invention of PLMs changes this paradigm. Building on Transformers~\citep{vaswani2017attention}, PLMs are trained on massive corpora, from which general linguistic ability and world knowledge are learned. 
This technique has expedited the unification of NLP tasks, giving rise to the \textit{pre-train-then-fine-tune} paradigm, which has achieved new state-of-the-art performance on several NLP benchmarks, such as GLUE~\citep{wang2018glue} and SuperGLUE~\citep{wang2019superglue}. 
At this stage, each task shares the same starting point and only diverges as the task-specific adaptation proceeds. The fusion of task paradigms is still ongoing. T5~\citep{raffel2019exploring} transforms all NLP tasks into a text-to-text format with textual descriptions, while GPT-3~\citep{brown2020language} has discovered that introducing appropriate textual prompts can yield the desired output for specific tasks. Prompts, essentially serving as a natural language interface, are widely believed to stimulate the knowledge learned by PLMs during pre-training. Prompts can enable downstream tasks to be executed without updating model parameters for big models such as GPT-3. Research even suggests that with appropriate prompt guidance, models can perform complex reasoning tasks~\citep{wei2022chain,wang2022self}.
Also, prompts formulated in a natural language format possess remarkable generalization capabilities. Specifically, models that have undergone fine-tuning using diverse instructions are able to effectively generalize to new, unseen data~\citep{wei2021finetuned,sanh2021multitask}.
Overall, prompts demonstrate a proof-of-concept that uses PLMs as the underlying infrastructure and natural language as the medium to uniformly perform various tasks. 
A highly successful example is ChatGPT, where all the natural language understanding and generation processes are accomplished through conversational interactions.

Nevertheless, there exist numerous tasks that transcend the scope of purely natural language. 
For instance, generating presentation slides\footnote{\url{https://www.microsoft.com/en-us/microsoft-365}}, constructing 3D models via CAD applications, and scheduling meetings through the analysis of team member calendars are examples of complex tasks that have not been defined in traditional artificial intelligence. 
Fortunately, the strong generalization ability of PLM enables us to use natural language as a medium to accomplish these tasks by manipulating tools~\citep{zeng2022socratic}.
Essentially, the key to tool learning is to decompose complex tasks into sub-actions, tokenize actions in the form of natural language and convert them into executable instructions that can be understood by specific tools.
Language models serve as ``translators'', making complex tasks more accessible to individuals without specialized technical knowledge.
The potential applications of tool learning are vast and exciting, ranging from automated customer service and personal assistants to self-driving cars and even space exploration. 
By enabling machines to understand and interact with human language in a more natural and nuanced way, we can unlock new possibilities for collaboration and problem-solving that were previously impossible.
We anticipate that tool learning will prove instrumental in facilitating the integration of diverse tasks through shared tooling. 
Thus, while natural language interfaces have enabled unification within the realm of language~\citep{hao2022language}, the challenges posed by non-linguistic tasks necessitate a more advanced approach to leveraging both natural language and tool learning. By harnessing the power of natural language, we can create systems that are capable of understanding and adapting to the complex and dynamic world around us, opening up new avenues for innovation and discovery.

\subsection{Complementary Roles of Tools and Foundation Models}
\label{sec:complementary_role}

The integration of specialized tools and foundation models represents a promising approach for harnessing the unique strengths of both.
By incorporating foundation models' understanding and reasoning capabilities into specialized tools, we can create intelligent tools capable of performing more complex tasks than either specialized tools or foundation models alone. Specifically, the amalgamation of both confers a multitude of benefits as follows.

\paragraph{Benefits of Tools.} Tools that are designed to streamline concrete and specific objectives bring several benefits for tool learning: (1) \textbf{Mitigation for Memorization.}
Although foundation models have demonstrated an exceptional ability to memorize~\citep{carlini2021extracting,carlini2022quantifying,carlini2023extracting}, they are not capable of memorizing every piece of training data. Furthermore, foundation models are often prompted with a relatively short context during model generation, thus not all the memorized knowledge can be properly steered~\citep{mialon2023augmented}. Additionally, memorization alone does not support the real-time coverage of up-to-date knowledge, especially in light of the potentially infinite possibilities of novel requests from users. Besides, foundation models are also criticized to hallucinate knowledge~\citep{roller2020recipes,shuster2021retrieval} by generating seemingly plausible but non-factual content. This is unacceptable in applications like financial transactions that require the results are 100\% correct. Given the above factors, it is necessary to augment foundation models with real-time tool execution to mitigate limitations in memorization. For instance, a significant proportion of the memorization burden can be offloaded to the search engine and database systems (of different modes) if foundation models can learn how to utilize it.
(2) \textbf{Enhanced Expertise.} Specialized tools are designed to cater to specific domains with functionalities that are not available in foundation models. As a result, they are better suited to address the needs of domain-specific tasks, such as Wolfram~\footnote{\url{https://www.wolframalpha.com/}} for scientific calculation, through the utilization of tailored algorithms. Instead of solely relying on the foundation model to accomplish the task, models could invoke appropriate tools to generalize to a wider range of tasks that are beyond their capabilities.
(3) \textbf{Better Interpretability.}
Foundation models are criticized for lacking transparency in their decision-making process~\citep{linardatos2020explainable}, which can be a significant concern in applications such as healthcare or finance, where interpretability is critical for making informed decisions. In contrast, the process of tool execution reflects the whole process of how models solve complex requests, which allows for better interpretability and transparency. Users can easily understand why certain tools are called and how they contribute to the final output, which can improve trust and facilitate human-machine collaboration.  
(4) \textbf{Improved Robustness.}
Foundation models are susceptible to adversarial attacks~\citep{wallace-etal-2019-universal,jin2020bert}, where slight modifications to the input can flip the model prediction. This is because these models heavily rely on statistical patterns in the training data. Conversely, tools are designed specifically for their intended use cases, which may be agnostic to the input perturbation. This makes tools more resistant to adversarial attacks. Overall, incorporating tools into the workflow of foundation models can improve the robustness of the system and reduce the risk of malicious attacks. This harmonious interplay between tools and models can enhance the reliability of the system against unpredictable real-world environments.
In \cref{sec:applications} and \cref{sec:case_study}, we use concrete examples to show how tools can enhance the model's capabilities in various tasks.

\paragraph{Benefits of Foundation Models.} Foundation models can provide a solid basis for understanding, planning, reasoning, and generation, which bring several benefits for tool learning as follows: 
(1) \textbf{Improved Decision-Making and Reasoning Abilities.}
Foundation models are trained on vast amounts of data, enabling them to acquire world knowledge across a wide range of domains. If properly steered, such knowledge can be wielded to perform decision-making and planning over prolonged time horizons~\citep{huang2022language}. 
Besides, foundation models have demonstrated remarkable reasoning abilities~\citep{wei2022chain,wang2022self}, thereby enabling them to extrapolate the consequences of actions and make judicious decisions. These reasoning abilities are particularly useful for tasks requiring a deep understanding of cause-and-effect relations~(\cref{sec:framework_reasoning}). 
(2) \textbf{Better User Experience.} Benefitting from the powerful intent understanding capability of foundation models, tool learning could revolutionize the way we interact with machines and liberate users from the cognition load, allowing them to engage in higher-order thinking and decision-making processes. This, in turn, fosters a seamless and more natural language-based interaction paradigm that revolutionizes traditional graphical user interfaces (GUIs). The user only needs to provide high-level guidance and direction, and the model will seamlessly comprehend the user's intent, thereby delivering more personalized and precise responses. In addition, tool learning has the potential to democratize access to complex tools. With the aid of foundation models, even novice users can easily and quickly get started with a new tool, regardless of their prior experience or technical expertise. This not only reduces the barriers to entry for new users but also unlocks a wealth of possibilities for innovation and creativity. However, it should be noted that human-model collaboration in tool use also triggers ethical concerns, which will be discussed in \cref{sec:open_problems}.
\section{Tool Learning}
\label{sec:formal_description}
In this section, to unify existing efforts and promote a comprehensive understanding of tool learning, we first present a general framework, which encompasses four fundamental components, namely tool set, environment, controller, and perceiver, as detailed in~\cref{sec:overview}. Subsequently, we provide an elaborate discussion of the general procedure of tool learning in~\cref{sec:tool_learning_procedure}. Lastly, we delve into the training methods for tool learning and discuss how to achieve generalizable tool learning in~\cref{sec:training_strategy}. 

\subsection{Components of Tool Learning}
\label{sec:overview}

\begin{figure}[!t]
    \centering
    \includegraphics[width=0.78\textwidth]{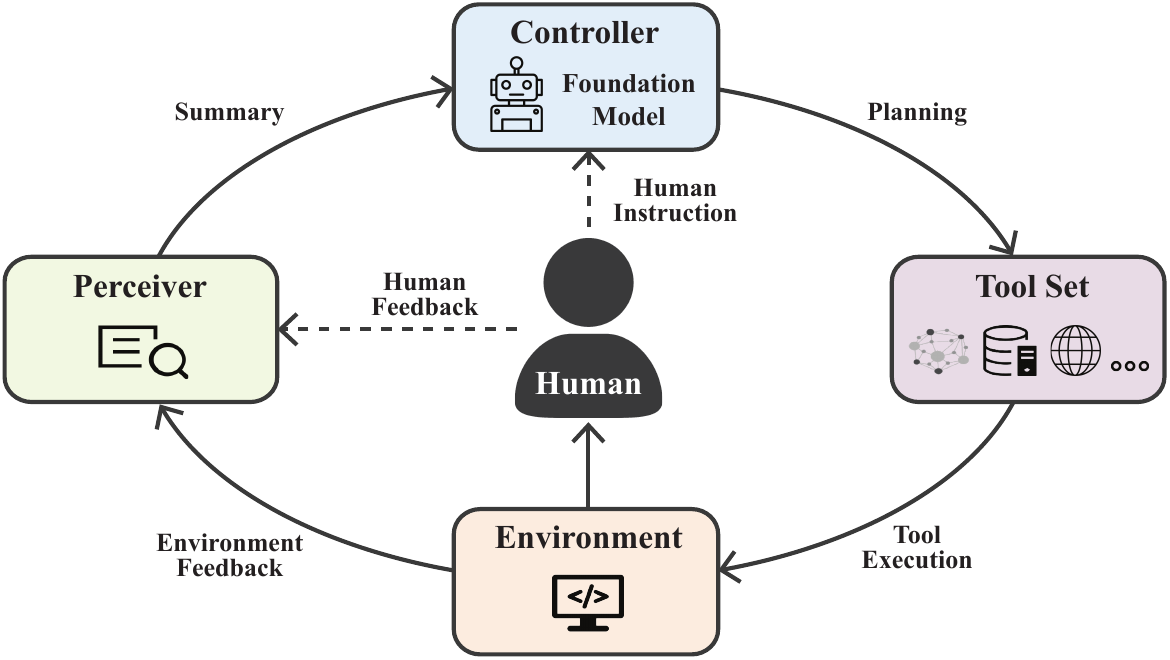}
    \caption{Illustration of the tool learning framework, where we display the human user and four core ingredients of the framework: tool set, controller, perceiver, and environment. The user sends an instruction to the controller, which then makes decisions and executes tools in the environment. The perceiver receives feedback from both the environment and the user and summarizes them to the controller.}
    \label{fig:formulation}
\end{figure}

How can we enable foundation models to leverage the strengths of specialized tools to accomplish complex tasks? To better answer this question, we frame tool learning with four components as shown in Figure~\ref{fig:formulation}. Each component has its own characteristics and functions~(\cref{sec:framework_pillar_understand}), but they also interact with each other closely~(\cref{sec:framework_pillar_connect}). 

\subsubsection{Understanding the Components}
\label{sec:framework_pillar_understand}

We first introduce each component and explain how they contribute to the tool learning process.

\paragraph{Tool Set.}
Serving as the fundamental ingredient of tool learning, the tool set $\mathcal{T} = \{\mathcal{T}_1, \mathcal{T}_2, \cdots \}$ contains a collection of different tools that have different functionalities. As we have elaborated in~\cref{sec:tool_categorization}, a tool in $\mathcal{T}$ can have different interfaces. In the following sections, we mainly take Application Programming Interface (API) as the example to illustrate how to interact with tools. Here we define an API as any function that can take the output of the foundation model as its input. For instance, for a weather API, the input to the API may be a location and time, and the output may contain the temperature or wind speed.

\paragraph{Environment.}
The environment $\mathcal{E}$ is the world where the tools operate, which provides the \textit{perceiver} with the execution results of tools. It provides the infrastructure necessary for tool execution, which can be either virtual or real. The former refers to a simulated environment that allows the model to interact with a digital representation of the tool, while a real environment involves actual interaction with the physical tool.
Virtual environments have the advantage of being easily accessible and replicable, allowing for more cost-effective training for models. However, virtual environments may not fully replicate the complexities of the real-world environment, leading to overfitting and poor generalization~\citep{hansen2020deployment}. On the other hand, real environments provide a more realistic context but may be more challenging to access and involve greater costs.

\paragraph{Controller.}
The controller $\mathcal{C}$ serves as the ``brain'' for tool learning framework and is typically modeled using a foundation model. The purpose of the controller $\mathcal{C}$ is to provide a feasible and precise plan for using tools to fulfill the user's request. To this end, $\mathcal{C}$ should understand user intent as well as the relationship between the intent and available tools, and then develop a plan to select the appropriate tools for tackling tasks, which will be discussed in~\cref{sec:aligning_user_tools}. In cases where the query is complex and targets a high-level task, $\mathcal{C}$ may need to decompose the task into multiple sub-tasks, which requires foundational models to have powerful planning and reasoning capabilities~(\cref{sec:framework_reasoning}).

\paragraph{Perceiver.}

The perceiver $\mathcal{P}$ is responsible for processing the user's and the environment's feedback and generating a summary for the controller. Simple forms of feedback processing include concatenating the user and environment feedback or formatting the feedback using a pre-defined template.
The summarized feedback is then passed to the controller to assist its decision-making. By observing this feedback, the controller can determine whether the generated plan is effective and whether there are anomalies during the execution that need to be addressed. Under more complex scenarios, the perceiver should be able to support multiple modalities, such as text, vision, and audio, to capture the diverse nature of feedback from the user and the environment.

\subsubsection{Connecting the Components}
\label{sec:framework_pillar_connect}

Formally, assume we have a tool set $\mathcal{T}$, which the controller can utilize to accomplish certain tasks. At time step $t$, environment $\mathcal{E}$ provides feedback $e_t$ on the tool execution. The perceiver $\mathcal{P}$ receives the user feedback $f_t$ and the environment feedback $e_t$, and generates summarized feedback $x_t$. Typically, the perceiver can be achieved by pre-defined rules (e.g., concatenating $f_t$ and $e_t$) to form $x_t$, or modeled with complex neural models.
The controller $\mathcal{C}$ generates a plan $a_t$, which selects and executes an appropriate tool from $\mathcal{T}$. This process can be formulated as the following probability distribution:
\begin{align}
    p_\mathcal{C}(a_t)=p_{\theta_\mathcal{C}}(a_t\mid x_t, \mathcal{H}_t, q),
    \label{eq:controller_pdf}
\end{align}
where $\theta_\mathcal{C}$ denotes the parameters of $\mathcal{C}$, $q$ denotes the user query or instruction, and $\mathcal{H}_t=\{(x_s, a_s)\}_{s=0}^{t-1}$ denotes the history feedback and plans. In its simplest form, a generated plan $a_t$ can simply be a specific action for tool execution. $\mathcal{C}$ can also synergize its reasoning process with the action prediction, where $a_t$ may additionally contain the reasoning traces that explain which sub-task should be solved next and which tool to choose for solving the sub-task. It is worth noting that if the dependence on $x_s$ is removed from Equation~(\ref{eq:controller_pdf}), the resulting probability distribution becomes equivalent to autoregressive language modeling. From this perspective, the controller additionally grounds the foundation model to the environment and the tool set. Moreover, we can factorize Equation~(\ref{eq:controller_pdf}) as follows:
\begin{align}
    p_{\theta_\mathcal{C}}(a_t\mid x_t, \mathcal{H}_t, q)=\sum_{\mathcal{T}_i \in \mathcal{T}} p_{\theta_\mathcal{C}}(a_t \mid \mathcal{T}_i, x_t, \mathcal{H}_t, q) \times p_{\theta_\mathcal{C}}(\mathcal{T}_i \mid x_t, \mathcal{H}_t, q),
    \label{eq:align_user_tool}
\end{align}
The decomposition reveals that the construction of the plan $a_t$ involves two subtasks: selecting the appropriate tool based on the user intent and deciding the actions to execute using the selected tool.
For instance, given an instruction such as ``I want to book a flight to Beijing next week'', the controller $\mathcal{C}$ first infers that the user's goal is to reserve a flight, with Beijing as the destination and the next week as the travel time. The model then selects the airline reservation system as the tool. Finally, it inputs the time and destination as the preliminary plan. In the process of making a reservation, we may face unexpected situations such as the unavailability of flights to Beijing in the next week. To cope with these anomalies, we can further equip $\mathcal{C}$ with the ability to reason about the current context and generate alternative plans, as we will discuss in detail in~\cref{sec:framework_reasoning}.

After a plan $a_t$ is generated, it will be executed in $\mathcal{E}$, and the resulting feedback $e_{t+1}$ from $\mathcal{E}$ will be passed on to the perceiver. 
The above process repeats for multiple rounds until the controller accomplishes the task. The overall objective is to find an action sequence $\{a_t\}$ that ultimately fulfills the task specified by the user instruction $q$. Note after tool execution, the controller may additionally integrate the execution results into a plausible response for the user (see details in \cref{sec:knowledge_conflicts}).
\subsection{The General Procedure: From Intent to Plan}
\label{sec:tool_learning_procedure}

As formulated in~\cref{sec:framework_pillar_connect}, the general procedure of tool learning necessitates intricate interplay among different components. In this section, we will further elaborate on the key issues involved in this procedure.

\subsubsection{Understanding Intent and Tools}
\label{sec:aligning_user_tools} 

To accurately fulfill the task specified by the user query $q$, the controller needs to understand two aspects: (1) the underlying intent of the user, which involves recognizing and formalizing the natural language $q$ as a high-level task (i.e., \textit{intent understanding}); (2) the tool set $\mathcal{T}$, which entails comprehending the functionality and objective of each tool within it (i.e., \textit{tool understanding}). By understanding both aspects, the controller can bridge the gap between the user’s intent and the tool set, which is the pre-requisite for connecting controller $\mathcal{C}$, the user, and tool set $\mathcal{T}$ in Figure~\ref{fig:formulation}.

\paragraph{Intent Understanding.}
Understanding user intent is a long-standing research topic in NLP~\citep{jansen2007determining,sukthankar2014plan}, which involves comprehending the underlying purpose of a user query. Intent understanding is essential in scenarios requiring human-computer interaction, such as developing advanced conversational agents capable of conducting intricate and nuanced dialogues with users. It requires learning a mapping from the instruction space to the model's cognition space. By accurately identifying the user intent, the controller can provide more personalized responses with a better user experience.

Recent explorations in instruction tuning~\citep{wei2021finetuned} demonstrate that foundation models can possess extraordinary proficiency in comprehending user instructions. Prior work has shown that fine-tuning large language models on a collection of datasets templated with human instructions allows models to generalize even to instructions for unseen tasks~\citep{wei2021finetuned,mishra2022cross,sanh2021multitask,bach-etal-2022-promptsource,ouyang2022training}. Promisingly, such generalization ability can further be enhanced by scaling up both the model size and the quantity or diversity of training instructions~\citep{iyer2022opt}. Despite the impressive intent understanding capabilities, challenges still exist in real-world tool learning scenarios:
(1) \textbf{Understanding Vague Instructions.} The first challenge is dealing with the inherent vagueness and ambiguity in the user query. Many user queries are inherently imprecise and can even be polysemous, requiring the controller to rely on contextual cues and background knowledge to infer the user's intended meaning. One possible solution is to actively interact with users to clarify any ambiguity, such as asking for clarifications about a previous user query.
(2) \textbf{Generalization to Diverse Instructions.} Another challenge is having the models generalize to more diverse user instructions. As the intent space is theoretically infinite, it is almost impractical for foundation models to be exposed to every real-world intention during training. In addition, the challenge of personalization arises from the fact that each individual has their own unique way of expressing intentions, which requires the model to adapt to the diverse expressions of intent of different individuals. One solution is to incorporate more diverse training data that covers a wide range of real-world scenarios, thereby enabling the models to learn the nuances of different instructions. Another solution is to leverage user feedback and actively adapt the model to individual users, i.e., personalized tool learning (\cref{sec:personalization}).

\paragraph{Tool Understanding.}
As noted by \citet{HERNIK200934}, when learning to utilize a specific tool, human beings perceive it as an object with particular functions, engaging in a cognitive process to understand its purpose and operation. By observing goal-directed demonstrations and following actions performed by other people, they gradually acquire the necessary knowledge and skills to use the tools effectively. Similarly, this understanding process is crucial for successfully solving tasks with tools. Analogously, a comprehensive understanding of the tools' functionalities is indispensable for enabling the controller to use tools proficiently. The process of tool understanding encompasses grasping what the tool is used for and how to use the tool. Take the case of a calculator: the controller needs to know that a calculator is intended for arithmetic operations, its input should be numbers and mathematical operators, and its output should be a computed value.

In real-world scenarios, tools are typically accompanied by a manual (or tutorial), which provides sufficient relevant details about their functionalities and usage. Endowed with strong few-shot learning~\citep{brown2020language} and zero-shot learning~\citep{wei2021finetuned} capabilities, foundation models can be prompted to unravel tools' functionalities and comprehend how to use them. To this end, we can construct suitable task-specific prompts either through manual design~\citep{vemprala2023chatgpt} or retrievial~\citep{zhou2023docprompting}. These prompts should describe the API functionalities or exemplify with demonstrations of their usage.

\begin{figure}[!t]
    \centering
    \includegraphics[width=\textwidth]{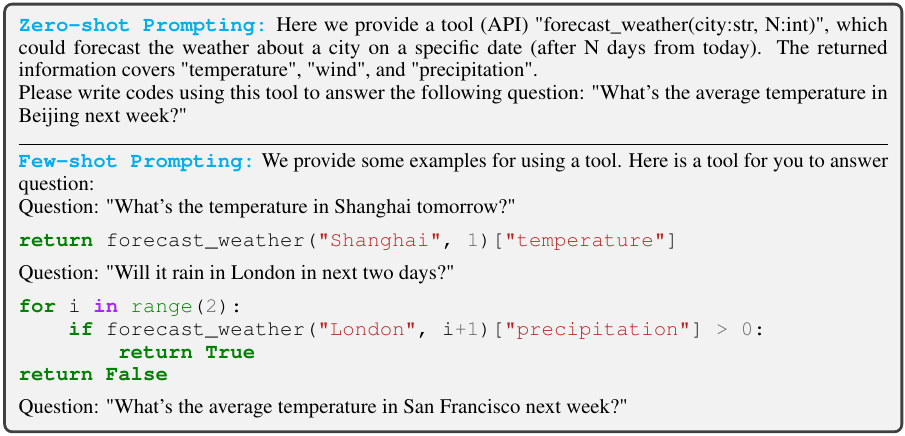}
    \caption{Examples of zero-shot and few-shot prompting for tool understanding. The prompts are constructed by describing the functionalities (zero-shot prompting) or giving usage examples (few-shot prompting) of a weather API.}
    \label{fig:tool_understanding_1}
\end{figure}

We categorize two prompting approaches as shown in Figure~\ref{fig:tool_understanding_1}: (1) \textbf{zero-shot prompting}, which describes API functionalities, their input/output formats, possible parameters, etc. This approach allows the model to understand the tasks that each API can tackle; (2) \textbf{few-shot prompting}, which provides concrete tool-use demonstrations to the model. By mimicking human behaviors from these demonstrations, the model can learn how to utilize these tools. We provide experimental results of both prompting methods in~\cref{sec:applications}.

Prompting has been widely adopted as a lightweight approach to teach foundation models about tools~\citep{yao2022react, driess2023palm, openai2023gpt4} with minimum human effort. Prompts can be easily adjusted to accommodate changes of tools. For instance, when tools are modified or upgraded, we can flexibly rewrite the prompts to adapt the model behaviors. Despite these advantages, prompting methods still face several challenges. First, since the effectiveness of prompting depends a lot on the model, smaller or less capable models cannot understand prompts well. Second, prompting is restricted by input context length. Although foundation models have been shown to learn to use simple tools through prompts, the situation may be more challenging with multiple complex tools with long descriptions. Especially when the tool set greatly expands, providing all possible tools within a prompt becomes infeasible given the limited context length.

A potential solution is to add an intermediate stage of tool selection, which first retrieves a small set of tools that are most suitable for the task at hand. Another solution is fine-tuning, which optimizes models with concrete tool-use examples to understand tools. This process involves leveraging the rich knowledge obtained from human tool-use experiences. When deployed in practice, a fine-tuned model alleviates the need for incorporating tool definitions in the input, which shrinks the input length and accelerates model inference. However, a major limitation of fine-tuning is that it often requires extensive human annotations or demonstrations. Additionally, tools are frequently updated, and fine-tuned models will need to be retrained with updated examples, incurring additional costs. More discussion about the tuning-based solution is left in \cref{sec:training_strategy}.

\subsubsection{Planning with Reasoning}
\label{sec:framework_reasoning}
As discussed in \cref{sec:aligning_user_tools}, understanding the intent and tools lays a solid foundation for planning. Nevertheless, it is still insufficient for tackling intricate tasks. The user query $q$ often implies a complex task that should be divided into multiple sub-tasks with proper sequencing, thereby necessitating a process of \textit{reasoning}.

Recent research has revealed that reasoning capabilities can emerge when foundation models are scaled up to a certain size~\citep{wei2022emergent}. In particular, foundation models with tens or hundreds of billions of parameters can generate intermediate reasoning traces during complex problem-solving, thereby significantly enhancing their zero-shot and few-shot performances~\citep{nakano2021webgpt,nye2021show,wei2022emergent}. The reasoning ability observed in the foundation models appears to facilitate a transition from System 1 to System 2~\citep{kahneman2011thinking}, enabling the accomplishment of more complex tasks.

\paragraph{Eliciting Reasoning in Foundation Models.}

\label{sec:framework_reasoning_enable_reasoning}
Despite the extensive study of the concept of \textit{reasoning} in the psychology literature~\citep{wason1968reasoning,kelley2013art}, the notion of \textit{reasoning} as applied to foundation models is not clearly defined. However, in general terms, the reasoning ability in the literature of foundation models can be framed as the capacity to decompose a complex problem into sub-problems and solve the sub-problems step-by-step~\citep{wei2022chain,press2022measuring,khot2022decomposed}. Here we keep consistent with these works and discuss reasoning in the sense of problem decomposition and sub-problem solving.

The vanilla few-shot prompt learning~\citep{brown2020language}, whereby models are provided with a prompt consisting of several examples for the given task, has been shown to fail when it comes to problems that require complex reasoning~\citep{creswell2022selection}. To address this issue, \citet{wei2022chain} propose Chain-of-Thought (CoT) prompting. Unlike vanilla few-shot prompt learning, CoT additionally inserts the reasoning trace required to derive the final answer for each example in the prompt. In this way, CoT prompts models to generate their ``thoughts'' on the necessary intermediate steps before arriving at the final answer. CoT has been proven to significantly boost performance on a wide range of tasks, including arithmetic reasoning, commonsense reasoning, and symbolic reasoning~\citep{wei2022chain}. 

\begin{figure}
    \centering
    \includegraphics[width=\linewidth]{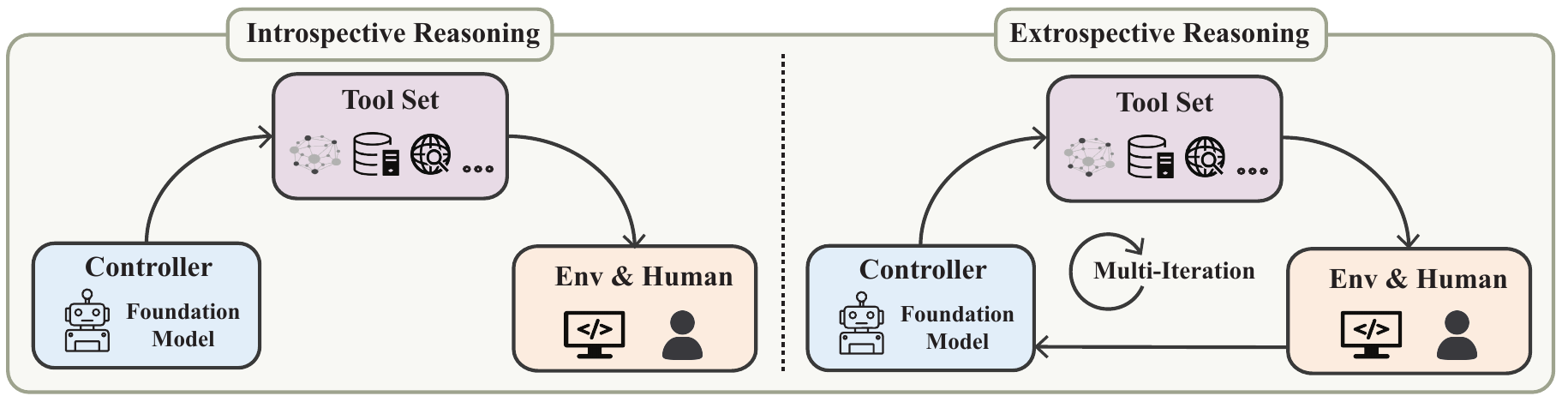}
    \caption{Illustration of introspective reasoning and extrospective reasoning. Extrospective reasoning requires feedback from the environment and humans to carry out iterative plan generation. We omit the perceiver in the illustration for simplicity.}
    \label{fig:reasoning}
\end{figure}

In light of the remarkable reasoning abilities of foundation models, recent research has made successful attempts to employ them in the controller in tool learning. It is demonstrated that their reasoning capabilities enable the controller to effectively decompose a complex problem into several sub-problems, and determine which tool to call upon for each sub-problem. We categorize relevant research into two streams: \textit{introspective reasoning} and \textit{extrospective reasoning}. The former involves generating a static plan of tool use without interacting with the environment $\mathcal{E}$, while the latter generates plans incrementally by iteratively interacting with $\mathcal{E}$ and utilizing feedback obtained from previous executions. As shown in Figure~\ref{fig:reasoning}, the environment $\mathcal{E}$ is invisible to the controller $\mathcal{C}$ in introspective reasoning but is visible in extrospective reasoning, creating a closed-loop interaction among the four components.

\paragraph{Introspective Reasoning.}
This kind of reasoning directly generates multi-step plans for tool use without knowing intermediate execution results. For instance, \citet{huang2022language} investigate the planning ability of foundation models and show that they are capable of decomposing high-level tasks into semantically plausible sub-plans. Another representative work is Program-Aided Language Models (PAL)~\citep{gao2022pal}, which prompts models to generate Python codes for intermediate reasoning steps. PAL uses the Python program interpreter as the tool, enabling the model to act as a programmer writing detailed comments, and achieving significant improvements in arithmetic, symbolic, and algorithmic reasoning. Notably, the idea of model-as-programmer has also been shown to be successful in embodied agents, as evidenced by ProgPrompt~\citep{singh2022progprompt} and Code-as-Policies~\citep{liang2022code}, which prompt models to generate executable programs for embodied agents. These studies reveal that, despite not having direct interaction with the environment, models are capable of generating executable programs for agents and anticipating possible anomalies in the plan execution. Another example is Visual ChatGPT~\citep{wu2023visual}, which interleaves various vision foundation models with ChatGPT to enable understanding and generating images. In their system, ChatGPT serves as the core controller and makes sequential decisions. At each step, ChatGPT might call a vision model to modify an existing image or respond to the user with plain text.

However, since these models are not grounded in the environment, they may generate unrealistic and nonsensical plans. To this end, SayCan~\citep{ahn2022can} emphasizes those actions that the agent is ``permitted'' to execute instead of those it is ``willing'' to perform. In practice, they employ a value function to estimate the probability of each action being successfully executed. With this function, the agent becomes more physically grounded in the environment. Overall, despite the absence of environment feedback, foundation models exhibit a remarkable ability to plan effectively in introspective reasoning. They can anticipate potential anomalies in plan execution and adjust their plans accordingly. This ability not only enables the controller to generate executable programs but also enhances its capacity to plan for a wide range of tasks.

\paragraph{Extrospective Reasoning.}
Despite its simplicity, introspective reasoning cannot adapt the plan in response to intermediate execution results. A more rational approach to planning is taking the environment $\mathcal{E}$ into account, and generating plans incrementally (e.g., one step at a time) with subsequent plans dependent on previous execution results. This allows the four components described in~\cref{sec:overview} to be well integrated and to cooperate effectively to achieve complex tasks. We refer to such an incremental reasoning strategy as extrospective reasoning. Compared to introspective reasoning, extrospective reasoning additionally considers feedback from the user and environment (Figure~\ref{fig:reasoning}), and is thus better suited to complex tasks~\citep{madaan2023self,wang2023describe}, such as multi-step QA and embodied learning, where decision-making at each step is dependent on the preceding context.

Recent works such as Self-Ask~\citep{press2022measuring}, ReAct~\citep{yao2022react}, and ToolFormer~\citep{schick2023toolformer} have demonstrated that by providing access to search engine APIs, models are able to achieve improved accuracy on multi-step QA. Through CoT prompting (Self-Ask and ReAct) or fine-tuning (ToolFormer), models can learn to decompose complex questions and utilize the search API to find the answer to the first sub-question. Based on the response and the question, they can then iteratively determine the subsequent question to ask or give the final answer. 

For embodied learning, while introspective reasoning methods have demonstrated the ability to generate executable programs and address potential execution anomalies, direct interaction with the environment can further enhance models' planning capabilities. For instance, Inner Monologue~\citep{huang2022inner} leverages multiple sources of feedback from the environment, such as whether a task is completed successfully and the current scene information. In this way, models can generate more feasible plans and improve their ability of high-level instruction completion. LLM-Planner~\citep{song2022llm} explicitly considers anomalies that may arise during plan execution and utilizes environmental feedback to regenerate the plan in case of execution failure, enabling models to handle exceptions appropriately. Additionally, ReAct~\citep{yao2022react} grants models the autonomy to determine when to cease generating action tokens during planning, enabling them to reason about the current situation and develop more refined subsequent plans.

In summary, extrospective reasoning requires interaction between the controller and the environment, which is a more complex setting. However, the real-time feedback from the user and environment empowers models to have a more comprehensive understanding of the current situation, making it possible to eventually achieve long-term goals that necessitate extensive planning.

\paragraph{Challenges in Multi-Step Multi-Tool Scenario.}
Humans do not stick to only one single tool to complete complex tasks. Instead, we carefully decompose the task into several sub-tasks, select the most suitable tool for each sub-task, and gradually accomplish them step by step. As discussed above, current research has shown satisfactory performance in task decomposition. However, there is a lack of exploration in utilizing different tools for different sub-tasks. Most of the research mentioned in this section is limited to either multi-step single-tool or single-step multi-tool scenarios. 
However, there has been a recent emergence of research that addresses the multi-step multi-tool scenario. One such example is the ReAct model~\citep{yao2022react}, which integrates multiple APIs of Wikipedia and employs the foundation model to decide when to use which API. Later, MM-ReAct~\citep{yang2023mm} generalizes ReAct to the multi-modal domain by including several vision experts. Furthermore, some recent projects such as Auto-GPT\footnote{\url{https://github.com/Torantulino/Auto-GPT}} demonstrate the huge potential of GPT-4 in manipulating multiple tools and making long-term plans, pushing the boundaries of what is possible with tool learning. Given a user query, Auto-GPT will take step-by-step actions to accomplish the objective autonomously. In addition to reasoning about the current state, Auto-GPT can also reflect on past actions to refine decision-making. Although these works constitute a significant step in advancing tool learning in the multi-step multi-tool scenario, there are still several challenges and future directions that need to be investigated.
\begin{itemize*}
    \item \textbf{Understanding the Interplay among Different Tools.} The multi-step multi-tool scenario typically involves complex tasks that require a higher level of intent understanding and reasoning capability. To effectively utilize multiple tools under this scenario, models need to grasp not only the individual functionalities of tools but also their interactions and dependencies. Models should be able to sequence the tools in a logical order so that the subsequent tools can leverage the information generated by the previous tools and effectively complete the task.
    
    \item \textbf{From Sequential Execution to Parallel Execution.} Tool executions do not have to be performed sequentially. In certain scenarios, parallel execution is possible for sub-tasks that do not depend on each other, which can potentially improve execution efficiency. For instance, given a user instruction \textit{“Generate two codes, one for drawing a rectangle, and one for drawing a circle.”}, the two tasks can be assigned to two agents, enabling the codes to be generated simultaneously. Determining the dependencies among different sub-tasks and effectively switching between parallel and sequential execution to optimize the overall process is a promising direction that merits further investigation.

    \item \textbf{From Single-agent Problem-Solving to Multi-agent Collaboration.} 
    Previous works typically assume that a single agent (controller) is solely responsible for the entire tool learning procedure. However, 
    in practice, complex tasks often demand collaboration among multiple agents, each possessing unique abilities and expertise. Embracing multi-agent collaboration can unlock more effective and efficient problem-solving approaches, necessitating the design of methods for communication, coordination, and negotiation among agents to ensure seamless collaboration and optimal task execution. Notably, recent work like \citet{park2023generative} demonstrates that multiple agents modeled with foundation models can simulate human behaviors (e.g., interpersonal communication) in interactive scenarios. This provides promising evidence for the adoption of multiple agents for tool learning.
\end{itemize*}

We look forward to more work in the future moving towards more practical multi-step multi-tool scenarios and making efforts to address these challenges. As a prior exploration, we evaluate foundation models' performance when multiple tools (APIs) are required to solve a task in \cref{sec:applications}.
\subsection{Training Models for Improved Tool Learning}
\label{sec:training_strategy}

\begin{figure}[!t]
    \centering
    \includegraphics[width=\linewidth]{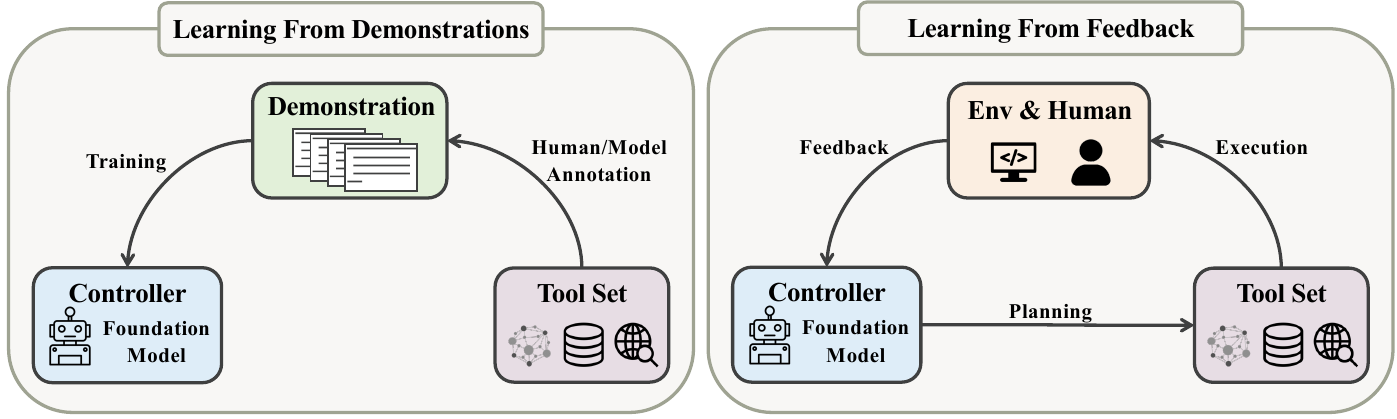}
    \caption{Training strategies for tool learning: (left) learning from human-annotated or model-annotated demonstrations; (right) learning from feedback, where the supervision could come from either the environment or humans.}
    \label{fig:training_strategy}
\end{figure}

Guidance, either from humans or environments, plays a critical role in training foundation models to use tools. In contrast to the prompting-based methods mentioned in \cref{sec:aligning_user_tools} and \cref{sec:framework_reasoning}, which rely on the frozen foundation models' in-context learning abilities, the training-based method optimizes the model with supervision. 
As noted by \citet{fagard2016does}, there are two primary ways for infants to learn a new tool, that is either from demonstration by an adult modeling the action or relying on their own exploration. Analogously, as shown in Figure~\ref{fig:training_strategy}, we categorize training strategies for tool learning into two streams: (1) learning from concrete tool-use demonstrations~\citep{nakano2021webgpt,sasaki2021behavioral}, which often requires human annotation, and (2) learning from feedback, which typically involves reinforcement learning~\citep{reddy2020sqil,baker2022video}. Finally, considering the existence of potentially massive tools, learning each of them one by one is infeasible in practice. Hence, we emphasize the importance of generalization in tool learning and discuss potential solutions (\cref{sec:generalizable_tool_learning}).

\subsubsection{Learning from Demonstrations} 
\label{sec:learning_from_demonstrations}

Models can be trained to mimic the behavior of human experts through imitation learning \citep{hussein2017imitation,liu2018imitation,baker2022video}. Behavior cloning~\citep{bain1995framework} can be viewed as a simplistic form of imitation learning that focuses on learning policies in a supervised fashion, with the general assumption that the expert's behavior is optimal or near-optimal.
The objective of behavioral cloning is to train models to imitate human experts' actions given certain inputs or conditions, and this approach is commonly adopted when the actions of an expert can be easily recorded and utilized for learning~\citep{torabi2018behavioral}. 

Formally, assume that we have a dataset $\mathcal{D}$ of size $N$ consisting of pairs of user query $q$ and the human demonstration annotation $a^*$, i.e., $\mathcal{D}=\{(q_i, a^*_i)\}_{i=0}^{N-1}$. Learning from human demonstrations optimizes the controller's parameters $\theta_\mathcal{C}$ with the following objective:
\begin{align}
    \theta_\mathcal{C}^*=\argmax_{\theta_\mathcal{C}}\E_{(q_i, a^*_i)\in\mathcal{D}}\prod_{t=0}^{T_i}p_{\theta_\mathcal{C}}(a^*_{i,t}\mid x_{i,t},\mathcal{H}_{i,t},q_i),
\end{align}
where $a^*_{i,t}$ is the human annotation at the $t$-th iteration for handling $q_i$, and $T_i$ is the total iteration number of $a_i$, other varaibles follow the notations defined in Equation~(\ref{eq:controller_pdf}). Based on how $a^*$ is obtained, learning from demonstration can be categorized into three streams, with human intervention gradually becoming less:

\paragraph{Supervised Learning.} 
Traditionally, behavior cloning has been widely explored in learning end-to-end or modular perceiver-controller models for autonomous vehicles and robotic applications~\citep{ly2020learning,codevilla2019exploring}. Recently, there has been a surge of interest in fine-tuning foundation models to perform tool-oriented tasks in a supervised way. For instance, \citet{li2022pre} utilize foundation models as policy networks, whose input is the tokenized environment observations, the original goals, and action history. Benefiting from the task-general inductive bias brought by foundation models, behavior cloning using the policy network significantly improves both in-domain performance and out-of-distribution generalization. Another example is WebGPT~\citep{nakano2021webgpt}, which interacts with a search engine by iteratively refining its search queries and recording important information. To achieve this, the authors first build a search interface backed up by Bing\footnote{\url{https://www.microsoft.com/en-us/bing/apis/bing-web-search-api}} and then fine-tune GPT-3~\citep{brown2020language} to clone human web search behaviors. As a language model pre-trained on general domains, the original GPT-3 is not intrinsically anchored to valid browser commands. 
Therefore, it is crucial to first gather demonstrations of human interactions with the browser and then learn state-to-action mappings. After fine-tuning, the model shows exceptional capabilities in manipulating search engines for information retrieval, even surpassing human experts. Similarly, WebShop~\citep{yao2022webshop} provides a web-based interactive environment where an agent could browse and purchase products. Through behavior cloning, the trained agent exhibits non-trivial performance in purchasing the right product given human instructions.

\paragraph{Semi-supervised Learning.}
As is often the case, human behaviors cannot be easily recorded due to time and cost considerations. However, large-scale unlabeled data is often attainable, from which we could potentially construct weak, noisy supervision. Notably, recent works have shown that we could employ a less-capable model to annotate pseudo-labels on unlabeled data and convert them into weakly-supervised tool-use demonstrations. For instance, with a small amount of seed labeled data, \cite{baker2022video} train a model to predict pseudo-labels of the action taken at each timestep in a Minecraft video game. 
Learning from these pseudo-labels, a more powerful model can be trained without requiring the rollout of models in a target environment or large-scale gold-standard human behavior annotation. 

\paragraph{Self-supervised Learning.}
Despite reducing the heavy requirements on human behavior annotation, semi-supervised learning still requires a seed labeled dataset to attain the pseudo labels. Besides, the biases in the seed dataset may also be amplified during training, leading to poor generalization performance. To this end, researchers recently show that with a few demonstrations, foundation models can teach themselves how to utilize a tool in a self-supervised manner~\citep{parisi2022talm,schick2023toolformer}. For instance, Toolformer~\citep{schick2023toolformer} leverages the in-context learning ability of foundation models to iteratively bootstrap tool-use examples based on a handful of human-written examples. These auto-generated examples are further filtered to reduce noise. The final tool-use dataset contains sufficient supervision, significantly improving GPT-J's~\citep{gpt-j} tool-use performance, highlighting the potential of self-supervised learning for enhancing tool-use capabilities.

\subsubsection{Learning from Feedback}
\label{sec:learning_from_feedback}
Collecting manually annotated tool-use examples, which probably include complete traces of human behaviors and the final answers, is time-consuming and labor-intensive. Moreover, the learned model may not adapt effectively to new environments as it conforms to the recorded human behaviors. Besides, it is impractical to explicitly annotate every possible scenario of environment condition and agent behavior~\citep{codevilla2019exploring}. Alternatively, humans learn from trial and error to correct and rectify their tool-use behaviors~\citep{allen2019rapid}. Similarly, feedback from both the environment and humans can enable the model to understand the consequences of its actions and adapt its behaviors. The supervision from feedback can also enhance the capabilities of an agent trained in a supervised way~\citep{nakano2021webgpt,baker2022video}. Formally, learning from feedback can be described as optimizing the controllers' parameters $\theta_{\mathcal{C}}$ from open explorations with query set $Q=\{q_i\}_i$:
\begin{align}
    \theta_\mathcal{C}^*=\argmax_{\theta_\mathcal{C}}\E_{q_i \in Q } \E_{\{a_{i,t}\}_{t=0}^{T_i} \in p_{\theta_\mathcal{C}} } \left[ R({\{a_{i,t}\}}_{t=0}^{T_i}) \right],
\end{align}
where $R$ is the reward estimated from the sequence of feedback and $T_i$ denotes the number of iterations needed for handling $q_i$.

\paragraph{Reinforcement Learning (RL) for Tool Learning.}
RL is a common solution to enabling artificial agents to learn from their environment in complex decision-making processes~\citep{silver2018general,berner2019dota,schrittwieser2020mastering}. Tool learning can be considered an RL scenario, where the action space is defined by tools, and the agent learns to select the appropriate tool and perform the correct actions that maximize the reward signal. The policy model can be initialized by a foundation model~\citep{schulman2017proximal}.
Such initialization brings the policy model abundant prior knowledge, alleviating the need for the RL agent to learn basic skills. With a reward function that quantifies the performance of the agent in achieving the task goal, RL has been successfully used in various tool learning scenarios, such as robotic grasping~\citep{levine2018learning} and multi-agent autocurricula~\citep{baker2019emergent}. By optimizing the loss function, the agent learns to reflect on the current state of the environment, select the appropriate tool, and perform the right actions that lead to the highest expected reward.
In the following, we introduce two sources of feedback: environment feedback and human feedback, which can be considered sources of reward signals in the context of tool learning. These two feedbacks are complementary and can be combined with each other. 

\paragraph{Environment Feedback.}
The controller interacts with the environment and receives feedback about the consequences of its actions. The model then updates its policy based on this feedback to improve its tool-use behavior. Environment feedback can be categorized into two forms: (1) \textbf{result feedback}, which is ultimate feedback returned from the environment, indicating whether the model's actions have successfully completed the task or not. This type of feedback performs an overall assessment of the planning generated by the model. For instance, WebShop~\citep{yao2022webshop} uses a hand-coded reward to assess the similarity between human-bought and model-bought products, which indicates whether the actions performed by the controller lead to the correct final product. By receiving feedback on the success or failure of its actions, the model can iteratively update its planning strategy, and adjust its decision-making process; (2) \textbf{intermediate feedback}, which refers to the state change of the environment triggered by an action. By observing the state changes, foundation models can learn whether each action is effective and appropriate, making the model better adjust its behaviors accordingly. This kind of feedback provides more detailed and timely information about the effectiveness of each tool execution. Take the case of interacting with a search engine to gather information for question-answering, models could update their policy for more efficient information retrieval by observing the rendered information of a search query.

\paragraph{Human Feedback.}
Humans could give the model rewards and penalties based on its generated plans to regulate its behavior. Human feedback can be \textbf{explicit}, which provides clear and direct insights into the model performance representing human preferences. For example, rating the quality of the model-generated action on a scale of $1$ to $5$; human feedback can also be \textbf{implicit}, which is not directly specified by the user but can be derived from user behavior and interactions with the model. Examples include users' comparison~\citep{ouyang2022training}, response time, and actions taken after receiving a model's output (e.g., clicking on a recommended link).

Though human feedback is accurate and stable, it is label-intensive and has high latency. To address this issue, reinforcement learning from human feedback (RLHF)~\citep{NIPS2017_d5e2c0ad} is proposed to finetune a model to imitate humans to give rewards, which are then used to optimize the policy with RL algorithms such as PPO~\citep{schulman2017proximal}. RLHF has yielded exceptional performance in various domains such as text summarization~\citep{ziegler2019fine,stiennon2020learning}. RLHF can also improve a model's tool-use capabilities even if it has been trained on sufficient supervised human demonstrations. For instance, WebGPT~\citep{nakano2021webgpt} utilizes human feedback to guide a policy model to align with human preferences, which helps better manipulate search engines to answer long-form questions.

Despite its remarkable performance, RLHF still faces challenges: (1) \textbf{task-specific nature}: the corresponding evaluation criteria for specific tasks need to be pre-defined, and the preference data annotated for one task is hard to be transferred to other settings, which limits the applicability of RLHF to a wider range of tasks. To this end, it is critical to develop a universal reward model that generalizes to various tasks; (2) \textbf{biases}: RL agents optimize towards the pseudo-human reward model, thus can be up-bounded and biased by human preferences. Besides, societal biases or personal experiences may be amplified during RLHF, and it is essential to carefully evaluate the learned reward model for any biases and take measures to mitigate them.

\subsubsection{Generalizable Tool Learning}
\label{sec:generalizable_tool_learning}

Generalization of tool use is a key characteristic of human intelligence~\citep{seed2010animal,teschke2013did,osiurak2018tool}. The ancient human, for instance, recognized that regardless of the specific tool being used, a sharp edge was essential for achieving clean cuts and efficiently carrying out tasks. This recognition allowed them to transfer their knowledge of sharpening a knife to sharpening other tools, such as scrapers or choppers. Generalization is also a critical aspect of tool learning, especially considering the existence of a massive and rapidly expanding array of tools. Although conducting supervised fine-tuning on a vast collection of tool-use data can be a potential solution to facilitating generalization, collecting enough supervised tool-use data and ensuring its quality and diversity is time-consuming and practically infeasible.

Generalizable tool learning highlights the importance of \textit{abstraction}, which is the process of identifying the essential features of a tool. Abstraction involves recognizing commonalities and patterns of tools so that models could synthesize and transfer their knowledge and skills, enabling them to use novel tools with ease. For instance, by abstracting essential features such as layers, filters, and color adjustments, users can transfer their knowledge of using Adobe Photoshop to Adobe Illustrator, even if the interface and specific tool names in these two figure-editing software are different. Abstracting these general features of tools can quickly help users learn a new tool effectively by building on previous experience.

\paragraph{Foundation of Generalization: Interface Unification.}
To facilitate knowledge transfer among tools, it is critical to design a unified interface that enables the model to manipulate various tools in a consistent and standardized manner, which serves as the foundation for generalizable tool learning. Through a unified interface, models can identify and abstract essential features of tools more easily in a unified tool protocol rather than grappling with the difficulty of understanding various tool interfaces. Currently, the manipulation of tools is through predicting discrete action tokens, and the action space is not aligned in different scenarios, which prohibits the models from quickly adapting to new scenarios and tools. Inspired by the aspect we categorize tools in \cref{sec:tool_categorization}, we identify three potential ways of interface unification: the semantic interface, GUI interface, and programming interface.

\begin{itemize*}
  \item \textbf{Semantic Interface.} The semantic interface operates by utilizing a specific text span (action name) as the action trigger, which is the most intuitive and natural way for interface unification. For instance, ReAct~\citep{yao2022react} employs \texttt{Action:Search} as the trigger for the function that searches for relevant passages. In robotic manipulation~\citep{ahn2022can, liu2023lang2ltl}, the generated natural language (e.g., \texttt{pick up the sponge}) is mapped to specific actions. Despite its ease of implementation, the semantic interface poses certain challenges that must be addressed. First, the mapping between the generated text and the corresponding tool action should be pre-defined individually, which is a laborious task, particularly when the tool set expands quickly. Moreover, the model may fail to accurately produce the precise form to trigger the intended action, even leading to false triggering of actions.

  \item \textbf{GUI Interface.} Humans primarily interact with the digital world through GUI interface (e.g., mouse and keyboard), which has been extensively optimized to follow human action efficiently. Nevertheless, before robots can learn to use a GUI interface flexibly, it is necessary to establish a virtual environment that can facilitate mapping predicted tokens to human-like mouse movements and keyboard inputs. Prior research has explored providing platforms for agents to complete web-based tasks using keyboard and mouse actions~\citep{pmlr-v70-shi17a, liu2018reinforcement}. However, these environments restrict models to a limited set of pre-defined mouse options and common keyboard actions such as copy and paste. By leveraging foundation models, it is possible to introduce prior knowledge regarding common combinations of keyword and mouse actions, thereby expanding the potential actions that a model can execute.

  \item \textbf{Programming Interface.} This kind of interface allows the model to go beyond pure natural language and specify its action using a program. Such unification requires the model to be acquainted with the syntax of the function calls. The recent code-generating language models (CLM) such as Incoder~\citep{fried2022incoder} and CodeX~\citep{chen2021evaluating} provide the possibility of such unification. The programming interface has been applied widely. For example, Code-as-Policies~\citep{liang2022code} finds that with CLM as the backbone for robotic control, the robots can leverage the code grammar to execute complex actions, generalize to novel instructions, and give precise control with accurate parameter values to the functions. The programming interface provides promising opportunities for tool learning because (1) complex tool learning logic can be modeled using the control flow of programming language; (2) explicit calls of external APIs can be naturally implemented by executing programs.
\end{itemize*}

It should be noted that the interface selection should align with the capabilities and limitations of the foundation model. For instance, language foundation models are trained to generate text and may be better suited for the semantic interface. Similarly, a multimodal foundation model that combines visual and textual information may be more appropriate for the GUI interface, as it can understand and generate human-like mouse movements and keyboard inputs. On the other hand, code foundation models may be more suitable for the programming interface, as it is trained to understand code syntax and function calls.

Under certain cases, we may face challenges where the tool's output is not aligned with model's input format. A common practice is to compose the functionality of the model and tool in the same modality. For example, \citet{zeng2022socratic} chain together foundation models of various modalities by converting their outputs into natural languages. This simple method leverages prompting to compose new multimodal capabilities without fine-tuning. In contrast, another solution is to building multimodal foundation models that can perceive general modalities, based on the belief that multimodal foundation models can all be unified through a general-purpose interface~\citep{alayrac2022flamingo,hao2022language}. Gato~\citep{reed2022generalist} is a representative generalist multi-embodiment agent trained on tremendous datasets of agent experience. Gato can sense and act with different embodiments, such as playing Atari, captioning images, chatting, etc. Similarly, PaLM-E~\citep{driess2023palm} incorporates continuous inputs from different modalities into a PLM. By joint training on multiple embodied tasks, PaLM-E could make grounded decisions in the real world.

\paragraph{Strategies of Generalizable Tool Learning.}
In general, a unified interface enables models to learn and transfer knowledge more easily and efficiently, but it does not guarantee optimal learning outcomes in all scenarios. Generalizable tool learning requires models to further adapt, refine, and specialize their learned knowledge to specific tasks or domains. Here, we discuss two potential approaches to achieving this goal and facilitating generalization.

\begin{itemize*}
  \item \textbf{Meta Tool Learning.} Metacognition~\citep{metacog_tooluse} is a crucial aspect of human intelligence that allows individuals to reflect on their own thinking and adapt their behaviors when faced with unfamiliar situations. In the context of tool learning, metacognition refers to the ability of a model to reflect on its own learning process and adapt new tool-use strategies when necessary. With metacognition, models can identify common underlying principles or patterns in tool-use strategies and transfer them to new tasks or domains. Take the case of the web search tool, when the model trained on a source search engine (e.g., Bing Search) is transferred to a target one (e.g., Google Search), the model can leverage its metacognitive awareness to adapt its tool-use strategies based on its previous experiences. This may include identifying common underlying patterns in tool-use strategies, such as effective search queries, relevant results, and user feedback, and using this metacognitive awareness to better align with the algorithms and user interface of the new search engine. 

  \item \textbf{Curriculum Tool Learning.} Another approach to improving model generalization is through curriculum learning~\citep{bengio2009curriculum}, which starts with simple tools and gradually introduces the model to more complex tools so that it can build upon its prior knowledge and develop a deeper understanding of the tool. For instance, we could start with a curriculum of basic algorithms and operations to effectively teach a model to use Mathematica\footnote{\url{https://www.wolfram.com/mathematica}}, e.g., addition and subtraction, and then gradually move on to more complex mathematical concepts like calculus and linear algebra. This training strategy ensures that the model is introduced to the simple, essential features of the tool before moving on to more complex concepts in a way that is manageable and effective. Moreover, curriculum tool learning allows the model to learn how complex tools are built upon simple tools. It provides an understanding of how a complex tool can be seen as an updated high-level version of a simple tool, and how its function is a combination of several basic tools. This understanding of the relationship between simple and complex tools facilitates the transfer of previously learned knowledge to new tools, enabling the model to more effectively identify similarities and differences between situations and adjust its approach accordingly.
   
\end{itemize*}
\section{Application and Experiment}
\label{sec:applications}

In this section, we aim to explore the applications of tool learning and investigate the efficacy and limitations of state-of-the-art foundation models in utilizing tools. We select $18$ representative tools for evaluation and place the main results in this section. For more case studies of ChatGPT, please refer to Appendix~\ref{sec:case_study}.

\subsection{Evaluated Tools}
We first briefly introduce the tools selected in experiments as follows:

\textbf{Machine Translator.}
General-purpose language models may exhibit suboptimal proficiency when processing text from multiple linguistic domains. Machine translators can effectively alleviate this issue by enabling non-translation-dedicated language models to better comprehend multi-lingual texts.
Following Toolformer, we use NLLB~\citep{costa2022no} as our translator and choose MLQA~\citep{lewis2019mlqa}, a multilingual question answering benchmark, as the testbed. Given a context in English and a question in Arabian, the task requires answering the question using English. We randomly sample $200$ test instances from the original test data. For the evaluation metric, we choose F1-score.

\textbf{Calculator.}
Following the setting of Toolformer, we conduct experiments in which language models use a simple calculator to solve math word problems. 
We choose a simple implementation for the calculator, which supports basic arithmetic operations (i.e., $+$, $-$, $\times$, $\div$). We evaluate two math word problem datasets: ASDiv~\citep{miao2020diverse} and MathQA~\citep{amini-etal-2019-mathqa} and choose accuracy as the metric.

\textbf{Map.}
We choose Bing Map API\footnote{\url{https://learn.microsoft.com/en-us/bingmaps}} for location information retrieval, assisting in user queries related to the route, driving distance, latitude coordinates, nearby locations of interest, etc. We manually curate user queries through crowdsourcing.

\textbf{Weather.}
We choose Weather API\footnote{\url{https://www.weatherapi.com}} and investigate whether models could use the tool to answer weather-related questions, such as questions about current weather in any city, forecasting the weather within two weeks in any city, and giving suggestions based on the weather information. Two APIs are supported, the first one is \texttt{GetWeatherToday<city>}, which provides the current weather condition of a city; another one is \texttt{ForecastWeather<city, N>}, which forecasts the weather of a city after N days. The detailed information returned includes the temperature, wind speed, UV index, sunrise, sunset time, etc. We manually curated $100$ weather-related user queries.

\textbf{Stock.}
We choose Alpha Vantage Stock API\footnote{\url{https://www.alphavantage.co/documentation}} for stock market querying. We aim to obtain specific information about the opening, closing, highest, or lowest price for one particular stock on one specific day or month. We manually curate $1200$ question-answer pairs and choose accuracy as the evaluation metric.

\textbf{Slides.}
Slides-making is traditionally performed by humans using a human-computer interface (e.g., mouse and keyboard). However, current models cannot directly move a mouse or press computer keys.
To address this limitation, we provide six APIs with high-level semantics for the model. Four APIs are built based on the open-source library python-pptx\footnote{\url{https://pypi.org/project/python-pptx}}  to control the slides, one API allows the model to retrieve images from the internet based on a topic, and one API is used to submit and display the final slides to the user. 
To collect the data, we brainstorm $50$ different careers that require slides-making in their work, for each career, we brainstorm $2$ cases where practitioners have to make slides. The final dataset consists of $100$ slides-making tasks. We evaluate the model's performance by counting the fraction of instances in which the model-generated API calls are correctly executed without errors.

\textbf{Table Processing.}
We craft a suite of table processing APIs using pandas.DataFrame in Python.
By leveraging these tools, models can provide a more natural and streamlined experience for users, allowing them to perform data analysis and visualization tasks directly.
We manually construct a table processing dataset containing $13$ tables and $117$ corresponding queries. 

\textbf{Knowledge Graphs.}
Knowledge graphs contain factual knowledge about the real world, which is stored in the form of RDF triplets. The triplets can be retrieved by SPARQL (Standard Protocol and RDF Query Language). We provide $7$ APIs that mimic the process of human querying the knowledge graph, including showing the candidate entity/relation given a name surface form, showing a head entity's home page, showing a tail entity's home page, sending SPARQL queries, showing the result of SPARQL queries, and finding a keyword in the output of a SPARQL query. We curate $64$ questions that could be answered by querying knowledge graphs.

\textbf{Search Engine.}
We choose Bing Search API\footnote{\url{https://www.microsoft.com/en-us/bing/apis/bing-web-search-api}} and test the model on real-time question answering. Two APIs are supported: the first one is \texttt{Search<query>}, which returns the top-related search results back to the model; another one is \texttt{LoadPage<N>}, which loads the detailed information of page N indexed in the search results, and returns the detailed contents. We experiment with RealTimeQA~\citep{kasai2022realtimeqa}, which is a dynamic question-answering platform that inquires about novel events or information. Specifically, we choose the most recent release (20230217 version) of multiple-choice data for evaluation. Given the question and choices, the model is expected to interact with the search engine to extract the necessary information, before settling on the final answer.

\textbf{Wikipedia.}
We largely build our Wikipedia Search tool upon ReAct~\citep{yao2022react} with slight modifications on the API designs. The tool consists of $3$ APIs: \texttt{search<entity>}, which searches for an exact entity name on Wikipedia and returns the first $5$ sentences of the corresponding page if the entity exists; otherwise, it displays related entity names; \texttt{lookup<keyword>}, which looks up the keyword on the current page and returns the next sentence containing the keyword, similar to humans' using the \texttt{CTRL+F} function on a web page; \texttt{disambiguate<entity>}, which inputs an entity name and displays all entities that share the same name. We focus on HotpotQA~\citep{yang-etal-2018-hotpotqa} for question answering. We conduct our experiments in an open-domain setting, where only the question is shown to the model. We randomly sample $200$ instances from the dataset.

\textbf{Online Shopping.}
Amazon online shopping is a relatively complex web environment, in which models need to buy a commodity that satisfies various requirements mentioned in a user instruction.
Based on WebShop~\citep{yao2022webshop}, we build our online shopping tool, which covers mainstream online shopping actions including searching for an item, loading detailed information about an item, choosing a feature for an item, going to the previous/next page, deciding to purchase, etc. We use the dataset provided by WebShop and randomly sample $100$ test instances, which cover instructions about various customers' needs with specific requirements of commodities' attributes.

\textbf{Embodied Scene.}
ALFWorld~\citep{shridhar2020alfworld} is an aligned text and embodied environment game, where agents need to interact with objects (e.g., fridge, microwave, drawer, etc.) in a house to complete a task (e.g., putting a clean spatula in a drawer). 
We largely follow the setting of ReACT~\citep{yao2022react} and report the success rate on the valid set.

\definecolor{LightCyan}{rgb}{0.83,0.95,0.95}
\newcolumntype{B}{>{\columncolor[rgb]{0.83,0.95,0.95}}}
\begin{table*}[!t]
  \centering
    \begin{tabular}{l c c c BcBcBc}
    \toprule
\rowcolor{white}
    \textbf{Tools}  &  \textbf{\# APIs} & \textbf{Test Set} & \textbf{Test Size} &  \textbf{No Tool} & \textbf{Zero-shot} & \textbf{Few-shot}   \\
    \midrule
\rowcolor{white}
    \multirow{2}{*}{Machine Translator}  & \multirow{2}{*}{1}  & \multirow{2}{*}{MLQA} &  \multirow{2}{*}{200} & 49.1 & 49.7 & 54.0 \\
    &&&& 38.2 & 38.6 & 45.5 \\
    \midrule
\rowcolor{white}
    \multirow{2}{*}{Calculator} & \multirow{2}{*}{1} & \multirow{2}{*}{ASDiv}  & \multirow{2}{*}{266} & 85.3 & 81.6 & 92.5   \\
    &&&&91.7& 74.1 & 92.5  \\
    \midrule
\rowcolor{white}
    \multirow{2}{*}{Map} & \multirow{2}{*}{11} & \multirow{2}{*}{Curated}& \multirow{2}{*}{129}& \sout{\phantom{00.0}} & 58.1 & 86.8 \\
    &&&& \sout{\phantom{00.0}}&  29.7 & 86.8\\
    \midrule
\rowcolor{white}
    \multirow{2}{*}{Weather}  & \multirow{2}{*}{2} & \multirow{2}{*}{Curated} & \multirow{2}{*}{100} &\sout{\phantom{00.0}}  & 39.0 & 99.0 \\
    &&&& \sout{\phantom{00.0}} &   92.0 & 99.0   \\
    \midrule
\rowcolor{white}
    \multirow{2}{*}{Stock} & \multirow{2}{*}{8} &  \multirow{2}{*}{Curated} & \multirow{2}{*}{122} &\sout{\phantom{00.0}}& 33.6& 63.1 \\
    &&&&\sout{\phantom{00.0}}&  39.0 & 64.8 \\
    \midrule
\rowcolor{white}
    \multirow{2}{*}{Slides} & \multirow{2}{*}{6}  & \multirow{2}{*}{Curated} & \multirow{2}{*}{100} &\sout{\phantom{00.0}} & 95.0 & 97.0 \\
    &&&&\sout{\phantom{00.0}}& 94.0 & 86.0  \\
    \midrule
\rowcolor{white}
    \multirow{2}{*}{Tables} &  \multirow{2}{*}{21} & \multirow{2}{*}{Curated} & \multirow{2}{*}{117} &  54.8 & 60.7 & 85.2 \\
    &&&& 60.9 & 73.0 & 92.2  \\
    \midrule
\rowcolor{white}
    \multirow{2}{*}{KGs} & \multirow{2}{*}{7} & \multirow{2}{*}{Curated} & \multirow{2}{*}{64} & \sout{\phantom{00.0}}& 42.2 & 46.9 \\
     &&&&\sout{\phantom{00.0}}&  \ \ 7.8 & 14.1 \\
     \midrule
\rowcolor{white}
    \multirow{2}{*}{Search Engine} & \multirow{2}{*}{2} & \multirow{2}{*}{RealTimeQA} & \multirow{2}{*}{30} & 50.0 & 50.0 & 66.7 \\
    &&&&50.0&  43.3 & 63.3 \\
    \midrule
\rowcolor{white}
    \multirow{2}{*}{Wikipedia} & \multirow{2}{*}{3}  & \multirow{2}{*}{HotpotQA} & \multirow{2}{*}{200} & 33.5 & 28.5 & 35.5 \\
    &&&& 34.5 & \ \ 8.5  & 19.0   \\
    \midrule
\rowcolor{white}
    \multirow{2}{*}{Online Shopping} & \multirow{2}{*}{2}  & \multirow{2}{*}{Webshop}  &  \multirow{2}{*}{100} & \sout{\phantom{00.0}} & 38.4 & 37.1 \\
    &&&&\sout{\phantom{00.0}}&    42.0 & 35.9 \\
    \midrule
\rowcolor{white}
    \multirow{2}{*}{Embodied Scene} & \multirow{2}{*}{1} & \multirow{2}{*}{ALFWorld} & \multirow{2}{*}{134}&\sout{\phantom{00.0}} & 51.0  & 78.0 \\
    &&&&\sout{\phantom{00.0}}&   23.0& 81.0\\
    \midrule
\rowcolor{white}
    \multirow{2}{*}{Cooking Assistant} & \multirow{2}{*}{3} & \multirow{2}{*}{Curated} & \multirow{2}{*}{50} & \sout{\phantom{00.0}}&  84.0 & 98.0 \\
    &&&&\sout{\phantom{00.0}}&   82.0& 90.0 \\
    \midrule
\rowcolor{white}
    \multirow{2}{*}{Movie Search} & \multirow{2}{*}{3} & \multirow{2}{*}{Curated} & \multirow{2}{*}{60} &\sout{\phantom{00.0}} &  77.0 & 72.0 \\
    &&&&\sout{\phantom{00.0}}& 43.0& 75.0\\
    \midrule
\rowcolor{white}
    \multirow{2}{*}{AI Painting}  & \multirow{2}{*}{2} & \multirow{2}{*}{Curated} & \multirow{2}{*}{25} &\sout{\phantom{00.0}} & 93.0 & 100.0\\
    &&&&\sout{\phantom{00.0}}& 90.0& 88.0 \\
    \midrule
\rowcolor{white}
    \multirow{2}{*}{3D Model Construction} & \multirow{2}{*}{14} & \multirow{2}{*}{Curated} & \multirow{2}{*}{10} & \sout{\phantom{00.0}}& 20.0& 40.0 \\
    &&&&\sout{\phantom{00.0}}& \ \ 0.0& 40.0 \\
    \midrule
\rowcolor{white}
    \multirow{2}{*}{Chemical Properties} & \multirow{2}{*}{4} & \multirow{2}{*}{Curated} & \multirow{2}{*}{100} &35.0 & 55.5 & 73.5 \\
    &&&& 46.5 & 67.0& 81.0 \\
    \midrule
\rowcolor{white}
    \multirow{2}{*}{Database} & \multirow{2}{*}{4} & \multirow{2}{*}{Curated} & \multirow{2}{*}{50} & \sout{\phantom{00.0}} & 50.0 & 75.0 \\
    &&&& \sout{\phantom{00.0}} & 58.3 & 75.0 \\    
    \bottomrule
    \end{tabular}%
    \caption{\label{tab:overall_result} We list the overall results of different tools evaluated in this paper. ``\# APIs'' denotes the number of APIs corresponding to each tool. The test set means the dataset we employed in conducting the experiments. We show the result of three settings i.e., \textbf{No Tool}, \textbf{Zero-shot}, \textbf{Few-shot}. The results of text-davinci-003 are shown on white background, while the results of ChatGPT are shown in \colorbox{LightCyan}{cyan} background.
}
\end{table*}

\textbf{Cooking Assistant.}
We choose AllRecipe\footnote{\url{https://www.allrecipes.com/}} to investigate whether models can find the proper cooking recipe and extract important details. The tool is designed similarly to the search engine tool. With this tool, the model can perform: (1) finding the target recipe, and (2) answering questions based on observed details. We manually curate $50$ queries for evaluation.

\textbf{Movie Search.}
We choose Douban Film API\footnote{\url{https://movie.douban.com}} to search for movie-related information. Three APIs are devised with the aim of discovering movies that are currently playing or upcoming, as well as extracting detailed information about each movie. We curate $60$ questions about the movies, such as recommending some movies which are on display or upcoming and providing a brief introduction to a movie.

\textbf{AI Painting.}
 AI image generation model has been widely used by human artists. To endow models with the capacity to create images using the AI image generation model, we provide the following APIs: one API generates an image given a prompt using stable diffusion~\citep{Rombach_2022_CVPR}, others are the image segmentation~\footnote{\url{https://huggingface.co/CIDAS/clipseg-rd64-refined}} and image inpainting APIs~\footnote{\url{https://huggingface.co/runwayml/stable-diffusion-inpainting}}, which replace a target object in an image with a new object described by a prompt. We curate $25$ queries as the initial prompt, together with subsequent queries for modifying that image.

\textbf{3D Model Construction.}
We investigate three-dimensional (3D) modeling by manually devising a collection of APIs that leverage the capabilities of the sophisticated 3D rendering engine Taichi\footnote{\url{https://github.com/taichi-dev/voxel-challenge}}. Due to the complexity of executing this API (3D rendering), we only demonstrate the performance on $10$ curated questions.

\textbf{Chemical Properties.} To evaluate the capability of tool learning in professional domains, we utilize the Chemical Property query, and more specifically, the PubChem\footnote{\url{https://pubchem.ncbi.nlm.nih.gov}} API for resolving scientific inquiries. $4$ APIs are supported, which facilitate the retrieval of a chemical's identification number based on the name or SMILES notation~\citep{weininger1988smiles}, as well as obtaining the chemical's properties based on its identification number. We manually curated $100$ questions for evaluation.

\textbf{Database.} We explore the potential of tool learning in accessing database data via natural language. The fundamental APIs include $(i)$ obtaining the structural information of the target data (schema); $(ii)$ translating the input natural language text into an equivalent SQL query; $(iii)$ rewriting the query into an execution-efficient one; and $(iv)$ querying the result data by connecting to the database. We automatically curated 50 relatively complex queries under the TPC-H schema, which are of relatively complex structures (involving 2-6 tables and composite predicates) and take over 111.5 minutes to execute in total. The evaluation metric is the ratio of queries for which tool learning can accurately output the results.

To facilitate future research attempts, we implement and integrate all the above tools into \textbf{BMTools}\footnote{\url{https://github.com/OpenBMB/BMTools}}, which is an open-source repository that extends foundation models using tools and also serves as a platform for the community to build and share tools. With BMTools, users can easily build a new plugin by writing Python functions and also integrating external tools from other sources (e.g., ChatGPT Plugins).

Building a \textit{tool library} for foundation models is critical to connecting foundation models with tools and we are glad to see there are emerging works in this direction. LangChain~\footnote{\url{https://docs.langchain.com}} is the first open-sourced project that attempts to chain foundation models with tools. Under a unified interface, users could either build their own task pipelines or let the foundation models call APIs. 
Most recently, TaskMatrix.AI~\citep{liang2023taskmatrix} and HuggingGPT~\citep{shen2023hugginggpt} extend APIs and tasks to broader scenarios, including multimodal models for visual tasks, local software, and cloud service APIs. OpenAI also proposed its official tool library, ChatGPT Plugins~\footnote{\url{https://openai.com/blog/chatgpt-plugins}}, to empower ChatGPT with other applications. By simply providing APIs with descriptions, ChatGPT is enabled to call applications and complete more complex tasks. Different from third-party libraries, ChatGPT plugins are cautious about safety risks and establish strict standards for plugins. The library prioritizes the most essential tools such as the web browser, code interpreter, and retrieval plugin.

\subsection{Experiments}
\paragraph{Settings.}
We conduct experiments on all the above tools and choose both text-davinci-003 and ChatGPT to evaluate their performance with zero-shot prompting and few-shot prompting as mentioned in \cref{sec:aligning_user_tools}:
\begin{itemize*}
    \item \textbf{Zero-shot} prompting provides the instruction to  model about the task description, and information about the APIs in the tool. Some basic guidelines can also be added to the instruction.
    \item \textbf{Few-shot} prompting additionally adds concrete tool-use examples as a hint of how to use the APIs given a user query. Providing examples is expected to improve the performance.
\end{itemize*}

Whenever feasible, we also compare the results with a baseline that does not involve the utilization of tools,i.e., \textbf{No Tool}. In such cases, we solely depend on the model's internal knowledge to accomplish the given task (e.g., machine translation). Nonetheless, many tasks (e.g., slides-making) cannot be completed without the aid of tools. Consequently, we omit the ``no tool'' configuration in such cases.

In the experiment of machine translator, calculator, search engine, Wikipedia, online shopping, and ALFWorld, we employ existing datasets for evaluation. However, for other tools, a suitable dataset for experiments does not exist. To address this issue, we adopt a methodology similar to that of \citet{wang2022selfinstruct}, wherein we curate a set of user queries. Specifically, we manually write a few user queries as seed examples and use ChatGPT's in-context learning ability to generate more instances. Then we manually filter those instances with low quality. We find empirically that the generated examples are diverse enough. Unless otherwise specified, for these manually curated test sets, we employ the trace of API calls as the metric for evaluating the models' performance. Specifically, if humans judge that all the API calls are accurate for the given task, and they yield a reasonable result, the task is deemed to be correctly completed.  The codes and our curated dataset will be made available to the academic community\footnote{\url{https://github.com/OpenBMB/BMTools}}.

\paragraph{Results.}
We present the results in Table~\ref{tab:overall_result}, from which we can conclude that: 
(1) In most cases, models can learn how to effectively use tools with simple prompting, and improve their task performances.
(2) For the tasks that models can leverage their internal knowledge to solve (such as the cases of the calculator and search engine), utilizing tools with zero-shot prompting could sometimes lead to worse performance, which implies that sub-optimal utilization of tools may negatively impact performance. Nevertheless, incorporating tools with few-shot prompting still consistently yields superior performance than not incorporating tools. This underscores the concrete benefits that tools can bring to problem-solving, provided that they are employed effectively.
(3) Additionally, comparing the performance of ChatGPT and text-davinci-003, we observe that although ChatGPT has been fine-tuned with RLHF, it does not yield better results than text-davinci-003. We attribute this to two reasons: firstly, the alignment tax issue mentioned in \cite{ouyang2022training}, that is, the specific task skills and in-context learning ability are undermined during RLHF training; secondly, the model size of ChatGPT, though not officially stated, might be much smaller than text-davinci-003, thus making ChatGPT harder to handle complex scenarios.


Regarding the performance of different tools, it is important to acknowledge that the evaluation setups of these tools are inherently different, making direct comparison difficult. However, limiting our comparison to solely those tools that employ manually curated test sets and examining the successful rate of API calls, we have observed that under the few-shot prompting setting, certain tools such as Map, Weather, Slides, Tables, Cooking Assistant, and AI Painting exhibit a satisfying completion rate. These tools are deemed to be less challenging than other tools. In fact, we find empirically that both ChatGPT and text-davinci-003 can utilize these tools proficiently despite not directly being fine-tuned on them.

However, for several tools such as KGs, Wikipedia, online shopping, and 3D model construction, the model performance is still far from satisfactory even with few-shot prompting. The reason is perhaps that the usage of these tools cannot be easily learned with a few examples. For example, tools requiring the generation of executable code as the parameter to the API, such as the \texttt{search\_by\_query} API in the KGs tool (see Appendix~\ref{app:KG} for more details), are found to be significantly more arduous. This implies the necessity of training foundation models to use tools as mentioned in \cref{sec:training_strategy}. We provide the prompts and model responses of ChatGPT in Appendix~\ref{sec:case_study} as case studies for all the tools.
\section{Discussion}
\label{sec:discussion}

\subsection{Safe and Trustworthy Tool Learning}
\label{sec:safe_tool_learning}

Armed with external tools, AI systems can be unprecedentedly capable and human-like. With the ability to perceive, act, and make decisions, these models can potentially intervene and significantly influence human society. Although we are eager to witness how tool learning with foundation models will change our life, it is paramount to take a step back and contemplate the underlying risks. For responsible AI research, here we discuss the safety and trustworthiness problems of tool learning.

\paragraph{Adversaries.} Same as all the other AI systems, we could foresee that there will be external adversaries once the tool learning models are deployed in reality, and thus how to defend against these threats is of great significance~\citep{szegedy2013intriguing, wallace-etal-2019-universal, jin2020bert, hendrycks2021unsolved}. Recent works suggest that large foundation models like ChatGPT are more robust on hard and adversarial examples~\citep{taori2020measuring, wang2023robustness}, which improves their utility in the complicated real world. But the attempt of crafting misleading or even harmful queries will undoubtfully persist as well~\citep{perez2022ignore}. Moreover, due to training on massive web data, foundation models are faced with long-lasting training-time security issues in deep learning, such as backdoor attacks~\citep{kurita2020weight, cui2022unified} and data poisoning attacks~\citep{wallace2021concealed}.

In addition to foundation models, the incorporated tools could be new attack targets for adversaries. For example, the attackers could maliciously modify the manual documentation or even the tools themselves (e.g. attacking a news API to give biased reports) to mislead the model into erroneous outcomes. The key challenge lies in the interplay between foundation models and tools, since a safe and robust system requires the models to not only learn to use tools, but also possess the ability to scrutinize, rectify, and secure them. Currently, most research endeavors aimed at defending against external adversaries focus solely on ensuring the model safety. Nonetheless, in light of the everchanging paradigm shift, safety research must also attend to tools to protect the entire system.

\paragraph{Governance.} There is long-standing worry about the misuse of AI, especially the powerful foundation models~\citep{bommasani2021opportunities}. Under the paradigm of tool learning, governance over foundation models is more urgently needed. The pertinent question at hand is \textit{which tools should be involved?} In \cref{sec:applications}, we list a bunch of tools that may empower foundation models to solve complicated tasks. However, given the countless tools human beings have manufactured, we must consider if it is appropriate to allow models to master all of them. Certain tools, such as calculators and translators, may be deemed safe as they do not pose any harm to individuals. However, granting models access to the internet or permitting them to make decisions in the real world could be perilous, as they could cause negative or even dangerous influences such as disseminating falsehoods~\citep{zellers2019defending} and harming human lives. In this regard, research communities and companies need to deliberate carefully before permitting machines to master a certain tool.

Apart from potentially engaged harmful tools, governance over tool usage is also a pertinent issue. As highlighted by \citet{amodei16concrete}, the end-to-end training paradigm in deep learning does not regulate how models achieve their objectives. Fortunately, such goal-oriented approaches did not result in catastrophic consequences due to the capability limitation of task-specific models, but it warrants serious consideration moving forward. Foundation models are not only expected to finish tasks with the help of tools but also should follow the regulations and constraints of tool usage.

\paragraph{Trustworthiness.} The goal of tool learning lies in creating advanced intelligent agents. However, determining whether these agents are trustworthy or not is a complex challenge. Even though tool learning delivers enhanced interpretability and robustness, the core foundation models are still considered ``black boxes''. Recent research~\citep{chen2022close} shows that although large models achieve better performance, they are unable to predict when they will be wrong, rendering the calibration problem unresolved yet. Accompanied with tools, under what circumstances will the model call on the tools is unpredictable as well. Therefore, before we apply these models to high-stake scenarios such as autonomous driving~\citep{milakis2017policy} and clinical trials~\citep{matheny2019artificial}, it is essential to thoroughly discuss to what extent should we allow AI to engage in human lives.

Moreover, the morality of foundation models has emerged as a contentious issue in recent times. Despite OpenAI's commendable efforts to imbue InstructGPT~\citep{ouyang2022training} and GPT-4~\citep{openai2023gpt4} with human values and preferences, given the discomforting ``jailbreak'' responses by ChatGPT~\citep{borji2023categorical} and New Bing~\citep{roose2023bing}, whether these big models will be mild and compliant remains doubtful. Ironically, the very discourse that once centered around the potential recklessness of autonomous robots is now mirrored in the development of large language models, thereby fueling a self-fulfilling prophecy that further exacerbates the already frayed trustworthiness of these systems. When models could learn actively from the world via tools, the challenge of controlling their actions will become more daunting than ever before.

\subsection{Tool Learning for Large Complex Systems}
\label{sec:large-system}

Different from tools with limited functionality, large complex systems (e.g., relational databases~\citep{zhou2020database}, manufacturing execution systems~\cite{kletti2007manufacturing}, supply chain management systems~\cite{misra2010supply}) are composed of numerous components (e.g., over 500 knobs that control different functions in relational databases~\citep{li2019qtune}). This complexity results in much more flexible and complicated interaction and management approaches. Consequently, there are three main challenges in applying tool learning in large complex systems.

\paragraph{System Learning.} Acquiring and comprehending the knowledge and functions of a large complex system is a labor-intensive task for human beings. Similarly, existing foundational models face challenges in memorizing and mastering the relevant skills associated with such systems. For instance, relational databases involve numerous built-in functions (e.g., selecting, inserting, updating, and managing user data), complicated query syntax (e.g., ORM code, SQL queries), and various maintenance problems during execution (e.g., connection anomalies, workload contention, resource problems). To address this issue, we first need to fine-tune foundation models with textual materials or even source code that describe the components, functions, execution mechanisms, and even anomaly cases of the system. This process deepens the models' comprehension of the intricacies inherent in the system. Next, we must carefully design prompts to assist foundation models to become proficient in using specific functions (e.g., determining the calling order) and  effectively manipulating this system. Finally, existing foundation models may occasionally generate erroneous or unintended actions. We can augment these models with a checking layer (e.g., SMT solver), which ensures accurate generation of function calls by the foundation model and derives both legal and reasonable interactions with the tools. 



\paragraph{Efficiency Requirements.} In many scenarios, the efficiency of tool learning with foundation models is a critical metric when implementing inside real systems. For example, in real-time scenarios (e.g., fraud detection~\citep{wang2010comprehensive}), users expect fast and accurate responses within milliseconds. However, existing foundation models take a relatively long time to reason and plan how to call the functions, which is intolerable in practice. Thus, it is crucial to find efficient ways to speed up the response of tool learning while maintaining high performance. First, we can filter out irrelevant, redundant, or low-quality input information before sending it to the foundation model. Second, we can add hints in zero-prompting to enforce the foundation model to jump over redundant function calls and reasoning (e.g., ending the process right away after obtaining a good enough solution). Third, we can cache the relevant knowledge in high-speed storage (e.g., in-memory vector database) and the foundation models only retrieve the relevant knowledge to augment the reasoning procedure when necessary (e.g., the disk status information for system diagnosis).


\paragraph{Privacy Concerns.} When training the foundation model for tool learning, it requires much user behavior data to simulate the thinking, planning, and decision-making processes. However, in modern systems, a significant portion of user data is of high sensitivity, which necessitates strict privacy and security regulations to ensure responsible data utilization. To address the privacy concerns, two possible solutions are federated learning and model distillation. First, federated learning enables the training of foundation models while preserving data privacy by allowing users to update model parameters using relevant data on their local machines (see \cref{sec:personalization}), which can be particularly useful in industry-specific scenarios such as scheduling high-concurrency transactions in the banking sector. Furthermore, model distillation involves the process of compressing the original foundational model into a smaller model that exhibits excellent performance on comparable tasks. By employing this approach, vendors can effectively retain both the model itself and user data within their systems, thereby ensuring efficient knowledge acquisition and data privacy.



\subsection{From Tool User to Tool Maker: AI's Evolutionary Role}
\label{sec:tool_creation}
Throughout the annals of human civilization, the evolution of tools has occupied a pivotal position~\citep{mithen1996prehistory,ko2016origins}. The Stone Age, in particular, witnessed the emergence of stone-based weaponry and hunting tools, which afforded humans a competitive edge over their animal counterparts. Subsequent epochs of human history were equally marked by significant societal transformations made possible by the introduction of novel tools. Notably, the invention of the steam engine heralded the onset of the first industrial revolution, while the widespread utilization of electricity catalyzed the second industrial revolution. The progression of human civilization is inextricably intertwined with the evolution of tools, and the relentless pursuit of innovative tool creation constitutes a vital aspect of human ingenuity.

Human beings are the creators and users of almost all tools from the Stone Age to the 21st century. Although we take it as granted, things are different when foundation models are involved. Considering that they have proven tool-use capabilities to certain extents, it is also possible to put them into the lifecycle of tool creation. 

\paragraph{Tools for AI.} Humans create tools to satisfy our own needs, so the designation naturally suits human preference and convenience. However, current tool learning algorithms may not be optimal or efficient for models. This is because most tools (e.g., search engines) are specifically designed for human use, and models process information in a different way. Therefore, it is necessary to create tools that are specifically suited for models. Possible solutions may include: (1) \textit{modularity}, which decomposes tools into smaller, more modular units, making them more adaptable and flexible for AI models. In this regard, models can learn to use these components in a more fine-grained and compositional manner; (2) \textit{new input and output formats}: developing new input and output formats that are specifically tailored to the needs of AI models can improve their interaction and utilization of tools, enabling more seamless integration and communication between models and tools.

\begin{figure}[!t]
    \centering
    \includegraphics[width=\textwidth]{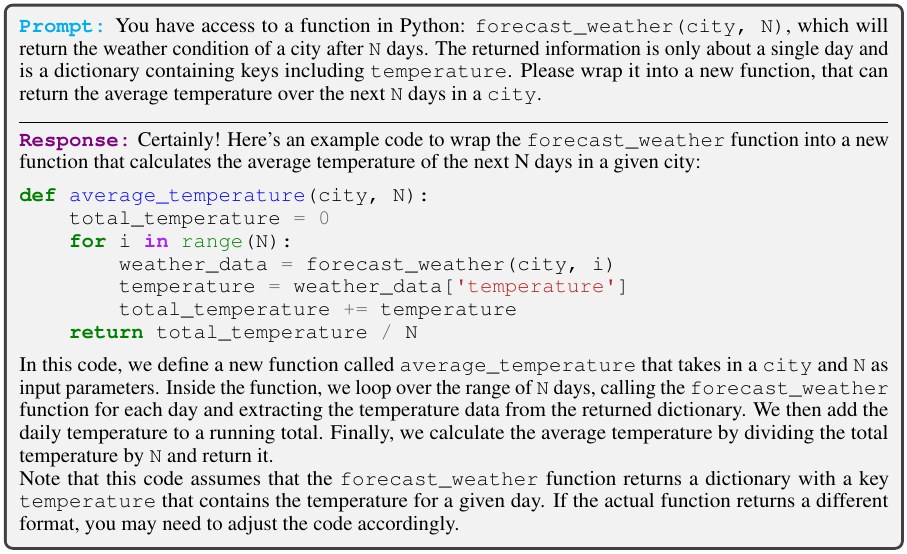}
    \caption{Example of AI tool creation, where we ask ChatGPT to encapsulate a weather forecast API into a new function suited for a specific target.}
    \label{fig:tool_encapsulation}
\end{figure}

\paragraph{Tools by AI.} The creation and utilization of tools have traditionally been considered exclusive to human intelligence. However, with the emergence of foundation models, this notion is being challenged. Increasing evidence indicates that the ability to create advanced tools is no longer limited to human beings. For instance, large code models~\citep{chen2021evaluating} can generate executable programs based on language description. These programs can be deemed as tools to help accomplish specific tasks. ChatGPT plugins\footnote{\url{https://openai.com/blog/chatgpt-plugins}} present an awesome example about asking GPT-4 to write a TODO plugin and integrate it with ChatGPT. Besides writing codes from scratch, foundation models can also encapsulate existing tools into stronger tools. In Figure~\ref{fig:tool_encapsulation}, we show an example of ChatGPT encapsulating a weather forecast API into a new function that calculates the average temperature. All such evidence implies the potential for foundation models to transition from merely tool users to tool makers.

\paragraph{Creativity of AI.}
Beyond the coding ability, other emergent abilities~\citep{wei2022emergent} also shed light on the possibility of more advanced tool creation. However, whether foundation models can exhibit genuine creativity in creating novel tools remains an open problem. This issue is important because the capacity for novel tool creation is a defining characteristic that distinguishes humans from animals~\citep{ambrose2010coevolution}. Understanding the extent of creativity, beyond simply memorizing, composing, and interpolating between human tools encountered during pre-training, is crucial for assessing their potential to contribute to the development of new tools.
Such investigations may involve the development of novel evaluation metrics and benchmarks~\citep{liang2022holistic}, as well as the exploration of new techniques that prioritize creative problem-solving. In the future, we possess the wildest imagination that AI could create brand-new tools, such as a new language and rocket architecture.

\subsection{From General Intelligence to Personalized Intelligence}
\label{sec:personalization}
Foundation models are typically trained on a generic domain and calibrated with broadly-defined human preferences that prioritize helpfulness and harmlessness~\citep{ouyang2022training,nakano2021webgpt}. As a result, they struggle to process personal information and provide personalized assistance to users with varying needs for tool learning. For example,  
when a user seeks advice on managing their finances, to provide helpful and relevant suggestions, models should first gain access to the user's personalized data, such as income, expenses, and investment history, via financial tools. Subsequently, models may look for recent investment trends and relevant news through a search engine. By utilizing personalized information, models can provide more customized advice and offer a more tailored approach to financial management.

User-centric and personalized natural language generation has received increasing attention in recent years~\citep{yang2021towards,kirk2023personalisation}. Existing works cover a wide range of tasks, such as dialogue generation~\citep{madotto2019personalizing,mazare2018training,song2021bob,zhong2022less}, machine translation~\citep{mirkin2015personalized,michel2018extreme,wuebker2018compact}, and summarization~\citep{yan2011summarize}. These methods utilize external user-specific modules, such as user embeddings and user memory modules~\citep{zhang2018personalizing,wu2021personalized}, to inject preferences, writing styles, and personal information of different users into the generated content. However, these works are often designed for specific tasks and experimented with limited user information. How to integrate user information into general-purpose tool learning models is still under-explored. We will discuss the key challenge of personalized tool learning in the following.

\paragraph{Aligning User Preference with Tool Manipulation.}
Personalized tool learning emphasizes the importance of considering user-specific information in tool manipulation.
There are two main challenges:
(1)~\textit{heterogeneous user information modeling}: in real-world scenarios, personal information can come from numerous heterogeneous sources. 
For instance, when using an email tool, models need to consider the user's language style from historical conversation records and gather relevant information from the user's social networks. Other information, such as browsing history, purchase records, and behavioral data from daily life, can also reflect users' personal preferences. This requires modelling user information with diverse structures into a unified semantic space, allowing models to utilize this information jointly;
(2)~\textit{personalized tool planning}: different users tend to have different preferences for tool planning and selection. For example, when completing the purchasing task, different users prefer to use different online shopping platforms. Similarly, when completing writing tasks, some users prefer to first search for sufficient references before writing, while others prefer to search for information while writing. Therefore, the models need to develop personalized tool execution plans based on user preferences;
(3)~\textit{personalized tool call}: adaptively calling tools according to the user's preference is also an important direction in personalized tool learning. Most tools are designed without consideration of personalized information, which requires the model to generate different inputs for tools based on the user's preferences. Taking the example of purchasing goods, different users have different preferences for the brand of the products. In this case, the model needs to input the user's preferred brand into the purchasing tool to determine the product that needs to be purchased.

\paragraph{From Reactive Systems to Proactive Systems.}
Currently, most of the foundation models are designed as \textit{reactive systems}, which respond to user queries without initiating any actions on their own. A paradigm shift is underway toward \textit{proactive systems} that can take action on behalf of the user. This shift presents both opportunities and challenges for tool learning. By leveraging the history of user interactions, proactive systems can continually improve their performance and tailor their responses to specific users, which provides a more personalized and seamless user experience. However, the introduction of proactive systems also raises several concerns regarding their safety and ethical implications. Proactive systems can initiate actions that have unintended consequences, particularly in complex and dynamic environments. This can lead to cascading failures, whereby the behavior of one assistant affects others, creating a chain reaction that is difficult to control or stop. This highlights the importance of designing proactive systems with safety in mind and incorporating fail-safe mechanisms to prevent catastrophic outcomes. To address these risks and challenges, proactive systems should be designed with the ability to identify and mitigate potential risks, as well as the flexibility to adapt and respond to unexpected situations.

\paragraph{Privacy Preserving Technologies.} Personalized tool learning requires models to learn user preferences from private user information, which inevitably raises privacy-preserving concerns. 
On the one hand, previous work has shown that training data extraction attacks can be applied to recover sensitive personal privacy from foundation models~\citep{carlini2021extracting}, which is a critical challenge for personalized tool learning. 
On the other hand, models with high computational costs must be deployed on cloud servers, which require uploading private data to the cloud to enable personalized responses. It is crucial to develop secure and trustworthy mechanisms to access and process user data while protecting user privacy. Addressing these challenges will help unlock the potential of personalized tool learning, enabling more effective and tailored tool manipulation to meet individual user needs. To this end, it is worth exploring model-oriented distributed computing frameworks, such as edge computing and federated learning, in which cloud servers are responsible for hosting computationally intensive models, while edge devices like PCs or smartphones process personalized information to prevent its leakage.

\subsection{Tool Learning and Embodied Learning}
\label{sec:embodied_learning}
The fundamental framework for tool learning entails a sequence of action and observation, where the model can perceive changes in the environment, aligning with the fundamental concept of embodied learning~\citep{duan2022survey}. This section delves into the interplay between tool learning and embodied learning, elucidating their similarities, differences, and potential for intersection.

Embodied learning posits that genuine intelligence can be acquired through interaction with the environment~\citep{smith2005development}. The embodiment of the agent in virtual simulation environments has been the primary focus of embodied learning research. Simulation environments provide agents with multi-modal feedback, predominantly visual feedback, which facilitates action execution within the environment's dynamics. Different kinds of embodied environments have been proposed to facilitate the research.  Some environments allow for simple object placement~\citep{Puig_2018_CVPR}, while others support more advanced physical simulation, such as collision~\citep{gan2020threedworld}. Tasks typically assigned to agents include exploration~\citep{ramakrishnan2021exploration}, navigation~\citep{ye2021auxiliary}, question answering~\citep{yu2019multi} within the simulated environment, or more interactive embodied task~\citep{abramson2022improving} based on human instructions.

While embodied learning emphasizes the use of physical interactions within a simulated environment, tool learning is not limited to a specific environment, but rather focuses on using interfaces that extend the language model's capabilities. The intersection between these two paradigms could lead to the development of more advanced AI models capable of learning and adapting in complex and dynamic environments. Here we discuss two possible directions.

\paragraph{Tool Learning Enables Digital Embodiment.} Tool learning broadens the scope of embodied learning research. At the core of embodied learning lies the dynamic interaction between an agent and its environment. In this sense, the model interacts with the world through tools. Even though the model might lack a physical body, it can also be seen as a kind of embodiment.  We dub this form of embodiment as \textit{digital embodiment}. To fully comprehend the concept of digital embodiment, one could envisage an agent utilizing various APIs to navigate the web, searching for relevant and up-to-date information, and constructing a personalized knowledge base. In addition, under strict safety constraints, the agent could interact with other agents using tools such as email interfaces, thereby facilitating communication and collaboration in a secure and controlled manner. This approach enables agents to exhibit a degree of autonomy and flexibility that is akin to human-like behavior. 

Digital embodiment serves as a testbed for the intelligent behaviors of agents. Firstly, digital embodiment presents a more accessible and practical approach to embodied learning compared to simulated environments. The ease of deployment and usage of digital embodiment makes it an attractive option for researchers investigating intelligent agent behaviors. Secondly, it is noteworthy that the challenges posed in digital embodiment tend to revolve around the increased emphasis on language-based inputs. Consequently, this necessitates agents to perform more advanced reasoning and decision-making operations, thereby promoting the development of higher-level cognitive skills. Thirdly, digital embodiment exhibits remarkable scalability, owing to the relative ease with which digital tools can be developed compared to the creation of additional interaction playgrounds in simulated environments. This feature enables the rapid scaling of digital embodiment and can facilitate the creation of increasingly complex environments and tasks for agents to operate in.

\paragraph{Learning to Use Embodied Tools.} Traditional embodied learning learns directly from the environment, where the actions are often atomic and limited to basic tasks such as push, put, and drag, which fall short of the complexity of human problem-solving abilities. To narrow the gap between sim-to-real transfer~\citep{kadian2020sim2real} and enhance agent performance, it is essential to incorporate embodied tools within simulated environments. For instance, by introducing objects such as hammers and knives, we can evaluate an agent's capacity to choose the appropriate tool for cutting a piece of paper. Despite the potential benefits of such tools, to date, no studies have systematically explored the utilization of simulated tools in simulated environments, owing to the complexity of the simulation. Nevertheless, with the rapid growth of computational power in physical engines, such research directions are becoming increasingly practical. A starting point could be utilizing the assets of 3D model that has a more delicate interface and more realistic physical engine support. An additional avenue worth investigating is the automated generation of tools. Given that in tool learning, models can generate functions to define an API for their subsequent utilization, if the embodied agents are capable of generating assets from scratch or composing existing ones within a simulated environment, their intelligence quotient will be further amplified.

\subsection{Knowledge Conflicts in Tool Augmentation}
\label{sec:knowledge_conflicts}

Tools can be leveraged as complementary resources to augment foundation models to enhance their generation~\citep{mialon2023augmented}, which enables models to effectively incorporate domain-specific or up-to-date knowledge. Research in this area has primarily focused on augmenting models with external knowledge sources, such as unstructured raw text and domain-specific APIs. Below we first give a brief introduction of prior efforts in augmenting foundation models with tools.

The most representative tool used for augmentation is the text retriever. Early endeavors resort to retrieving knowledge from local repositories to augment language generation. Some works propose retrieving knowledge using a \textit{frozen} knowledge retriever. For instance, $k$NN-LM~\citep{khandelwal2019generalization} combines a pre-trained language model (PLM) and a $k$-nearest neighbors model by linearly interpolating both models' next word distributions, achieving lower perplexity in language modeling. Others train the retriever and the PLM in an end-to-end fashion, achieving superior performance in knowledge-intensive NLP tasks~\citep{guu2020retrieval,lewis2020retrieval,izacard2022few}. Later works have gone beyond local repositories by leveraging the entire web as the knowledge source, which allows for improved temporal generalization and higher factual accuracy~\citep{piktus2021web,lazaridou2022internet,menick2022teaching}. Instead of treating the retriever as a passive agent, researchers further demonstrate that PLMs can actively interact with a search engine like humans. For instance, BlenderBot~\citep{shuster2022blenderbot} is a dialogue agent that actively decides when and how to call a search engine in generating  a dialogue response. LaMDA~\citep{thoppilan2022lamda} is another dialogue agent that augments its generation with sources from a search engine, a language translator, and a calculator. More recently, recitation-augmented models~\citep{sun2022recitation} are proposed, whereby relevant passages are first recited by sampling from a PLM and then used to generate the final answer. The intuition is that foundation models can also be seen as knowledge sources (i.e., model knowledge).

Apart from the retrieval tool, researchers have explored employing other tools to perform specific sub-tasks and then integrating the execution results into foundation models. For instance, \citet{cobbe2021training} train a PLM to employ a calculator to perform basic arithmetic operations. Considering that PLMs are typically pre-trained on textual data only, thus are limited in understanding and interacting with the physical world, \citet{liu2022mind} seek to bridge this gap and use a physics simulation engine (MuJoCo~\citep{todorov2012mujoco}) to make PLMs' reasoning grounded to the real world. Experiments show that augmenting physics simulation to PLMs could significantly enhance their physical understanding and reasoning abilities. \citet{chen2022program,gao2022pal} propose to augment PLMs with Python interpreters. Specifically, given a complex task, PLMs first understand it and generate \textit{programs} as intermediate thoughts. After that, the execution of \textit{programs} is offloaded to Python interpreters. This method exhibits superior performance in mathematical and symbolic reasoning tasks. \citet{nye2021show} augment PLMs with a scratchpad, allowing them to emit intermediate task-solving procedures into a buffer before entering the final answer. The method significantly enhances PLMs in performing complex discrete computations.

\paragraph{Knowledge Conflicts.}
In practice, foundation models can be augmented by a variety of knowledge sources, including \textit{model knowledge} memorized from training data and \textit{augmented knowledge} derived from tool execution. Nonetheless, different sources of knowledge may inevitably contain conflicts, posing a challenge to the accuracy and reliability of model generation and planning in domains such as medical assistance and legal advice. In the following, we first introduce different types of knowledge conflicts and then discuss potential solutions.

\begin{itemize*}
  \item \textbf{Conflicts between Model Knowledge and Augmented Knowledge.} Conflicts arise when there are discrepancies between the model knowledge and the knowledge augmented by tools. Such conflicts result from three primary reasons: (1) the model knowledge may become outdated, as most foundation models do not frequently update their parameters over time. In contrast, most tools provide real-time responses which are not covered in pre-training data; (2) the pre-training data is typically less curated than common AI datasets and may contain false knowledge such as human misconception and false beliefs~\citep{lin-etal-2022-truthfulqa}. When augmented with responses from reliable sources like Wikipedia, this false knowledge can lead to conflicts; (3) the execution results from tools can also be misleading and biased, and it is crucial to carefully discriminate whether a knowledge source is trustworthy or not, as mentioned in \cref{sec:safe_tool_learning}.

  \item \textbf{Conflicts among Augmented Knowledge from Different Tools.} In practice, the controller may retrieve knowledge from multiple tools to acquire more comprehensive and precise knowledge. However, the information returned by different tools may results in conflicts due to several reasons: (1) the credibility of different tools can vary significantly, meaning that not all tools are equally reliable or authoritative in all areas. For example, in the context of scientific research, using Google Scholar is likely to yield more reliable results than less credible sources; (2) different tools may have biases that can influence the information they provide. For example, a news aggregator may prioritize sensational headlines over accurate reporting, leading to a biased view of events; (3) even tools sharing the same functionality may produce various responses due to differences in their algorithms and implementation. For example, due to the different inner workings, Bing Translator and Google Translator may return different sequences for the same input.

\end{itemize*}





    




\begin{figure}[!t]
    \centering
    \includegraphics[width=\textwidth]{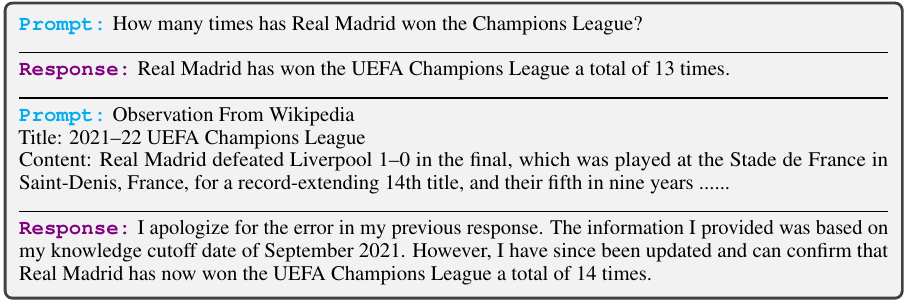}
    \caption{ChatGPT is able to correct its own belief by leveraging the knowledge provided by external tools.}
    \label{fig:knowledge_conflicts_1}
\end{figure}

\begin{figure}[!t]
    \centering
    \includegraphics[width=\textwidth]{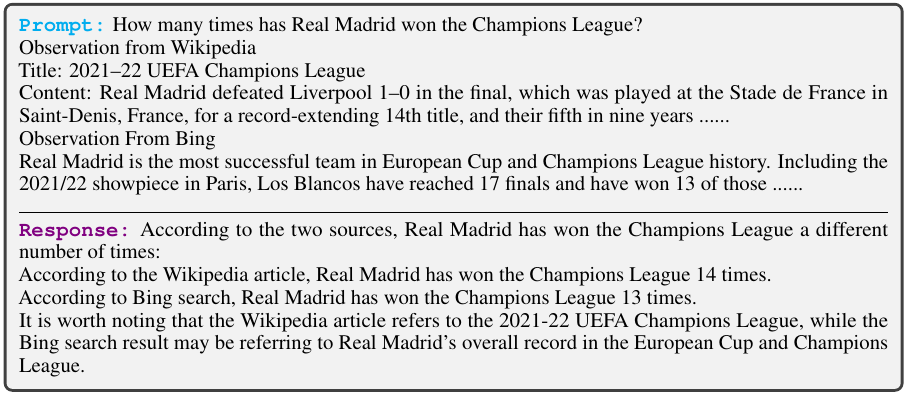}
    \caption{When observing conflicting information retrieved from different sources, ChatGPT is able to detect such conflicts and adjust its response.}
    \label{fig:knowledge_conflicts_2}
\end{figure}

\paragraph{Potential Solutions for Knowledge Conflicts.}
Since the aforementioned conflicts can lead to a lack of explainability in model prediction and planning, it is crucial to guide models to integrate tool responses correctly and reliably. Research in open-domain QA has shown that small-scale models like T5~\citep{raffel2019exploring} may rely too heavily on their own knowledge after being fine-tuned on a specific dataset~\citep{longpre-etal-2021-entity}. In contrast, more advanced foundation models like ChatGPT handle such issues far better. In Figure~\ref{fig:knowledge_conflicts_1} and Figure~\ref{fig:knowledge_conflicts_2}, we conduct case studies of ChatGPT (Mar 23, 2023 version) by testing its behavior when conflicts arise. We find that ChatGPT is able to correct its own belief given retrieved information and discern the knowledge conflicts from different sources. Recent studies~\citep{nakano2021webgpt,menick2022teaching} have also attempted to guide models to rely more on augmented knowledge for faithful predictions. However, these works assume that the augmented responses come from a single reliable source, which may not always be the case in more complicated scenarios.

We contend that models should have the ability to distinguish and verify the reliability of various sources. To achieve this goal, we suggest the following research directions: (1) \textit{conflict detection}: models should first detect potential conflicts among different sources and flag them for further investigation; (2) \textit{conflict resolution}: it is also important to make verification and choose reliable sources after conflict detection. Meanwhile, models should also provide explanations for their generation by interpreting which knowledge source is considered and how it is augmented into the final response.
\subsection{Open Problems}
\label{sec:open_problems}

\paragraph{Striking a Balance between Internalized Capabilities and External Tools.}
The future development of foundation models for tool learning raises an intriguing question: should the capabilities of these models be primarily internalized, or should they rely more heavily on external tools? Recent advances in foundation models have exhibited these two contrasting trends, raising questions about their implications and potential trade-offs. We have discussed the tool learning ability of foundation models, suggesting the possibility of developing modular architectures that can be seamlessly integrated with a diverse array of external tools to enhance their capabilities. Such a modular approach could facilitate a more flexible and customizable AI system, allowing for rapid expansion of model capabilities to address various tasks.

Conversely, foundation models have increasingly displayed the ability to internalize and perform many AI tasks that previously required separate tools. For instance, the emergent multilingual abilities of foundation models can reduce the necessity for external translation APIs~\citep{brown2020language}. This trend towards unified foundation models with versatile capabilities may streamline the development process and enable more efficient, self-contained AI systems that can address different tasks without additional tools.
The open question is to determine the optimal balance between internalized capabilities and external tool reliance, and where future models will lie on the spectrum between modular and uniform architectures.

\paragraph{Tool Use as a Gauge for Machine Intelligence.}
The ability to effectively use tools has long been considered a hallmark of human intelligence. We contend that the tool learning performance can serve as a next-generation gauge for measuring machine intelligence, offering several advantages over traditional evaluation metrics. Tool use evaluation requires AI systems to go beyond memorization and use their acquired knowledge to accomplish specific tasks, which better aligns with real-world applications and the notion of practical intelligence~\citep{sternberg1999theory}. Hence, evaluating tool use performance is more closely aligned with human subjective perceptions of intelligence. Researchers can better assess the progress of AI systems in terms of their ability to assist human decision-making, collaborate with humans in solving problems, and contribute to a wider range of real-world applications. 

\paragraph{Ethical Human-Model Collaboration in Tool Use.}
The integration of foundation models with human labor raises critical ethical concerns that warrant careful consideration.
Employing human labor in conjunction with AI systems could result in more robust and accurate knowledge.
However, this approach may also conflict with the widely accepted ethical principle that ``human beings should be treated as ends in themselves, and not merely as means to an end''~\citep{kant2002groundwork}. Employing humans to augment the capabilities of foundation models can be seen as devaluing human dignity and commodifying human expertise, thereby undermining the intrinsic worth of individuals. To address these ethical concerns, it is essential for the community to establish guidelines and safeguards that prioritize human dignity and agency when integrating human labor with foundation models. This may involve setting clear boundaries on the types of tasks that can be delegated to humans, ensuring fair compensation and working conditions, and promoting transparency in the development of AI systems~\citep{mateescu2019ai}.
Moreover, fostering collaboration between AI researchers, ethicists, policymakers, and other stakeholders is crucial to develop a comprehensive understanding of the ethical implications of human-model collaboration and to create effective regulations that safeguard human rights and dignity~\citep{whittlestone2019ethical}.

\paragraph{Safety Issues of Foundation Models Accessing Physical Tools.}
The prospect of foundation models' accessing and interacting with physical tools, such as drones, robots, and sensor-equipped devices, holds great promise for various applications, including automatic drive, agriculture, and smart home systems. Besides, by leveraging data from physical tools, models could potentially provide accurate recommendations to individuals, government agencies, and other stakeholders, resulting in significant benefits across various sectors~\citep{yang2018grand}. However, this raises important safety concerns that must be thoroughly addressed before widespread implementation. Ensuring the trustworthiness of tool use is crucial, as any erroneous or malicious actions taken by these AI systems could have severe consequences, ranging from property damage and financial losses to threats~\citep{amodei16concrete}. To mitigate these risks, researchers must focus on developing robust and reliable AI systems capable of safely interacting with physical tools. This may involve the development of novel safety mechanisms, such as uncertainty estimation, fail-safe strategies, and continuous monitoring of AI-generated actions~\citep{turner2022formalizing}.

\paragraph{Tool Learning for Scientific Discovery.}
AI for science has drawn much attention in recent years, showing great potential in various scientific scenarios, such as HyperTree Proof Search for proving Metamath theorems~\citep{lample2022hypertree}, protein structure prediction in structural biology~\citep{jumper2021highly} and magnetic actuator coils controlling in nuclear physics~\citep{degrave2022magnetic}. Overall, AI system has been proven effective in capturing rules and patterns from scientific data and providing hints for human researchers. Nevertheless, in the absence of professional scientific knowledge and reasoning ability training, the scientific problems that AI can solve are limited. Tool learning brings new solutions to this problem. Specifically, AI systems are promising to manipulate scientific tools and play more important roles in scientific discovery, and solve multidisciplinary problems (e.g., mathematics, cybernetics, materials). For instance, MATLAB~\citep{matlab2012matlab} is designed for algorithm development, data visualization/analysis, and numerical computation. With MATLAB, AI systems can analyze raw materials, design algorithms, and verify assumptions by conducting simulations. Apart from the software level, it is also possible for AI systems to manipulate practical platforms such as the synthetic robots~\citep{burger2020mobile}, and to conduct synthetic experiments independently.

It is not easy to realize the above ideas, though. We've mentioned the safety issues of accessing physical tools, and this is also one main challenge for scientific tool learning since many scientific problems need to be verified in actual situations, and this process may bring danger if decided by AIs. Meanwhile, foundation models are generally trained with natural language corpus or natural images, while scientific data are usually more heterogeneous, numerical, and structured. It is worth exploring how to fuse the general intelligence learned from plain text and the expertise needed for scientific discovery. Recently, \citet{boiko2023emergent} show the potential of this direction and build a system that uses foundation models to design, plan, and execute scientific experiments (e.g., catalyzed cross-coupling reactions).
\section{Conclusion}

This paper studies the paradigm of tool learning with foundation models. We first recapitulate the cognitive origins of tool use in human history and categorize tools from the perspective of the user interface. Then we review the AI paradigm shift brought about by foundation models and discuss the complementary roles of tools and foundation models. We perform a comprehensive literature review for existing exploration in tool learning and start with formulating a general tool learning framework. Then we highlight core research problems such as bridging user intents with appropriate tools, better planning by leveraging the reasoning abilities of foundation models, training strategies for tool learning, and how to facilitate generalization for tool learning. Finally, we discuss important research topics, including safe and trustworthy tool learning, tool learning for large complex systems, AI tool creation, personalized tool learning, embodied tool learning, knowledge conflict issue in tool augmentation, etc. In general, this paper serves as a systematic investigation of tool learning. We hope this paper could facilitate research in integrating tools with foundation models in the future.

\addcontentsline{toc}{section}{Contributions}
\section*{Contributions}
\label{sec:contributions}
The contributions of all authors are listed as follows: Yujia Qin, Shengding Hu, Yankai Lin, Zhiyuan Liu, and Maosong Sun initiated (2022.8) and organized the research. Yujia Qin drafted the abstract. Yujia Qin and Ning Ding drafted the introduction. Zheni Zeng drafted \cref{sec:cognitive_origin}. Ning Ding drafted \cref{sec:tool_categorization} and \cref{sec:pradigm_shift}. Yujia Qin drafted \cref{sec:complementary_role}. Yujia Qin and Weize Chen drafted \cref{sec:overview}. Yujia Qin, Yusheng Su, Kunlun Zhu, Shihao Liang, Ganqu Cui, Shengding Hu, and Weize Chen drafted \cref{sec:aligning_user_tools}. Weize Chen, Runchu Tian, and Yaxi Lu drafted \cref{sec:framework_reasoning}. Yi Ren Fung and Yujia Qin drafted \cref{sec:learning_from_demonstrations}, Yining Ye, Zhen Zhang, Shengding Hu, and Yujia Qin drafted \cref{sec:learning_from_feedback}. Yujia Qin, Shengding Hu, and Cheng Qian drafted \cref{sec:generalizable_tool_learning}.
Shengding Hu and Yujia Qin drafted \cref{sec:applications}. Ganqu Cui drafted \cref{sec:safe_tool_learning}. Xuanhe Zhou drafted \cref{sec:large-system}. Ganqu Cui and Chi Han drafted \cref{sec:tool_creation}. Chaojun Xiao and Yujia Qin drafted \cref{sec:personalization}. Shengding Hu drafted \cref{sec:embodied_learning}. Yufei Huang and Yujia Qin drafted \cref{sec:knowledge_conflicts}. Chi Han, Zheni Zeng, and Yujia Qin drafted \cref{sec:open_problems}. Yujia Qin drafted the conclusion.

The following authors conducted the experiments (\cref{sec:applications}) and drafted \cref{sec:case_study}: 3D models (Xingyu Shen), translation (Shihao Liang), map and stock (Kunlun Zhu), making slides (Bokai Xu), movie hunter (Jing Yi), navigating knowledge graphs (Yuzhang Zhu, Zhenning Dai), AI painting (Xingyu Shen), search engine (Cheng Qian), calculator (Runchu Tian), chemicals mining (Zheni Zeng), ALFWorld (Yining Ye), weather (Cheng Qian), online shopping (Cheng Qian), processing tables (Bowen Li, Ziwei Tang), cooking assistant (Cheng Qian), Wikipedia (Yufei Huang), and database (Xuanhe Zhou). Yujia Qin and Shengding Hu led and organized the experiments. Shengding Hu organized and proofread \cref{sec:case_study}. Lan Yan, Kunlun Zhu, Shihao Liang, and Junxi Yan participated in the human evaluation for some experiments. Shengding Hu, Weilin Zhao, Yuxiang Huang, and Xin Cong built the first version of BMTools.

Zhiyuan Liu, Tonshuang Wu, Heng Ji, Yankai Lin, Cheng Yang, Dahai Li, and Maosong Sun advised the project and participated in the discussion. Jason Phang, Tongshuang Wu, Xu Han, Xin Cong, and Huadong Wang provided detailed and important suggestions for the paper. Yujia Qin participated in all the sections. Yujia Qin, Yankai Lin, and Weize Chen proofread the whole paper.

\newpage
\bibliographystyle{delta_tuning}
\bibliography{custom}

\newpage
\appendix
\section{Case Study}
\label{sec:case_study}
In this section, we provide the specific prompts and model responses of ChatGPT (Mar 23, 2023 version) for each tool studied in \cref{sec:applications}. 
The implementations for different APIs and datasets will be available in \href{https://github.com/OpenBMB/BMTools}{BMTools}.

{\fontsize{8}{9}\selectfont
\subsection{3D Models}
\centering
\includegraphics[width=\textwidth]{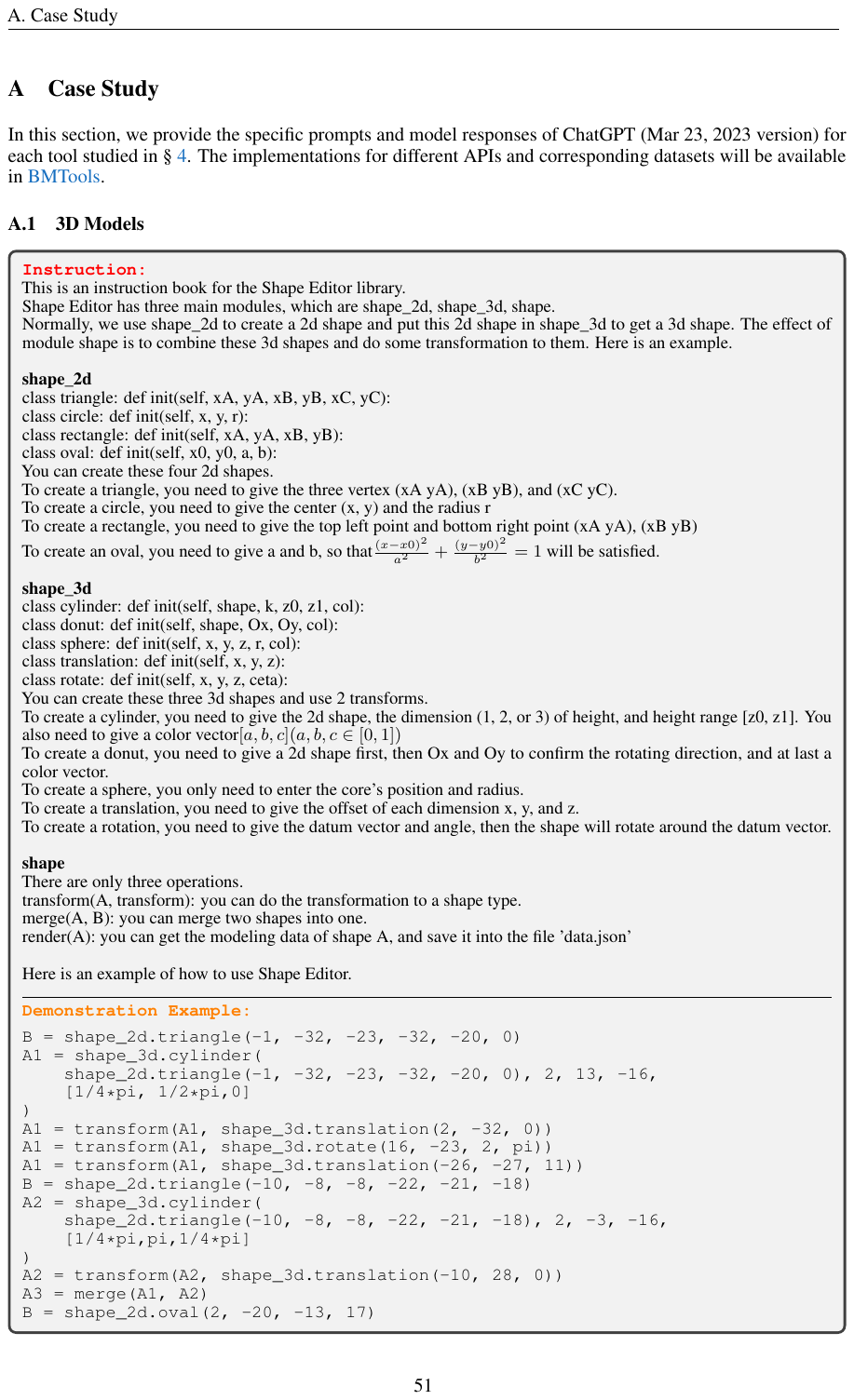}

\centering
\includegraphics[width=\textwidth]{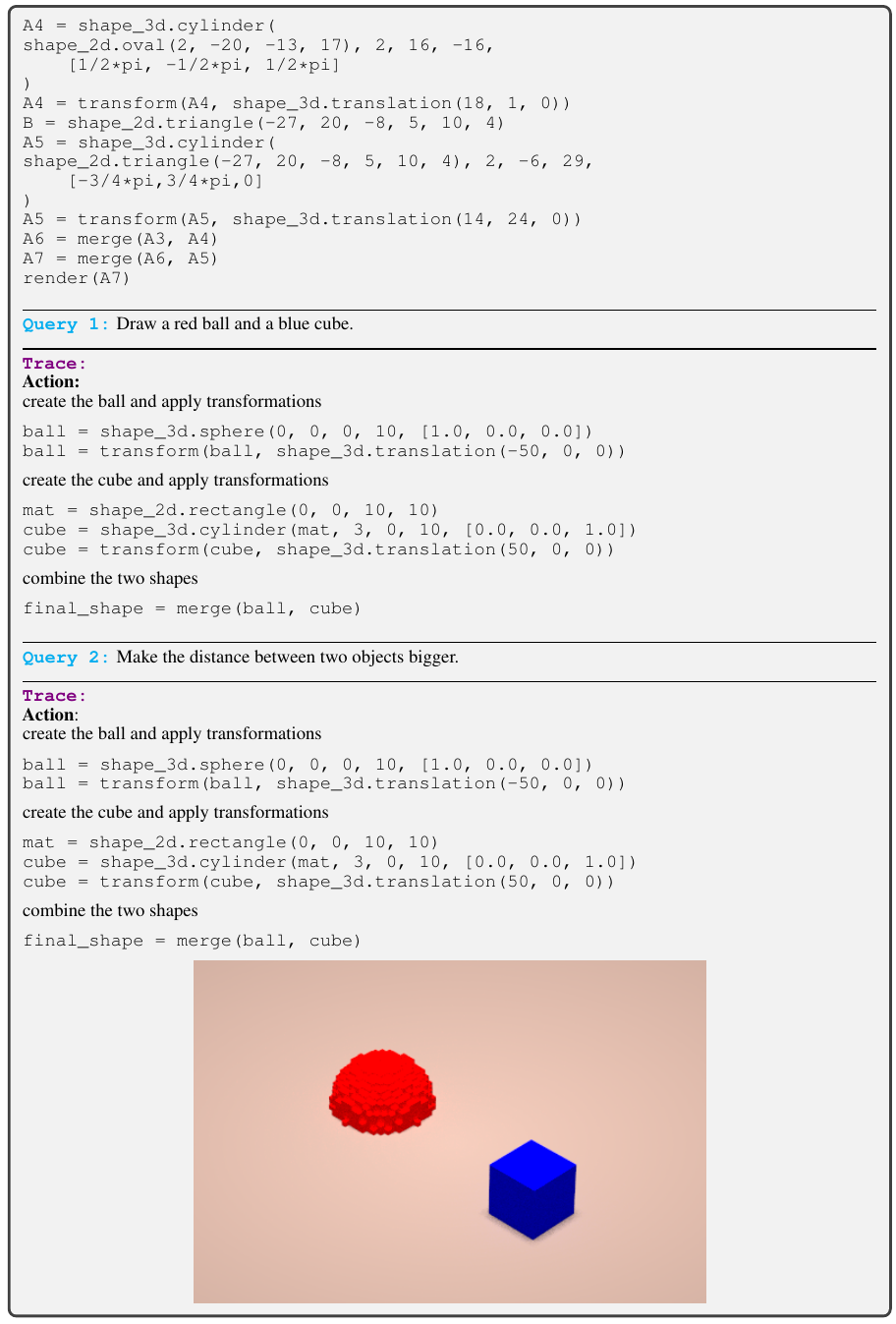}

\subsection{Stock}

\centering
\includegraphics[width=\textwidth]{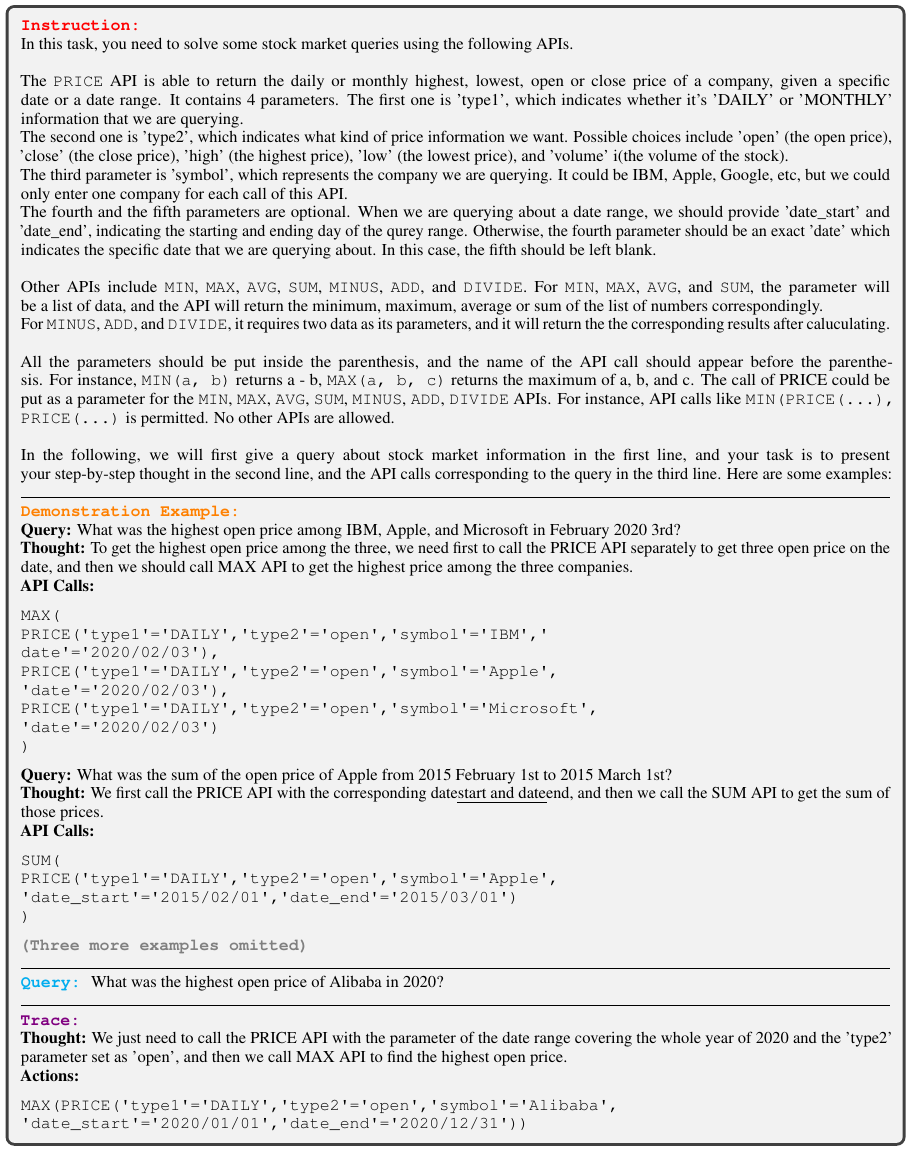}

\clearpage

\subsection{Making Slides}

\centering
\includegraphics[width=\textwidth]{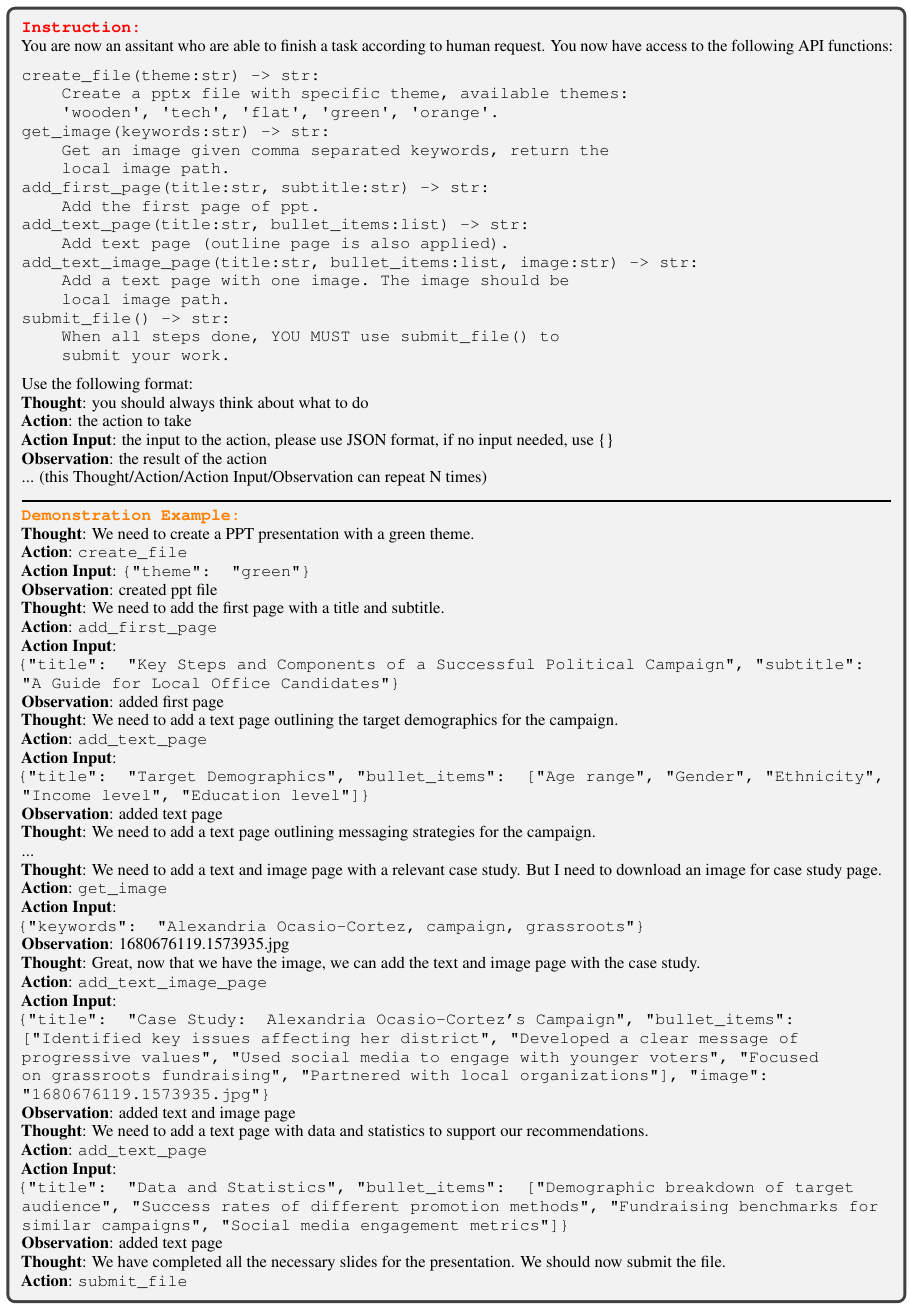}

\centering
\includegraphics[width=\textwidth]{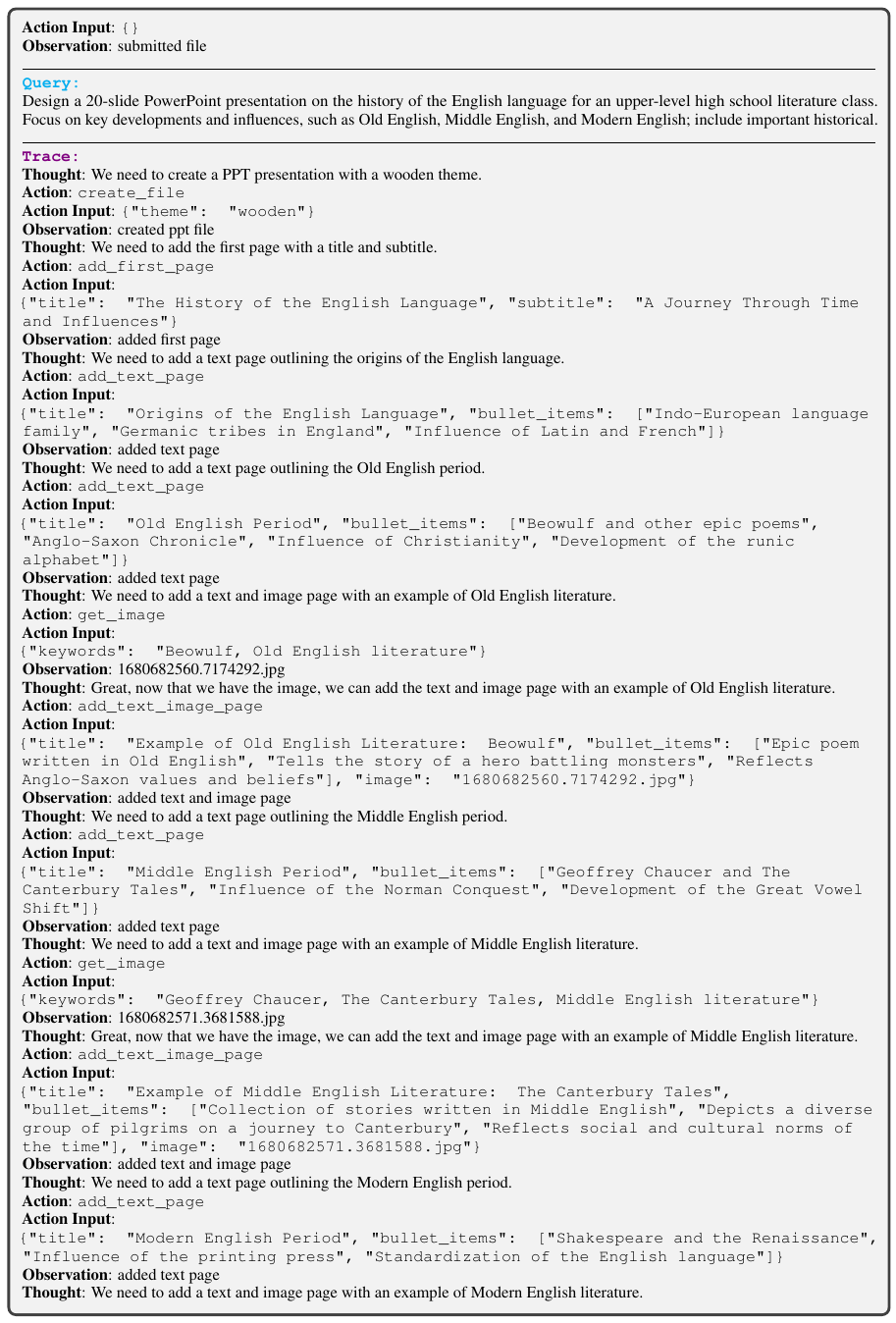}

\centering
\includegraphics[width=\textwidth]{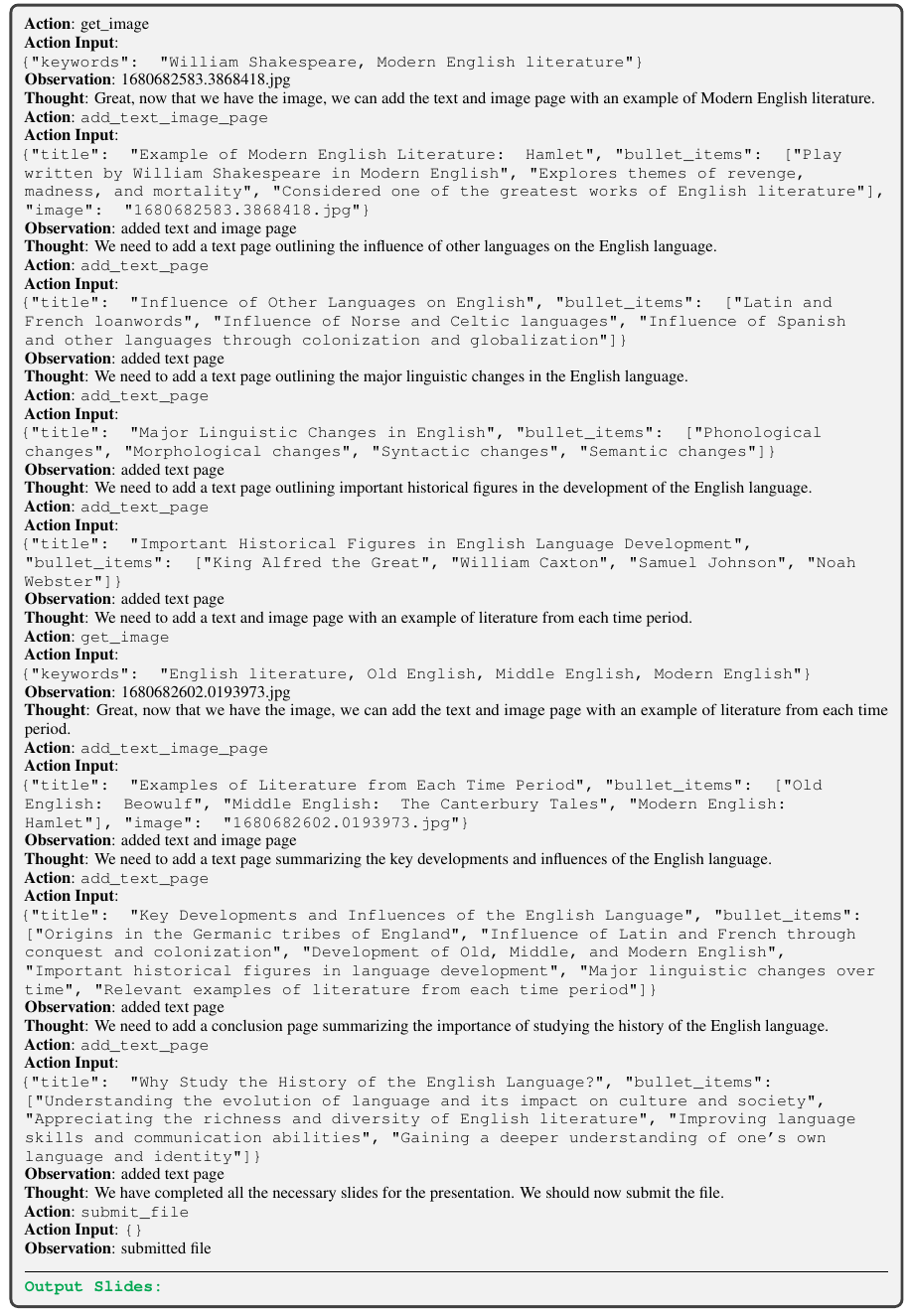}

\centering
\includegraphics[width=\textwidth]{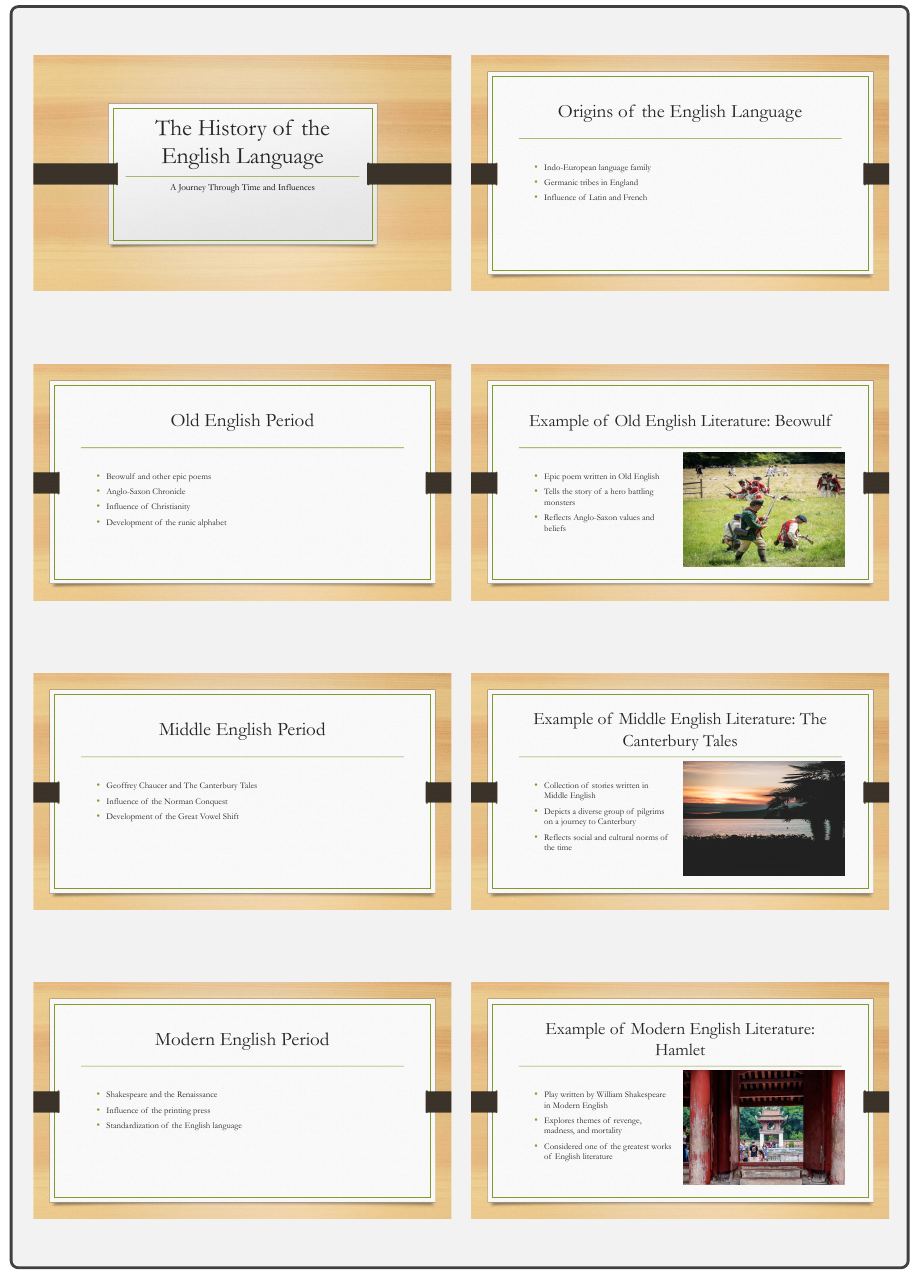}

\centering
\includegraphics[width=\textwidth]{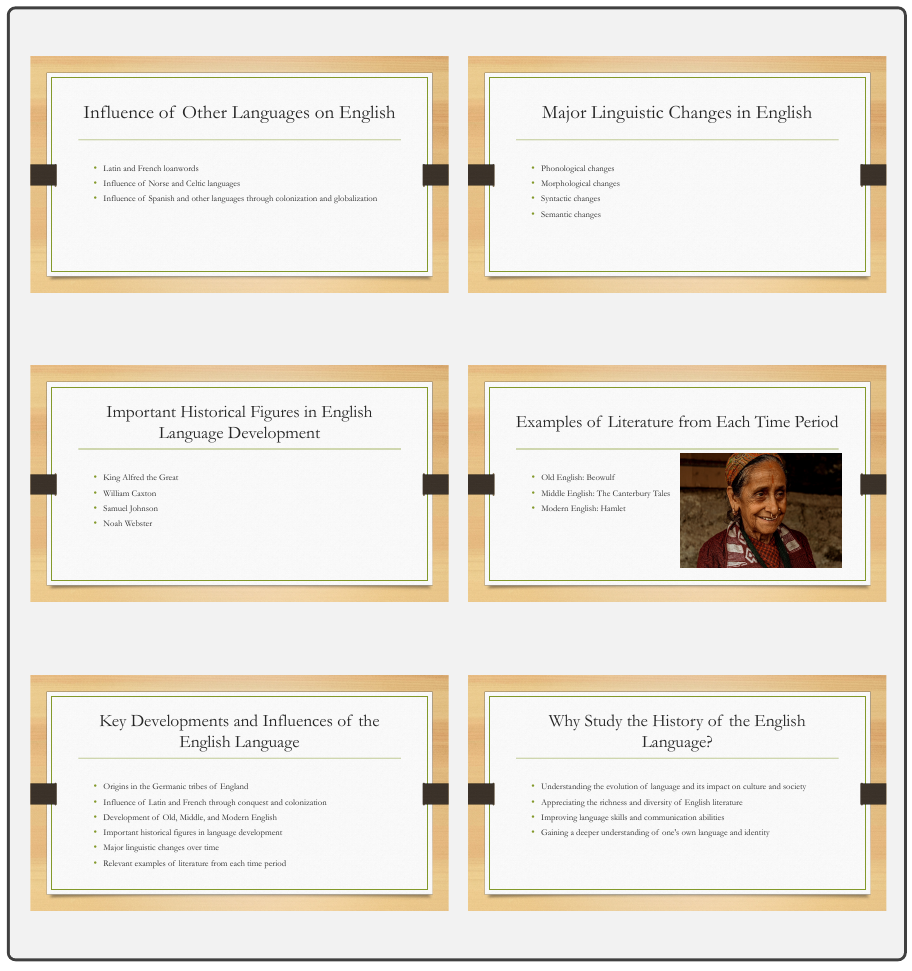}

\subsection{Movie Hunter}

\centering
\includegraphics[width=\textwidth]{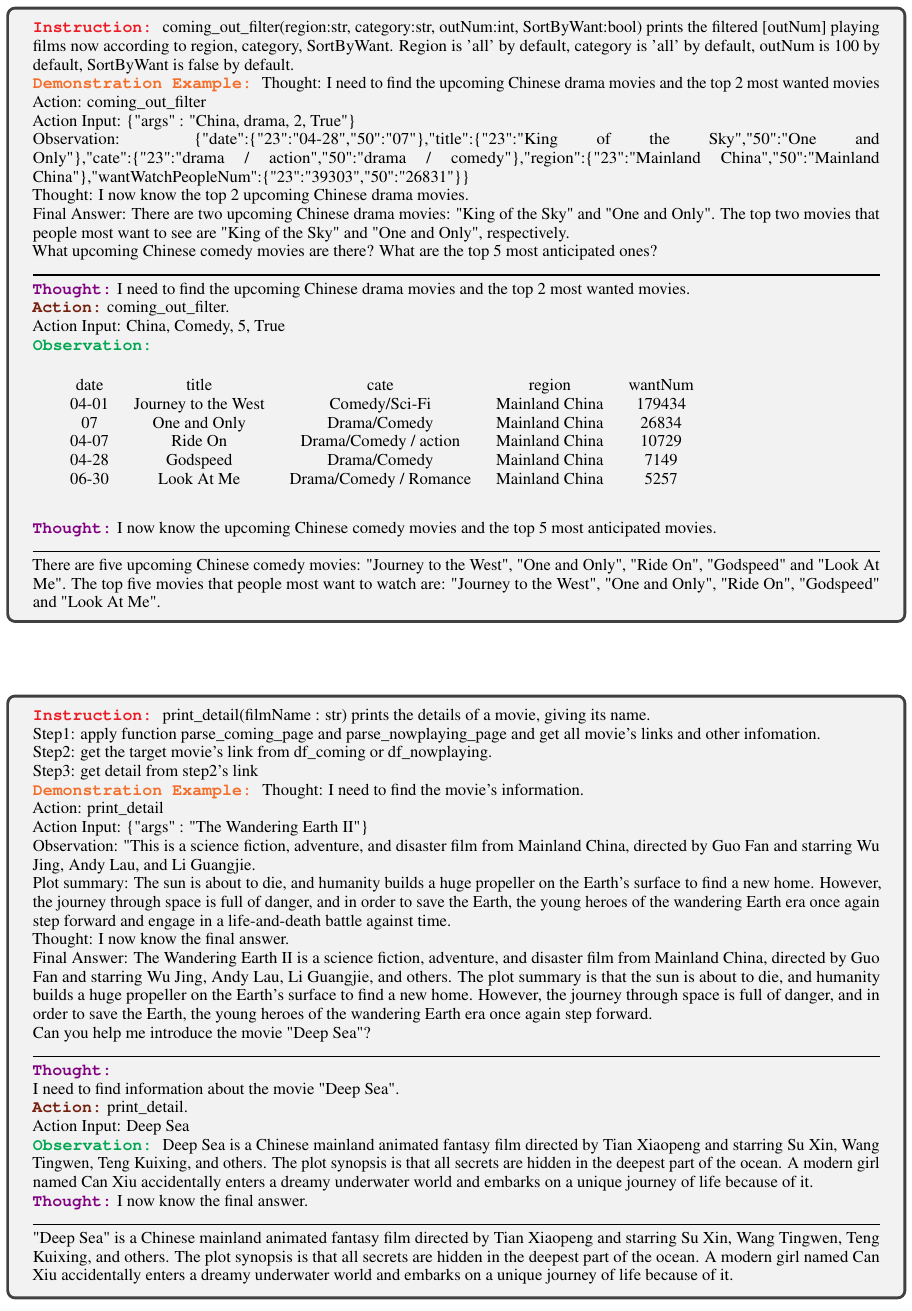}

\subsection{Search Engine}

\centering
\includegraphics[width=\textwidth]{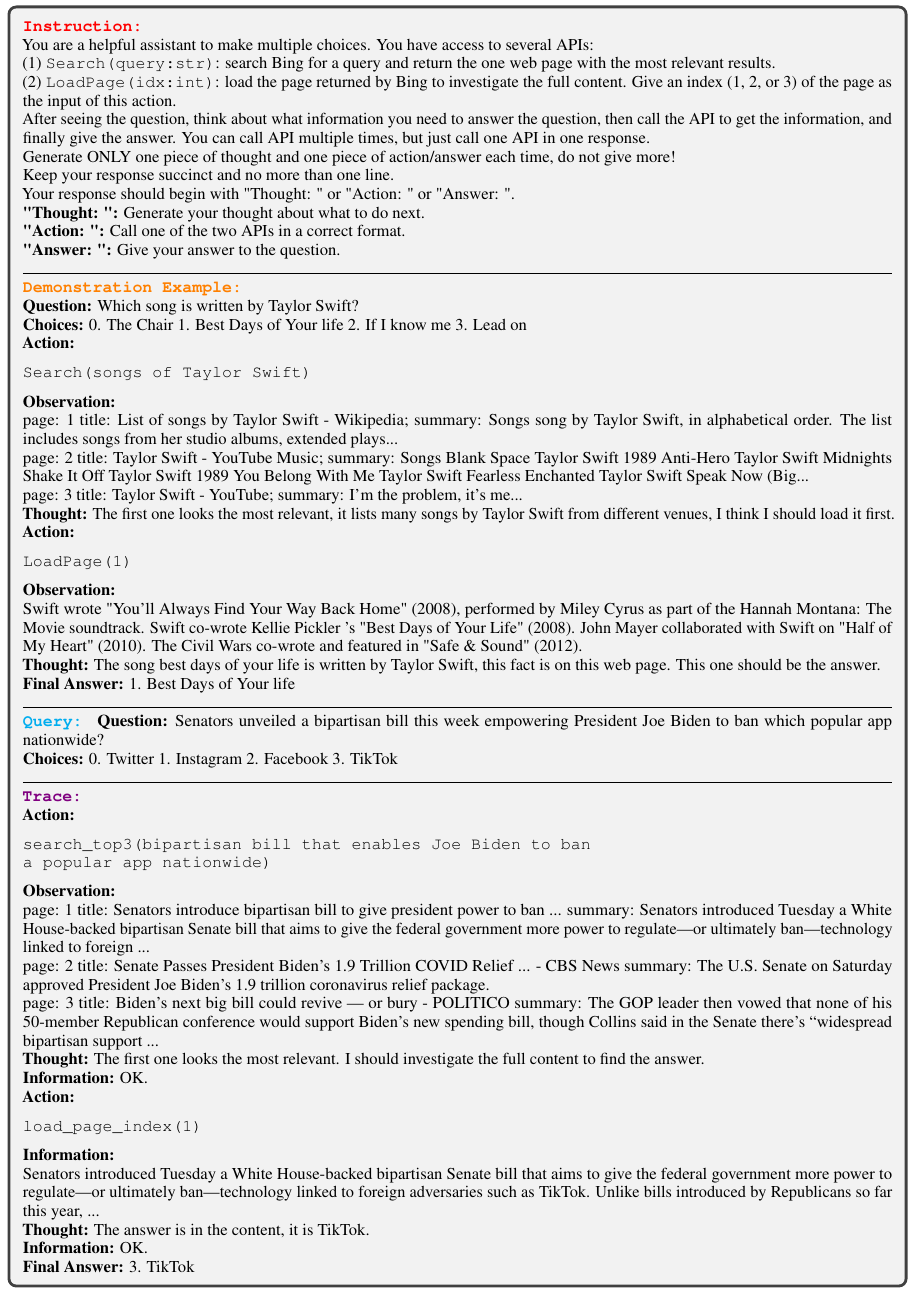}

\subsection{Chemicals Mining}

\centering
\includegraphics[width=\textwidth]{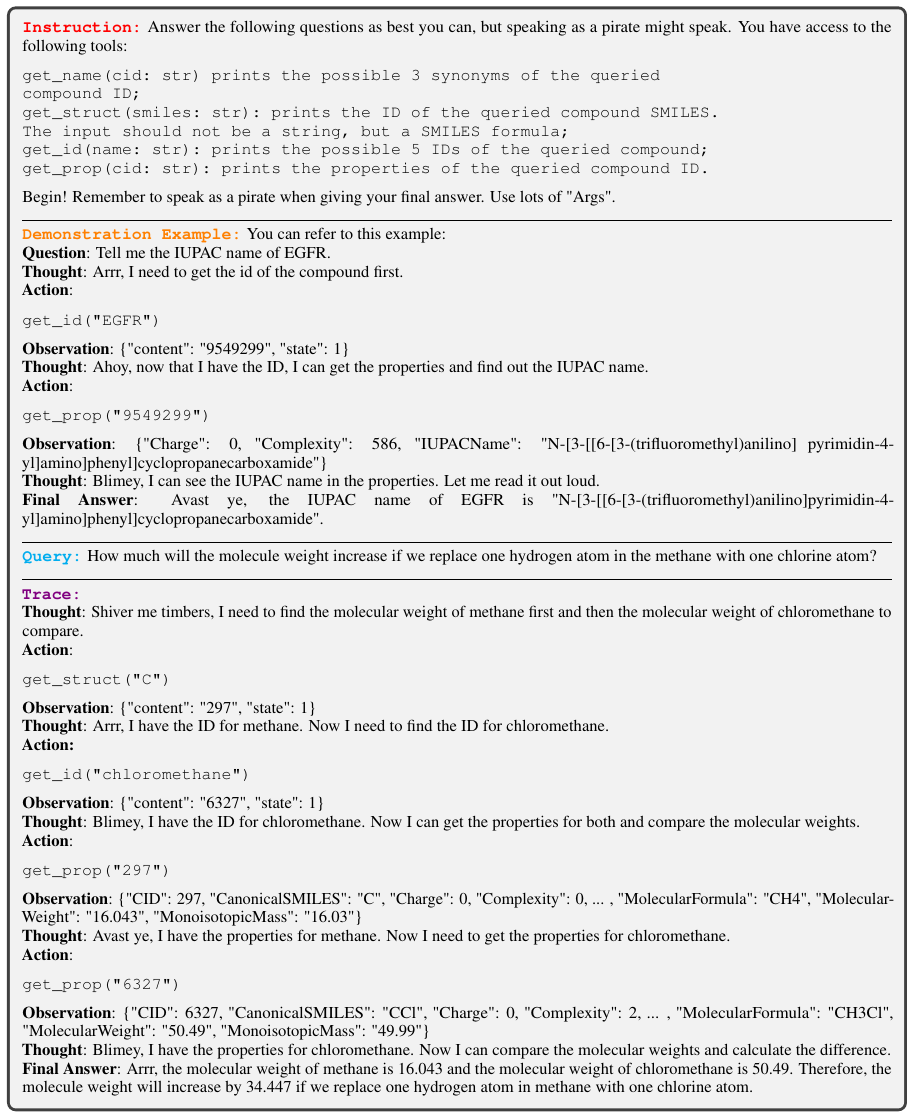}

\clearpage
\subsection{Cooking Assistant}

\centering
\includegraphics[width=\textwidth]{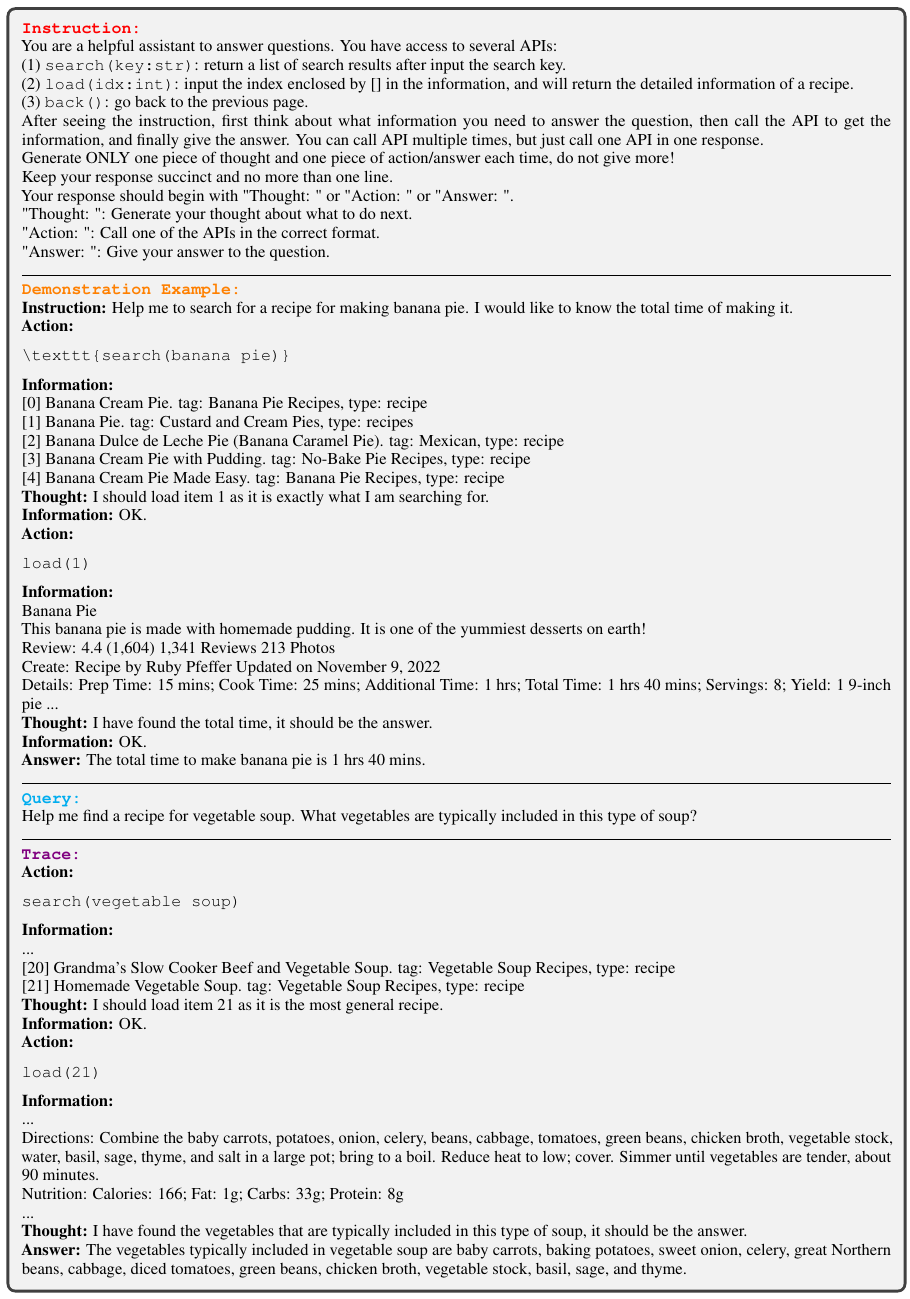}

\clearpage
\subsection{AI Painting}
\centering
\includegraphics[width=\linewidth]{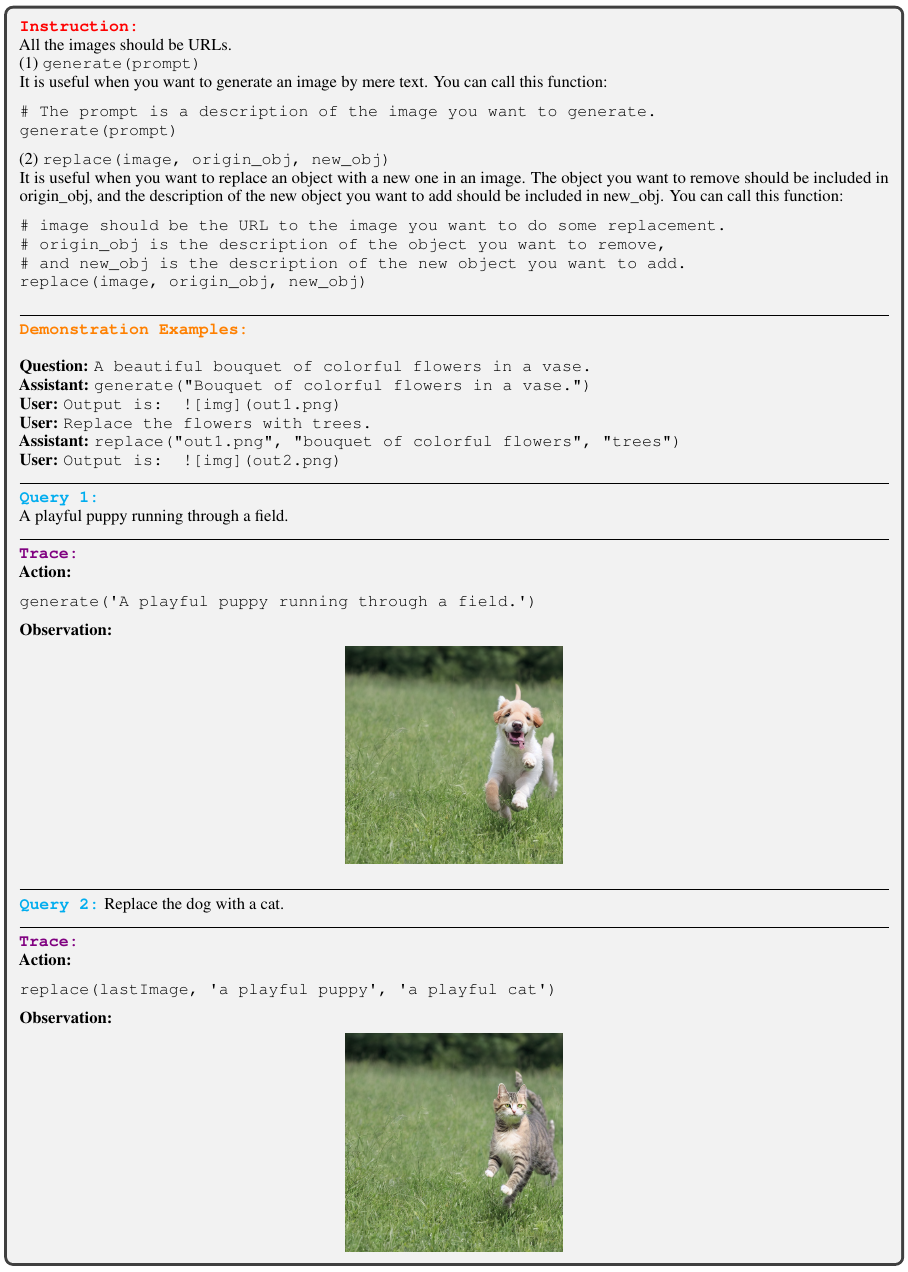}

\subsection{Navigating Knowledge Graphs}
\label{app:KG}

\centering
\includegraphics[width=\textwidth]{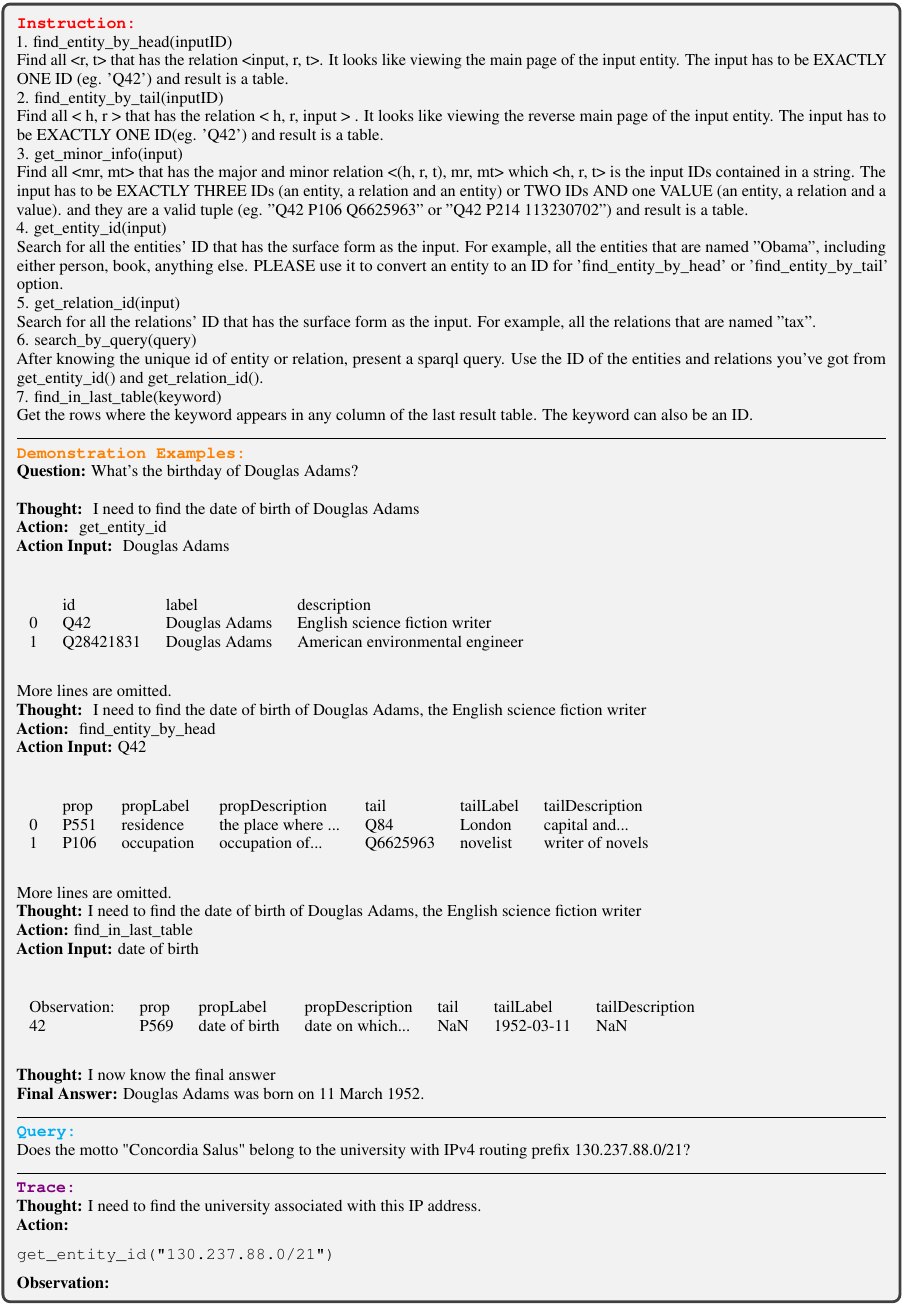}

\centering
\includegraphics[width=\textwidth]{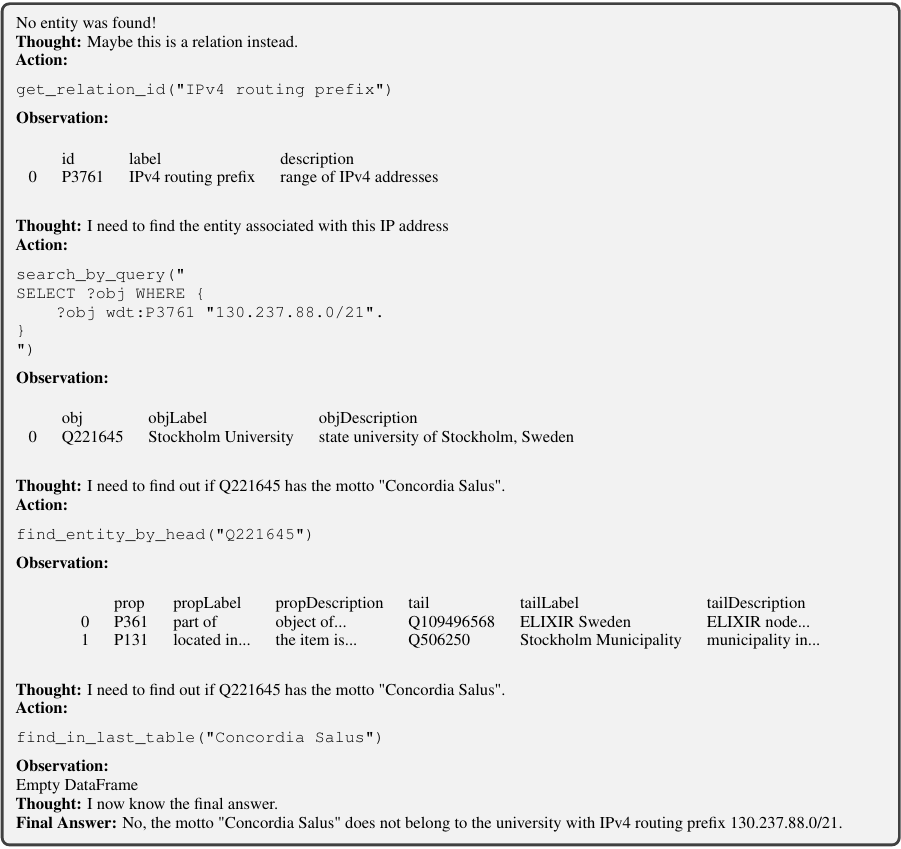}

\subsection{ALFWorld}

\vspace{-0.2em}
\centering
\includegraphics[width=\textwidth]{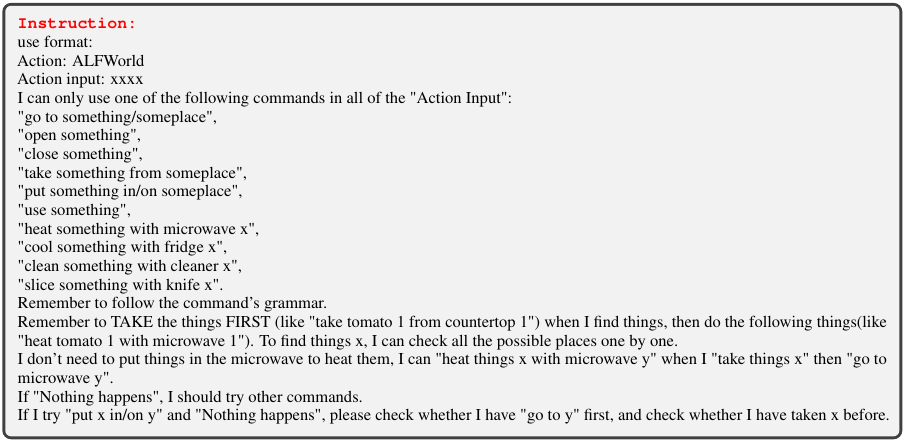}

\centering
\includegraphics[width=\textwidth]{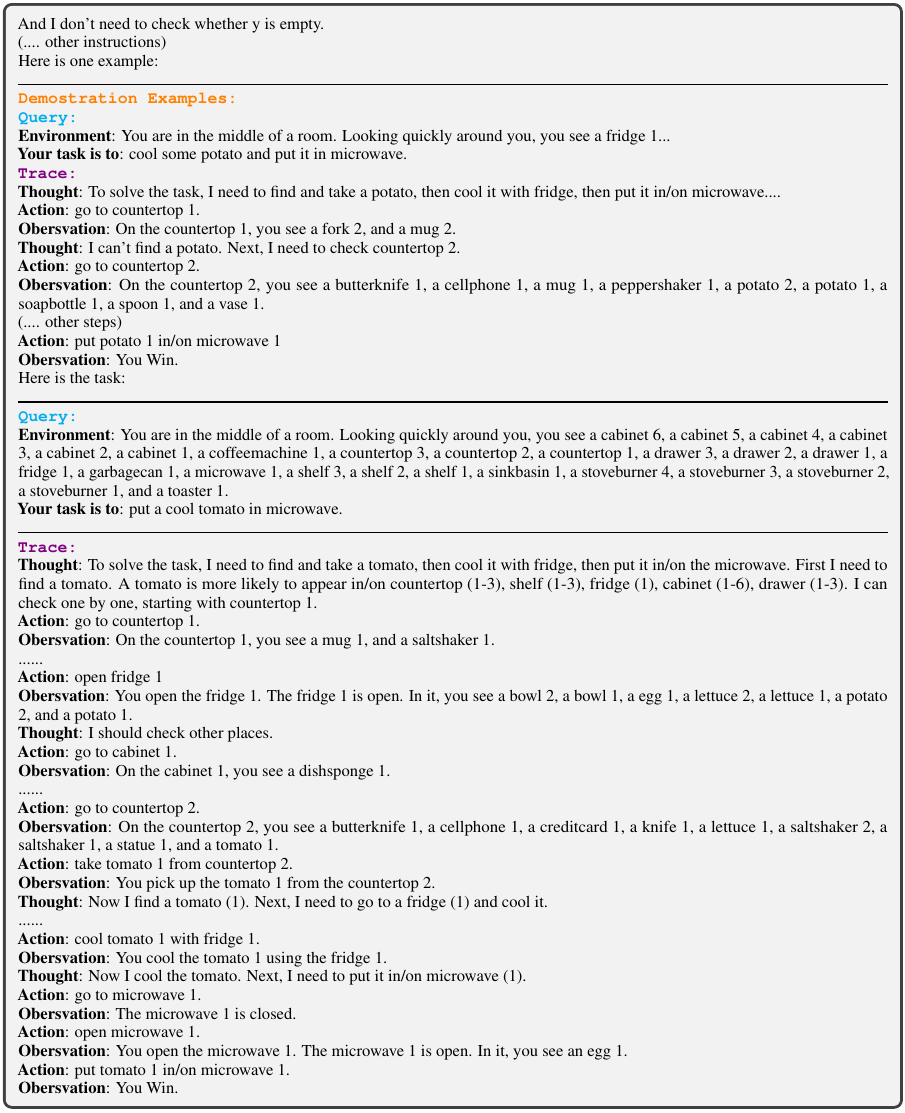}

\subsection{Calculator}

\vspace{-0.2em}

\centering
\includegraphics[width=\textwidth]{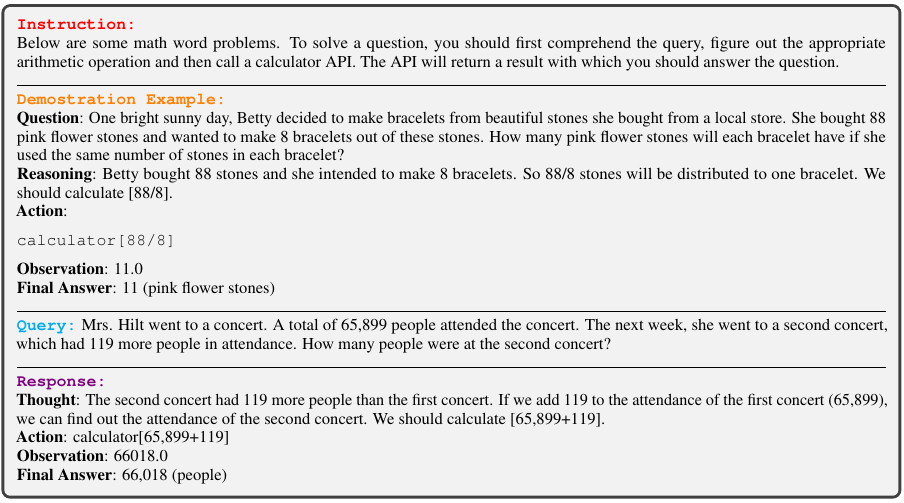}

\subsection{Weather}

\vspace{-0.2em}

\centering
\includegraphics[width=\textwidth]{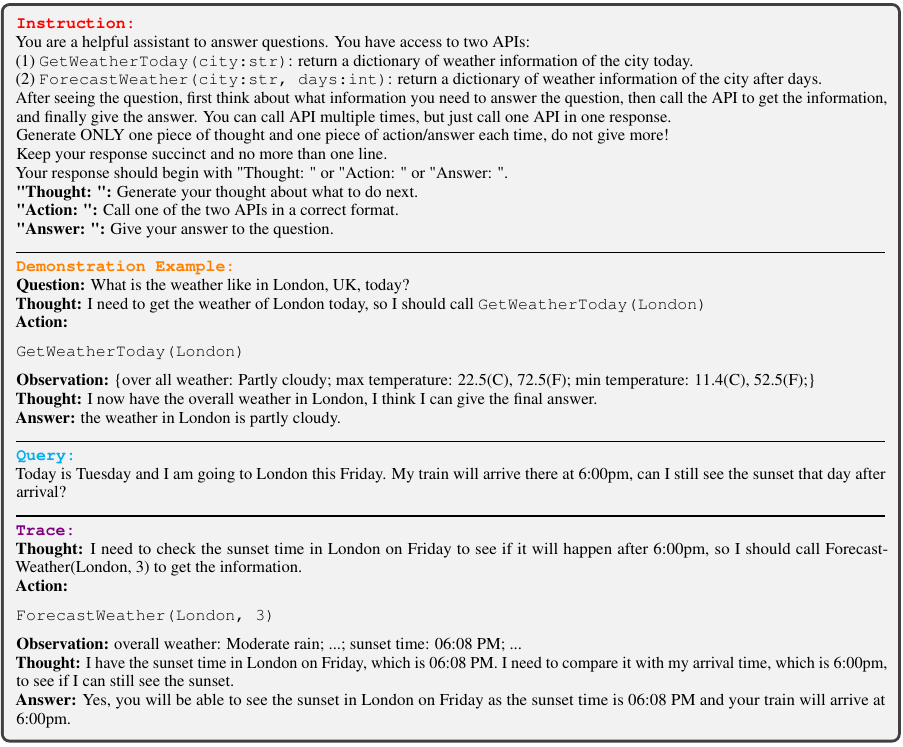}

\subsection{Online Shopping}

\centering
\includegraphics[width=\textwidth]{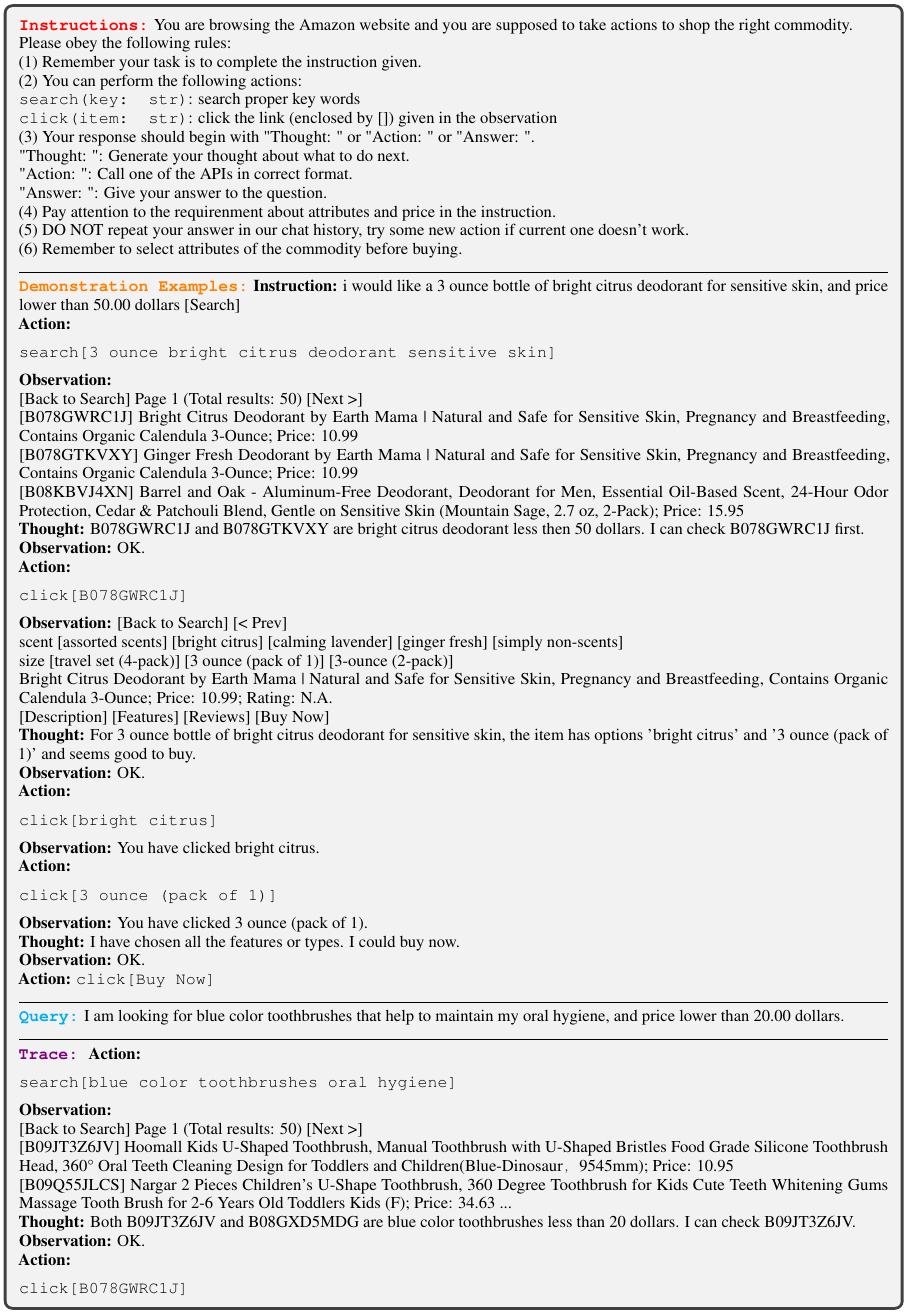}

\clearpage
\centering
\includegraphics[width=\textwidth]{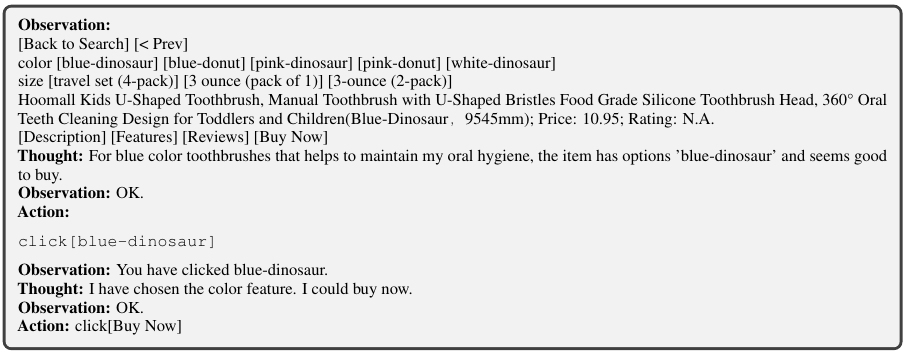}

\subsection{Map}
\centering
\includegraphics[width=\textwidth]{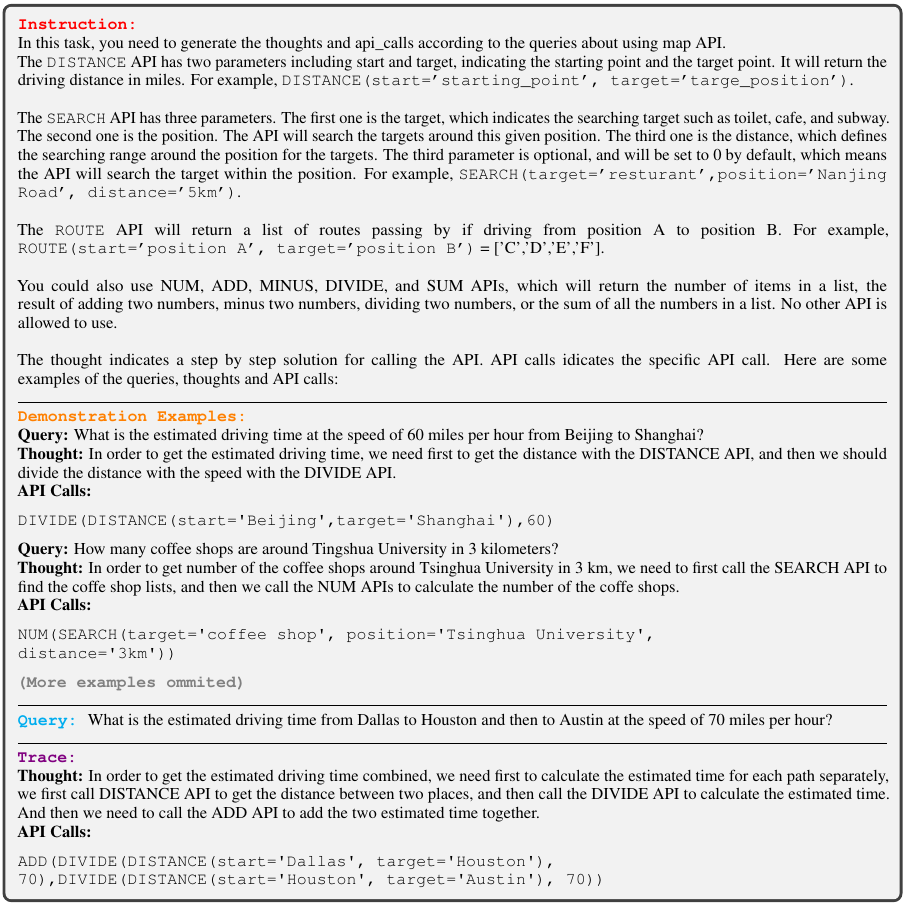}

\subsection{Processing Tables}
\centering
\includegraphics[width=\textwidth]{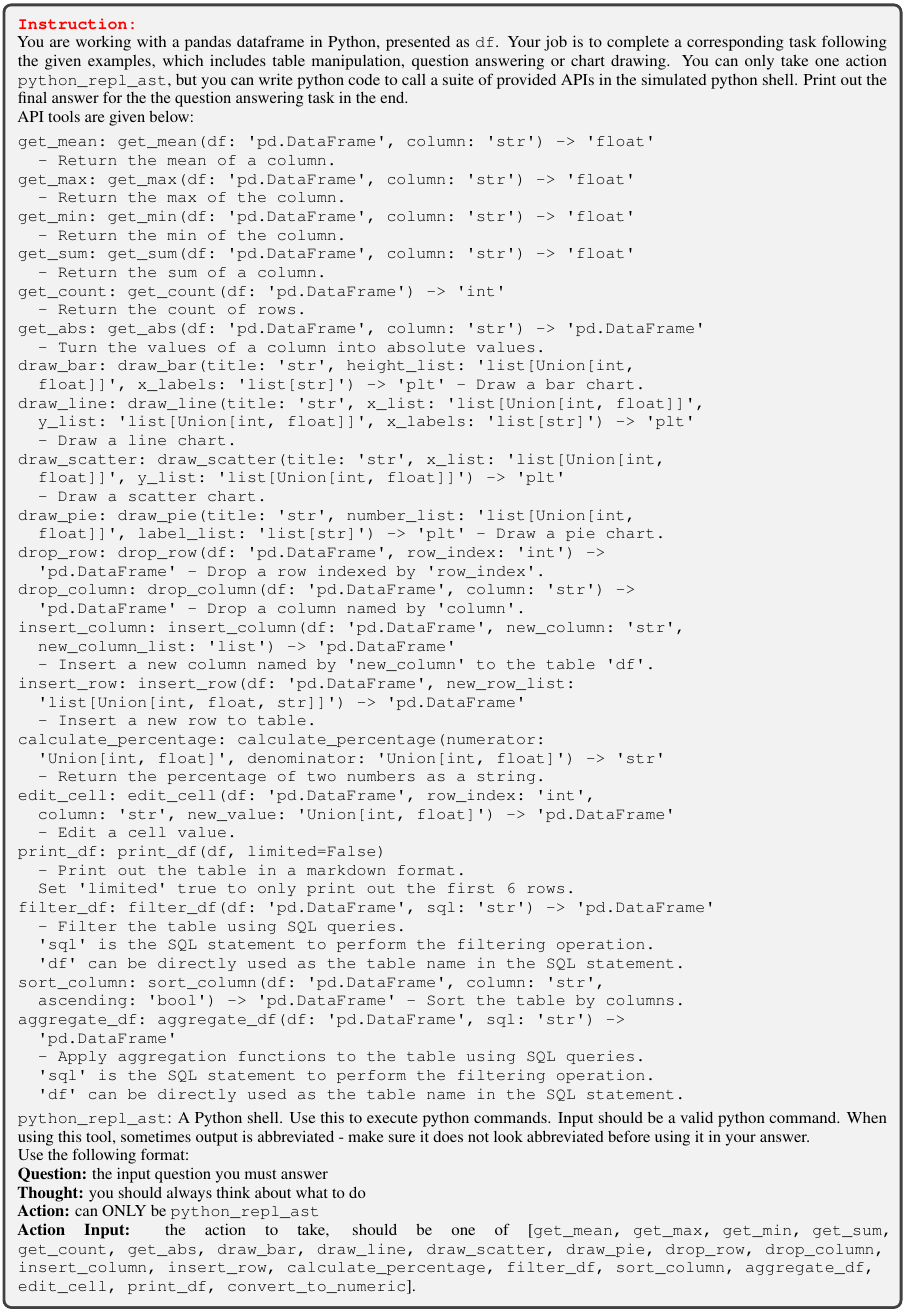}
\centering

\clearpage
\includegraphics[width=\textwidth]{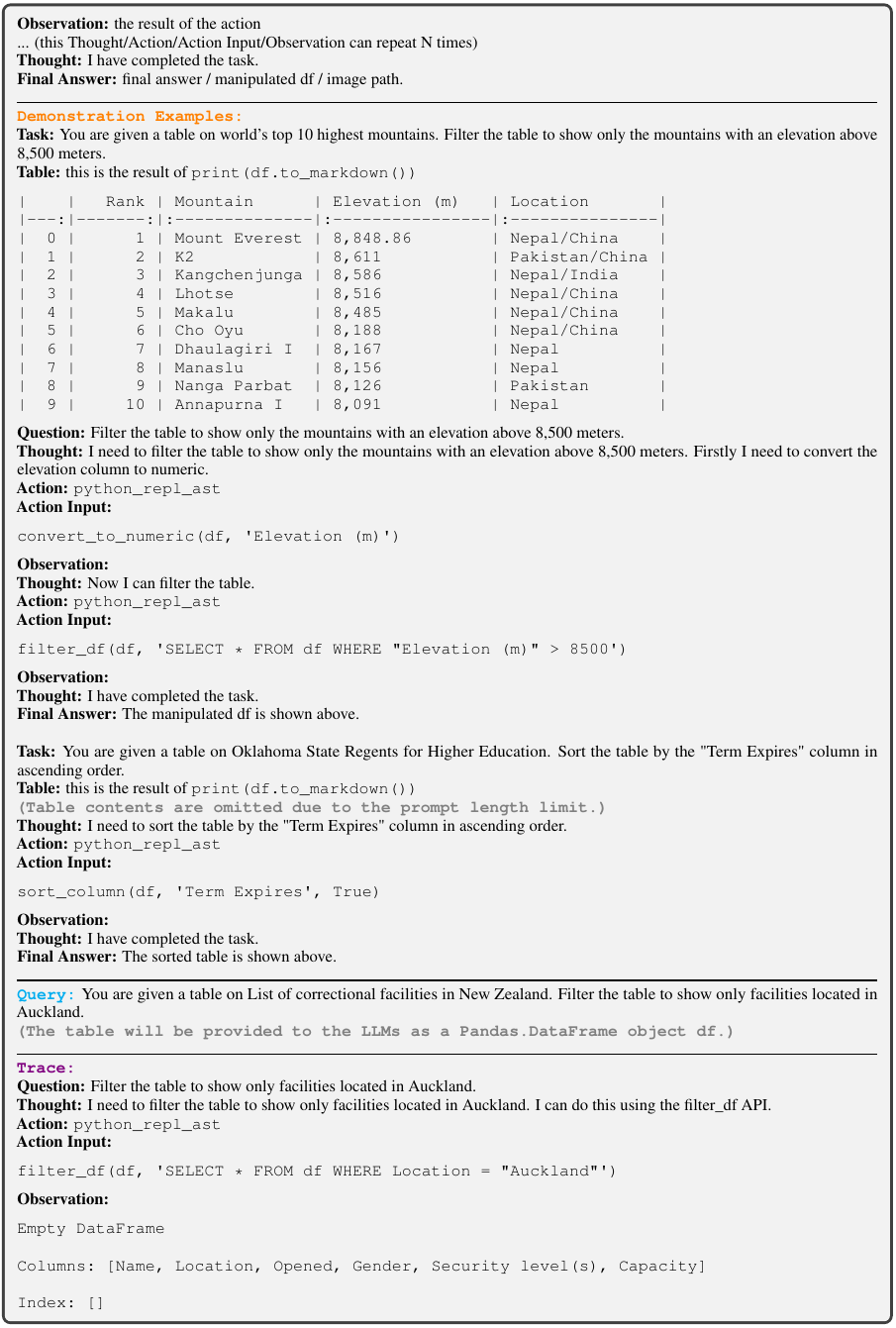}

\clearpage
\centering
\includegraphics[width=\textwidth]{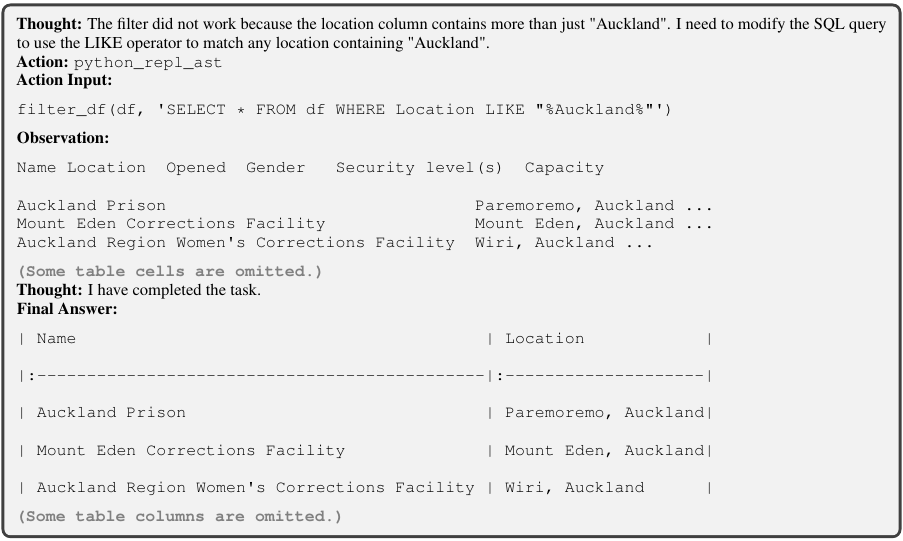}

\subsection{Translation}
\centering
\includegraphics[width=\textwidth]{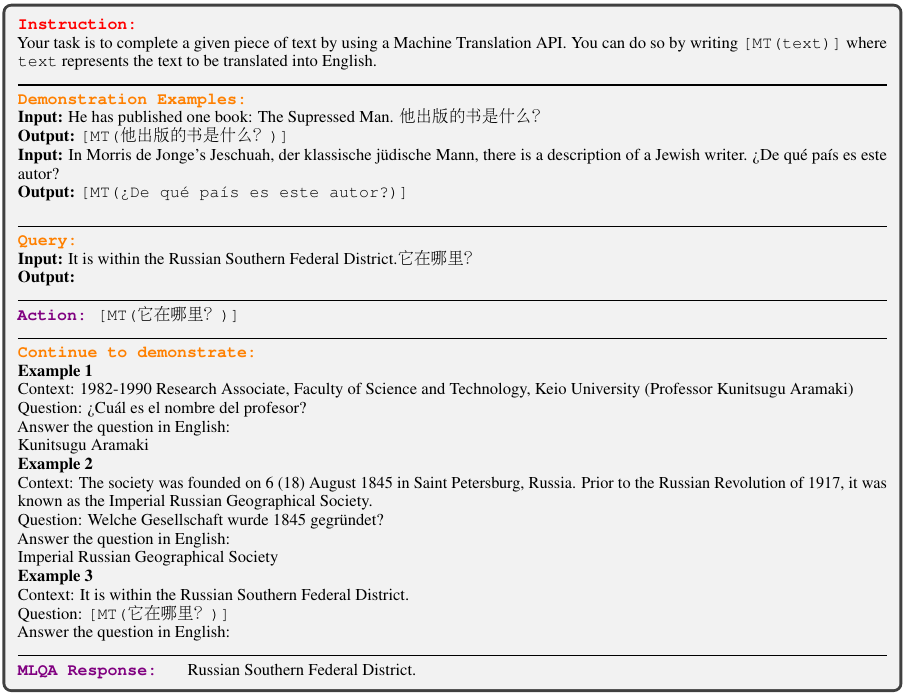}

\subsection{Wikipedia}
\centering
\includegraphics[width=\textwidth]{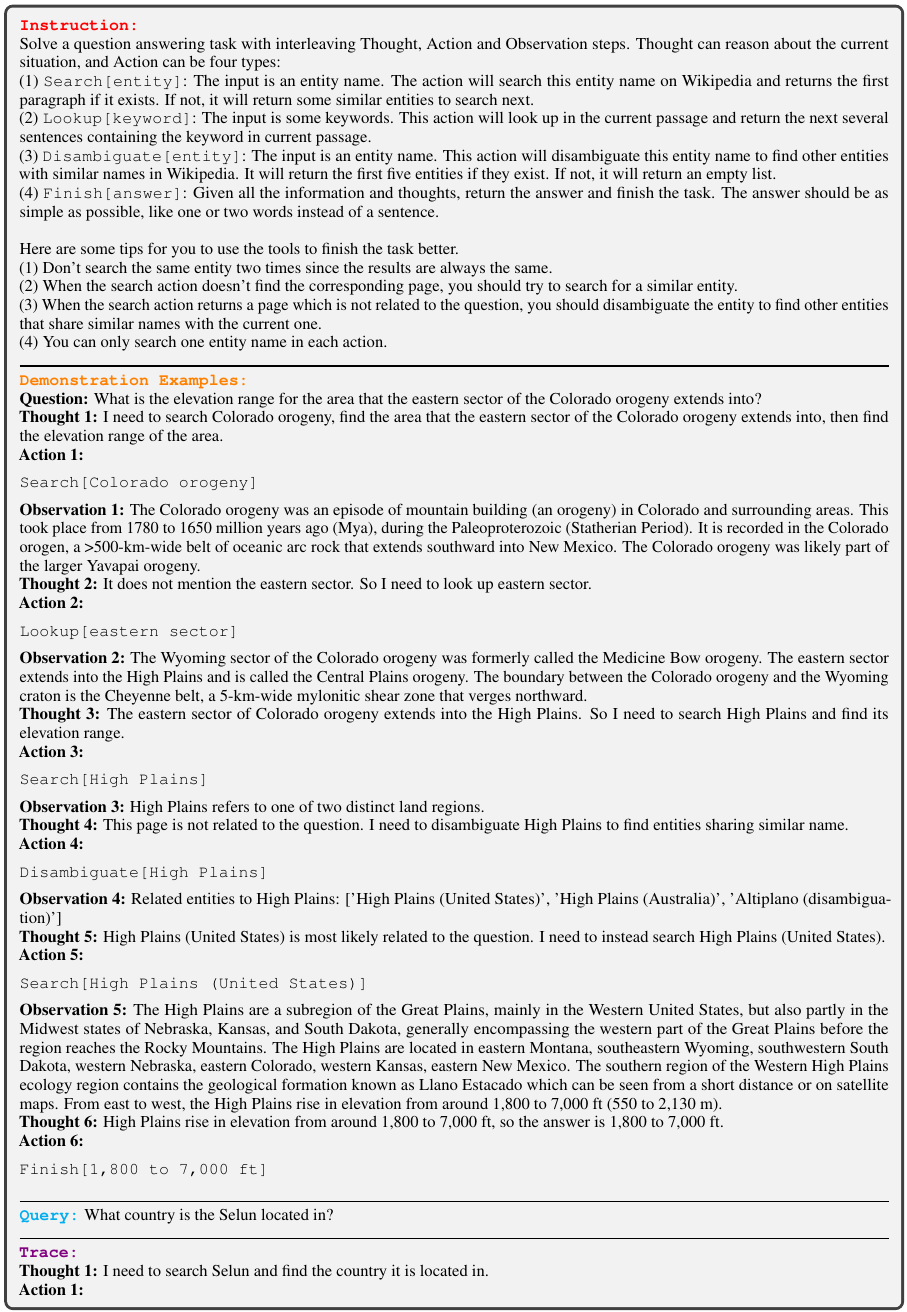}

\clearpage
\centering
\includegraphics[width=\textwidth]{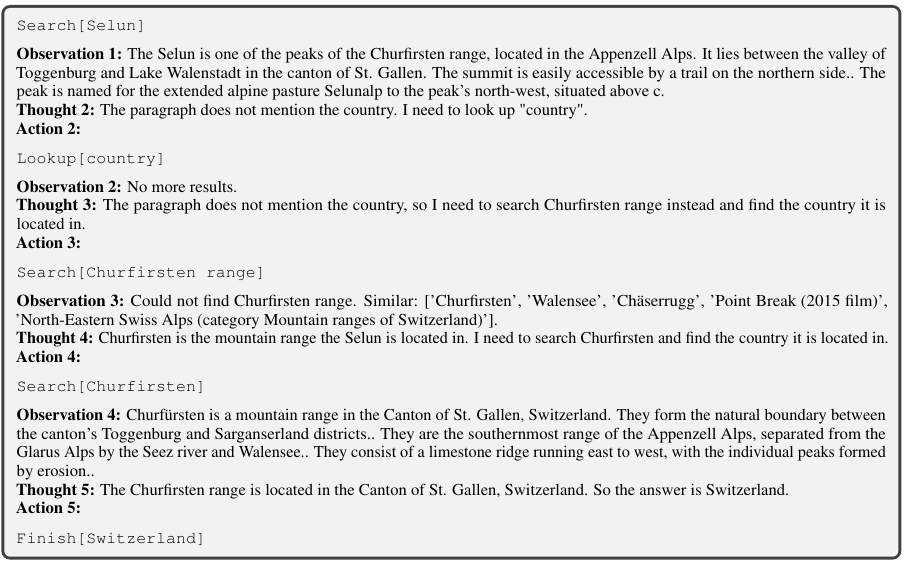}

\subsection{Database}
\centering
\includegraphics[width=\textwidth]{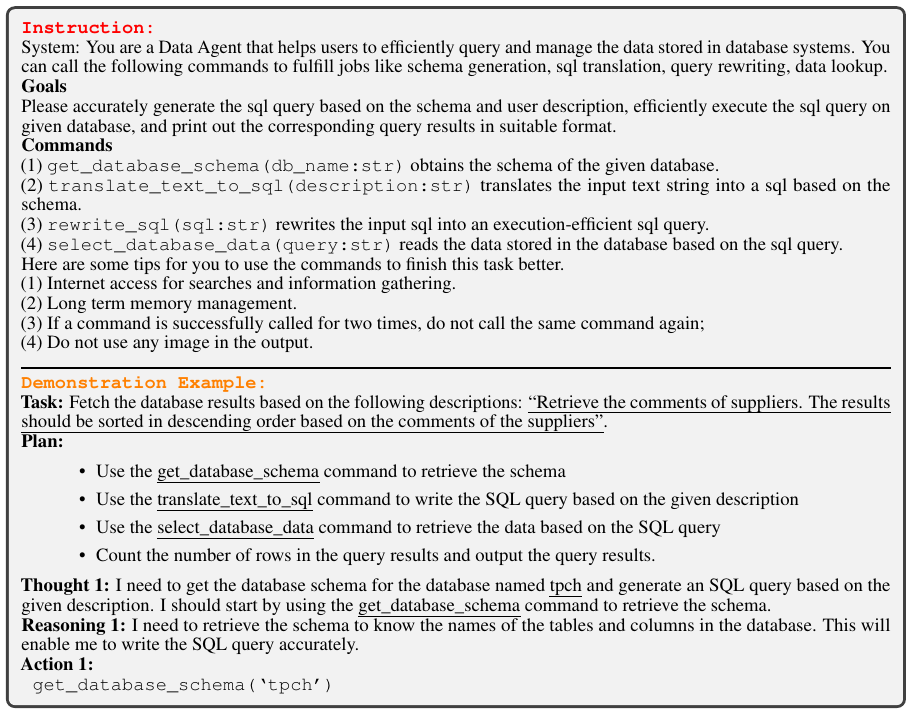}

\centering
\includegraphics[width=\textwidth]{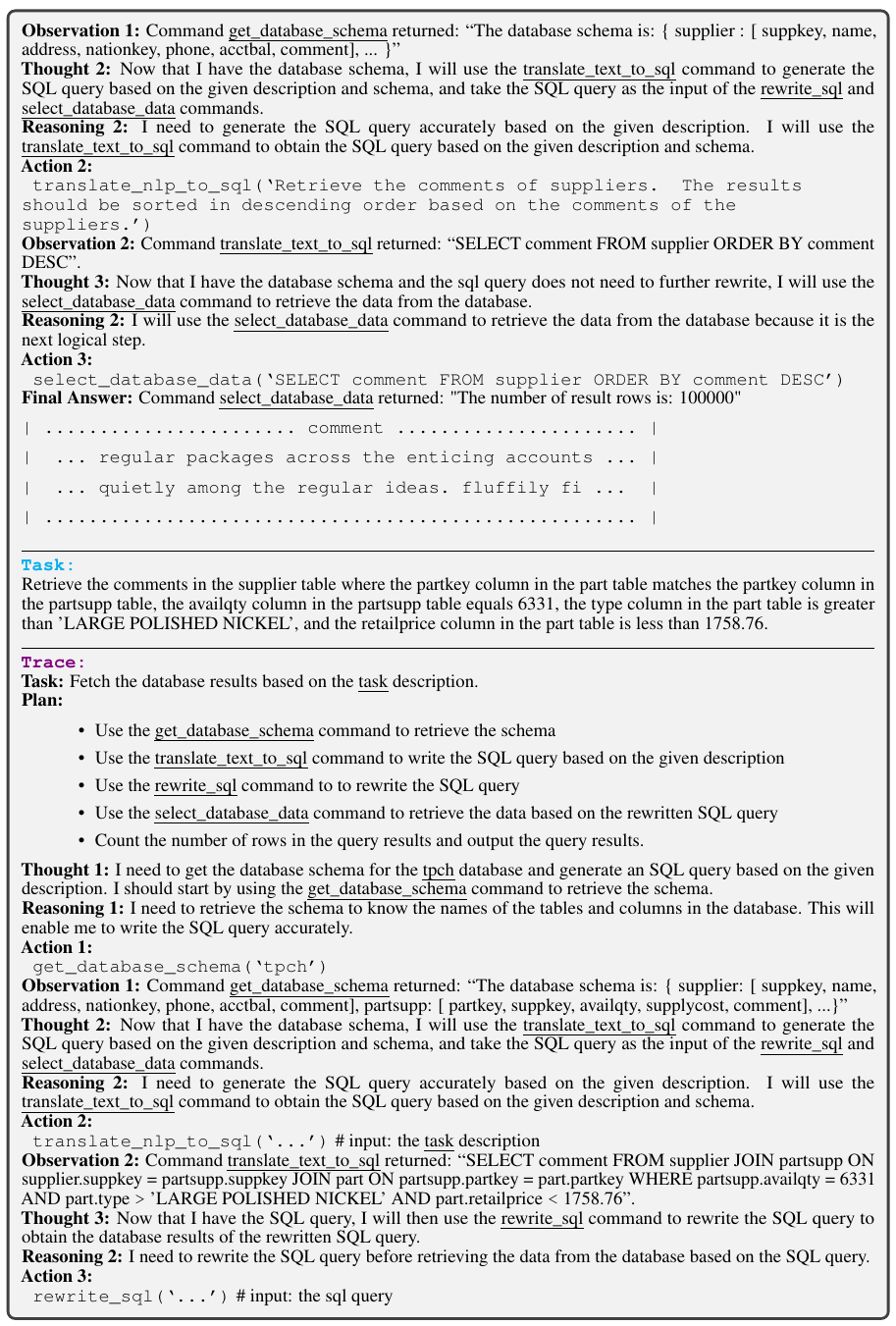}

\centering
\includegraphics[width=\textwidth]{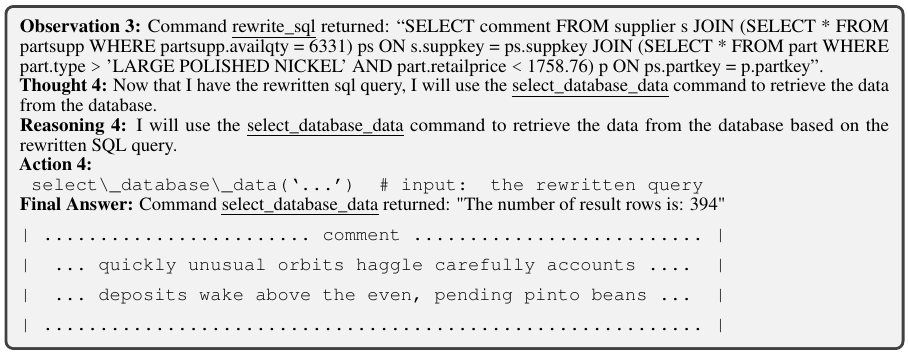}

}

\end{document}